\documentclass[10pt,twocolumn,letterpaper]{article}

\usepackage{iccv}

\newcommand{\finalcopy}{\iccvfinalcopy}
\makeatletter
\@namedef{ver@everyshi.sty}{}
\makeatother

\usepackage[dvipsnames,svgnames,x11names]{xcolor}
\usepackage{tikz}
\usetikzlibrary{arrows.meta,shapes,calc,matrix,fit,positioning,backgrounds,decorations.markings,fadings}
\usepackage{pgfplots}
\usepackage{pgfplotstable}
\pgfplotsset{compat=1.9}
\usepackage{xstring}

\usepgfplotslibrary{external}
\IfBeginWith*{\jobname}{fig/extern/}{\finalcopy}{}

\tikzstyle{every picture}+=[
	remember picture,
	every text node part/.style={align=center},
	every matrix/.append style={ampersand replacement=\&},
]
\tikzstyle{tight} = [inner sep=0pt,outer sep=0pt]
\tikzstyle{node}  = [draw,circle,tight,minimum size=12pt,anchor=center]
\tikzstyle{op}    = [draw,circle,tight]
\tikzstyle{dot}   = [fill,draw,circle,inner sep=1pt,outer sep=0]
\tikzstyle{pt}    = [fill,draw,circle,inner sep=1.5pt,outer sep=.2pt]
\tikzstyle{box}   = [draw,rectangle,inner sep=3pt]
\tikzstyle{high}  = [black!60]
\tikzstyle{group} = [high,box,opacity=.5]
\tikzstyle{dim1}  = [fill opacity=.3,text opacity=1]
\tikzstyle{dim2}  = [fill opacity=.5,text opacity=1]
\tikzstyle{dim3}  = [fill opacity=.7,text opacity=1]
\tikzstyle{rectc} = [tight,transform shape]
\tikzstyle{rect}  = [rectc,anchor=south west]

\newcommand{\leg}[1]{\addlegendentry{#1}}

\tikzset{every mark/.append style={solid}}
\pgfplotsset{%smooth,
	grid=both, width=\columnwidth, try min ticks=5,
	every axis/.append style={font=\small},
	every axis plot/.append style={thick,mark=none,mark size=1.8,tension=0.18},
	legend cell align=left, legend style={fill opacity=0.8},
	xticklabel={\pgfmathprintnumber[assume math mode=true]{\tick}},
	yticklabel={\pgfmathprintnumber[assume math mode=true]{\tick}},
	nodes near coords math/.style={
		nodes near coords={\pgfmathprintnumber[assume math mode=true]{\pgfplotspointmeta}},
	},
}

\pgfplotsset{
	dash/.style={mark=o,dashed,opacity=0.6},
	dott/.style={mark=o,dotted,opacity=0.6},
	nolim/.style={enlargelimits=false},
	plain/.style={every axis plot/.append style={},nolim,grid=none},
}

\pgfdeclarelayer{bg4}
\pgfdeclarelayer{bg3}
\pgfdeclarelayer{bg2}
\pgfdeclarelayer{bg1}
\pgfdeclarelayer{fg1}
\pgfdeclarelayer{fg2}
\pgfdeclarelayer{fg3}
\pgfdeclarelayer{fg4}
\pgfsetlayers{bg4,bg3,bg2,bg1,main,fg1,fg2,fg3,fg4}

\pgfkeys{/tikz/.cd, aspect/.store in=\aspect, aspect=1}
\pgfkeys{/tikz/.cd, depth/.store in=\depth, depth=.5}
\pgfkeys{/tikz/.cd, stepx/.store in=\step, stepx=1}

\tikzstyle{geom} = [line join=bevel,aspect=1,depth=.5,z={(\depth*\aspect,\depth)}]
\tikzstyle{wire} = [geom,draw,thick]

\def\cx[#1,#2,#3]{#1}
\def\cy[#1,#2,#3]{#2}
\def\cz[#1,#2,#3]{#3}
\def\ex[#1,#2,#3]{#1,0,0}
\def\ey[#1,#2,#3]{0,#2,0}
\def\ez[#1,#2,#3]{0,0,#3}

\makeatletter
\renewcommand\paragraph{\@startsection{paragraph}{4}{\z@}{1ex}{-1em}{\normalfont\normalsize\bfseries}}
\makeatother
\usepackage{times}
\usepackage{epsfig}
\usepackage{graphicx}
\usepackage{amsmath}
\usepackage{amssymb}

\usepackage{bbm}
\usepackage{booktabs}
\usepackage{multirow}
\usepackage{colortbl}
\usepackage[dvipsnames,svgnames,x11names]{xcolor}
\usepackage{adjustbox}
\usepackage{enumitem}
\usepackage[ruled,vlined,linesnumbered]{algorithm2e}
\usepackage{capt-of,etoolbox}
\usepackage[most]{tcolorbox}
\usepackage{colortbl}

\usepackage[pagebackref=true,breaklinks=true,letterpaper=true,colorlinks,bookmarks=false]{hyperref}

\iccvfinalcopy % *** Uncomment this line for the final submission

\def\iccvPaperID{1724} % *** Enter the ICCV Paper ID here

\newcommand{\eq}[1]{(\ref{eq:#1})}

\newcommand{\Th}[1]{\textsc{#1}}
\newcommand{\mr}[2]{\multirow{#1}{*}{#2}}
\newcommand{\mc}[2]{\multicolumn{#1}{c}{#2}}

\newcommand{\ch}{\checkmark}

\newcommand{\red}[1]{{\textcolor{red}{#1}}}
\newcommand{\blue}[1]{{\textcolor{blue}{#1}}}

\newcommand{\citeme}[1]{\red{[XX]}}
\newcommand{\refme}[1]{\red{(XX)}}

\newcommand{\fig}[2][1]{\includegraphics[width=#1\linewidth]{fig/#2}}

\newcommand{\tran}{^\top}

\newcommand{\ind}{\mathbbm{1}}

\newcommand{\nat}{\mathbb{N}}

\newcommand{\real}{\mathbb{R}}

\newcommand{\normal}{\mathcal{N}}

\newcommand{\mif}{\textrm{if}\ }

\newcommand{\id}{\operatorname{id}}

\newcommand{\diag}{\operatorname{diag}}

\newcommand{\defn}{\mathrel{:=}}

\newcommand{\inner}[1]{\left\langle{#1}\right\rangle}
\newcommand{\norm}[1]{\left\|{#1}\right\|}

\newcommand{\cA}{\mathcal{A}}

\newcommand{\cK}{\mathcal{K}}

\newcommand{\cP}{\mathcal{P}}
\newcommand{\cQ}{\mathcal{Q}}
\newcommand{\cR}{\mathcal{R}}
\newcommand{\cS}{\mathcal{S}}

\newcommand{\cV}{\mathcal{V}}

\newcommand{\cZ}{\mathcal{Z}}

\newcommand{\vK}{\mathbf{K}}

\newcommand{\vX}{\mathbf{X}}

\newcommand{\va}{\mathbf{a}}

\newcommand{\vd}{\mathbf{d}}

\newcommand{\vg}{\mathbf{g}}

\newcommand{\vk}{\mathbf{k}}

\newcommand{\vm}{\mathbf{m}}

\newcommand{\vq}{\mathbf{q}}

\newcommand{\Vs}{\mathbf{s}}

\newcommand{\vu}{\mathbf{u}}
\newcommand{\vv}{\mathbf{v}}

\newcommand{\vx}{\mathbf{x}}
\newcommand{\vy}{\mathbf{y}}
\newcommand{\vz}{\mathbf{z}}

\newcommand{\vone}{\mathbf{1}}
\newcommand{\vzero}{\mathbf{0}}

\newcommand{\vsigma}{{\boldsymbol{\sigma}}}

\makeatletter
\newcommand*\bdot{\mathpalette\bdot@{.7}}
\newcommand*\bdot@[2]{\mathbin{\vcenter{\hbox{\scalebox{#2}{$\m@th#1\bullet$}}}}}
\makeatother

\makeatletter
\DeclareRobustCommand\onedot{\futurelet\@let@token\@onedot}
\def\@onedot{\ifx\@let@token.\else.\null\fi\xspace}

\def\eg{\emph{e.g}\onedot} 
\def\ie{\emph{i.e}\onedot} 
 
 \def\vs{\emph{vs}\onedot}

\makeatother

\definecolor{TableColor}{rgb}{0.835, 0.894, 0.968}
\definecolor{CommonChoices}{rgb}{0.935, 0.935, 0.935}

\newcommand{\Col}[2]{{\color{#1}#2}}

\newcommand{\iavr}[1]{\Col{violet}{#1}}

\newcommand{\relu}{{\operatorname{relu}}}
\newcommand{\conv}{{\operatorname{conv}}}
\newcommand{\avg}{{\operatorname{avg}}}
\newcommand{\sink}{{\textsc{Sinkhorn}}}
\newcommand{\fc}{{\textsc{fc}}}

\newcommand{\gem}{{\textsc{GeM}}}
\newcommand{\lse}{{\textsc{lse}}}

\newcommand{\mlp}{{\textsc{mlp}}\xspace}
\newcommand{\msa}{{\textsc{msa}}}
\newcommand{\gru}{{\textsc{gru}}}
\newcommand{\LN}{{\textsc{ln}}}
\newcommand{\cls}{{\textsc{cls}}\xspace}
\newcommand{\our}{{\textsc{sp}}}
\newcommand{\ours}{{SimPool}}
\newcommand{\Ours}{{SimPool}\xspace}
\newcommand{\win}[2][RoyalBlue]{{\color{#1}#2}}              % winning choice
\newcommand{\hc}[2][CommonChoices]{{\cellcolor{#1}#2}}         % highlight cell
\newcommand{\hl}[2][CommonChoices]{\adjustbox{bgcolor=#1}{#2}} % highlight box
\newcommand{\lrn}[1]{{\color{ForestGreen}#1}}                % learned
\newcommand{\hp}[1]{{\color{DarkRed}#1}}                     % hyperparameter
\newcommand{\dt}[1]{{\color{RoyalBlue}#1}}                   % detailed in the appendix
\newcommand{\we}{\rowcolor{LightCyan}}                       % ours highlighted in table row
\newcommand{\us}[1]{{\cellcolor{LightCyan}#1}}               % ours highlighted in table cell
\newcommand{\all}{\tikz{\draw[white,fill=black,line width=0.4pt] (0,0) circle (1.2pt);}}

\newcommand{\imagenet}{{ImageNet-1k}\xspace}

\newcommand{\gain}[1]{}
\newcommand{\gp}[1]{{\color{ForestGreen}#1}}

\begin{document}

%%%%%%%%% TITLE
\title{Keep It \Ours: \\ Who Said Supervised Transformers Suffer from Attention Deficit?}

%\author{Bill Psomas\\
%National Technical University of Athens, IARAI\\
%Athens, Greece\\
%{\tt\small psomasbill@mail.ntua.gr}
%\and
%Ioannis Kakogeorgiou\\
%National Technical University of Athens\\
%Athens, Greece\\

\author{Bill~Psomas$^{1,2}$ \hspace{0.5em} Ioannis~Kakogeorgiou$^{1}$ \hspace{0.5em} Konstantinos~Karantzalos$^{1}$ \hspace{0.5em} Yannis~Avrithis$^{2}$ \vspace{0.5em} \\$^1$National Technical University of Athens \hspace{1.0em} \\ $^2$Institute of Advanced Research in Artificial Intelligence (IARAI) %\vspace{0.25em} \\ \tt{psomasbill@mail.ntua.gr}
}
%------------------------------------------------------------------------------
%------------------------------------------------------------------------------
% place teaser between title and abstract
\makeatletter
\apptocmd\@maketitle{{\teaser{}}}{}{}
\makeatother
%------------------------------------------------------------------------------
\newcommand{\teaser}{%
%------------------------------------------------------------------------------
\vspace{-6pt}
% \begin{figure*}[t]
\scriptsize
\centering
\setlength{\tabcolsep}{1.0pt}
\begin{tabular}{cccccccccc}

\fig[.096]{attmaps/ILSVRC2012_val_00019115_orig.PNG} &
\fig[.096]{attmaps/ILSVRC2012_val_00019115_supervised_official.PNG} &
\fig[.096]{attmaps/ILSVRC2012_val_00019115_supervised_simpool.PNG} &
\fig[.096]{attmaps/ILSVRC2012_val_00019115_dino_official.PNG} &
\fig[.096]{attmaps/ILSVRC2012_val_00019115_dino_simpool.PNG} &

\hspace{2pt}

\fig[.096]{attmaps/ILSVRC2012_val_00012528_orig.PNG} &
\fig[.096]{attmaps/ILSVRC2012_val_00012528_supervised_official.PNG} &
\fig[.096]{attmaps/ILSVRC2012_val_00012528_supervised_simpool.PNG} &
\fig[.096]{attmaps/ILSVRC2012_val_00012528_dino_official.PNG} &
\fig[.096]{attmaps/ILSVRC2012_val_00012528_dino_simpool.PNG} \\

\fig[.096]{attmaps/ILSVRC2012_val_00032993_orig.PNG} &
\fig[.096]{attmaps/ILSVRC2012_val_00032993_supervised_official.PNG} &
\fig[.096]{attmaps/ILSVRC2012_val_00032993_supervised_simpool.PNG} &
\fig[.096]{attmaps/ILSVRC2012_val_00032993_dino_official.PNG} &
\fig[.096]{attmaps/ILSVRC2012_val_00032993_dino_simpool.PNG} &

\hspace{2pt}

\fig[.096]{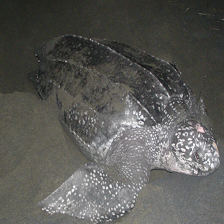} &
\fig[.096]{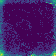} &
\fig[.096]{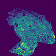} &
\fig[.096]{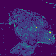} &
\fig[.096]{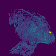} \\

input &
supervised &
supervised &
DINO~\cite{dino} &
DINO~\cite{dino} &

input &
supervised &
supervised &
DINO~\cite{dino} &
DINO~\cite{dino}\\

image &
\cls &
\Ours &
\cls &
\Ours &

image &
\cls &
\Ours &
\cls &
\Ours\\

% input &
% supervised &
% supervised &
% DINO~\cite{dino} &
% DINO~\cite{dino} &
% supervised &
% DINO~\cite{dino} \\
%
% image &
% ViT-S-\cls &
% ViT-S-\our &
% ViT-S-\cls &
% ViT-S-\our &
% ConvNeXt-S-\our &
% ConvNeXt-S-\our \\

\end{tabular}
\vspace{3pt}
\captionof{figure}{We introduce \Ours, a simple attention-based pooling method at the end of network, obtaining clean attention maps under supervision or self-supervision.
% , improving pre-training and downstream task performance.
Attention maps of ViT-S~\cite{vit} trained on ImageNet-1k~\cite{imagenet}. For baseline, we use the mean attention map of the \cls token. For \Ours, we use the attention map $\va$~\eq{sp-att}. Input image: $896 \times 896$; patches: $16 \times 16$; attention map: $56 \times 56$.
% ViT~\cite{vit} shown here but works equally well on convolutional networks (see \autoref{fig:vis-maps}).
}
\label{fig:teaser}
% \end{figure*}
%------------------------------------------------------------------------------
\par\vspace{24pt}}
%------------------------------------------------------------------------------

%------------------------------------------------------------------------------
\maketitle
%------------------------------------------------------------------------------

\ificcvfinal\thispagestyle{empty}\fi

%------------------------------------------------------------------------------
%%%%%%%%% ABSTRACT
%{\blfootnote{{Correspondence: \tt{psomasbill@mail.ntua.gr}}}}
\begin{abstract}
Convolutional networks and vision transformers have different forms of pairwise interactions, pooling across layers and pooling at the end of the network. Does the latter really need to be different?
As a by-product of pooling, vision transformers provide spatial attention for free, but this is most often of low quality unless self-supervised, which is not well studied. Is supervision really the problem?

In this work, we develop a generic pooling framework and then we formulate a number of existing methods as instantiations. By discussing the properties of each group of methods, we derive \Ours, a simple attention-based pooling mechanism as a replacement of the default one for both convolutional and transformer encoders. We find that, whether supervised or self-supervised, this improves performance on pre-training and downstream tasks and provides attention maps delineating object boundaries in all cases. One could thus call \Ours universal. To our knowledge, we are the first to obtain attention maps in supervised transformers of at least as good quality as self-supervised, without explicit losses or modifying the architecture. Code at: {\small \url{https://github.com/billpsomas/simpool}}.
\end{abstract}

%------------------------------------------------------------------------------
%%%%%%%%% BODY TEXT

\section{Introduction}
\label{sec:intro}

% %------------------------------------------------------------------------------
% \begin{table*}
% \centering
% \setlength{\tabcolsep}{3pt}
% \begin{tabular}{lccccc}
%        & \mc{2}{ViT} & \mc{2}{ResNet-50} & \\
%        %& \mc{2}{Supervised} & \mc{2}{Self-supervised} & \\ 
%        \multirow{2}{*}{\fig[.180]{attmaps/img475.png}}       &
% 	
%        \fig[.180]{attmaps/vits_official_supervised_img475.png}     &
% 	\fig[.180]{attmaps/vits_ours_supervised_img475.png}         &
% 	\fig[.180]{attmaps/img475.png}                              &
% 	\fig[.180]{attmaps/img475.png}                              & \\
% 
% 	& \fig[.180]{attmaps/vits_dino_official_img475.png}           &
% 	\fig[.180]{attmaps/vits_dino_ours_img475.png}               &
% 	\fig[.180]{attmaps/img475.png}                              &
% 	\fig[.180]{attmaps/img475.png}                              & \\
% 
% 
% 	& (a) Default                                                &
% 	(b) \Ours                                               &
% 	(c) Default                                               &
% 	(d) \Ours                                                &   \\
% \end{tabular}
% \end{table*}
% %------------------------------------------------------------------------------

Extracting visual representations and spatial pooling have been two interconnected processes since the study of 2D Gabor filters~\cite{gabor} and early convolutional networks~\cite{neo}. Modern \emph{convolutional networks}~\cite{resnet,convnext} gradually perform local pooling and downsampling throughout the architecture to extract a low-resolution feature tensor, followed by %a last step of
global spatial pooling. \emph{Vision transformers}~\cite{vit} only downsample at input tokenization and then preserve resolution, but pooling takes place again throughout the architecture via the interaction of patch tokens with a \cls token, inherited from language models \cite{bert}. 

The pooling operation has been studied extensively in instance-level tasks on convolutional networks~\cite{spoc,gem}, but less so in category-level tasks or transformers. Pooling in transformers is based on weighted averaging, using as weights the 2D \emph{attention map} of the \cls token at the last layer. However, this attention map is typically of low quality, unless under self-supervision~\cite{dino}.

In this work, we argue that vision transformers can be reformulated in two streams, where one is extracting a visual representation on patch tokens and the other is performing spatial pooling on the \cls token; whereas, convolutional networks undergo global spatial pooling at the very last step, before the classifier. In this sense, one can isolate the pooling process from both kinds of networks and replace it by a new one. This raises the following questions:
\emph{
\begin{enumerate}[itemsep=2pt, parsep=0pt, topsep=3pt]
	\item Can we derive a simple pooling process at the very last step of either convolutional or transformer encoders that improves over their default?
	\item Can this process provide high-quality attention maps that delineate object boundaries, for both networks?
	\item Do these properties hold under both supervised and self-supervised settings?
\end{enumerate}
}
To answer these questions, we develop a \emph{generic pooling framework}, parametrized by: (a) the number of vectors in the pooled representation; (b) whether pooling is iterative or not; (c) mappings at every stage of the process; (d) pairwise similarities, attention function and normalization; and (e) a function determining the pooling operation.

We then formulate a number of existing pooling methods as instantiations of this framework, including (a) simple pooling mechanisms in convolutional networks~\cite{resnet,mac,gem,lse,how}, (b) iterative methods on more than one vectors like $k$-means~\cite{otk,slot}, (c) feature re-weighting mechanisms originally desinged as network components rather than pooling~\cite{se,cbam}, and (d) vision transformers~\cite{vit, cait}. Finally, by discussing the properties of each group of methods, we derive a new, simple, attention-based pooling mechanism as a replacement of the default one for both convolutional and transformer encoders. \Ours provides high-quality attention maps that delineate object boundaries, under both supervised and self-supervised settings, as shown for ViT-S~\cite{vit} in \autoref{fig:teaser}.

In summary, we make the following contributions:
\begin{enumerate}[itemsep=2pt, parsep=0pt, topsep=3pt]
	\item We formulate a generic pooling framework that allows easy inspection and qualitative comparison of a wide range of methods.
	\item We introduce a simple, attention-based, non-iterative, universal pooling mechanism that provides a single vector representation and answers all the above questions in the affirmative.
	\item We conduct an extensive empirical study that validates the superior qualitative properties and quantitative performance of the proposed mechanism on standard benchmarks and downstream tasks.
\end{enumerate}

\section{Related Work}
\label{sec:related}

We discuss the most related work to pooling in convolutional networks and vision transformers. An extended version with more background is given in the appendix.

%------------------------------------------------------------------------------

\paragraph{Convolutional networks}

Early convolutional networks~\cite{neo, lenet} are based on learnable \emph{convolutional layers} interleaved with fixed \emph{spatial pooling layers} that downsample. The same design remains until today~\cite{alexnet, vgg, resnet, convnext}. Apart from mapping to a new space, convolutional layers involve a form of local pooling and pooling layers commonly take average~\cite{lenet} or maximum~\cite{hmax, alexnet}.

Early networks end in a fully-connected layer over a feature tensor of low resolution~\cite{lenet, alexnet, vgg}. This evolved into spatial pooling, \eg global / regional average followed by a classifier for category-level tasks like classification~\cite{nin, resnet} / detection~\cite{frcnn}, or global maximum followed by a pairwise loss~\cite{mac} for instance-level tasks. 

The spatial pooling operation at the end of the network is widely studied in instance level-tasks~\cite{spoc, mac, gem}, giving rise to forms of \emph{spatial attention}~\cite{crow, delf, delg, how, solar}, In category-level tasks, it is more common to study \emph{feature re-weighting} as components of the architecture~\cite{se, cbam, gatherexcite}. The two are closely related because \eg the weighted average is element-wise weighting followed by sum.

Pooling can be \emph{spatial}~\cite{gatherexcite, delf, delg, how, solar}, \emph{over channels}~\cite{se}, or both \cite{crow, cbam}. CBAM~\cite{cbam} is particularly related to our work in the sense that it includes global average pooling followed by a form of spatial attention, although the latter is not evident in its original formulation and although CBAM is 
not a pooling mechanism.

%------------------------------------------------------------------------------

\paragraph{Vision transformers}

\emph{Pairwise interactions} between features are forms of pooling or \emph{self-attention} over the spatial~\cite{nln, aa_net, san, n3_net} or channel dimensions~\cite{a2_net, eca}. Originating in language models~\cite{transformer}, \emph{vision transformers}~\cite{vit} streamlined these approaches and dominated the architecture landscape. Several variants often bring back ideas from convolutional networks~\cite{swin, rethinking, levit, cvt, convit, pit, metaformer}.

Transformers downsample only at the input, forming spatial \emph{patch tokens}. Pooling is based on a learnable \cls token, which, beginning at the input space, undergoes the same self-attention operation with patch tokens and provides a global image representation. That is, the network ends in global weighted average pooling, using as weights the attention of \cls over the patch tokens. 

Few works that have studied beyond \cls for pooling are mostly limited to global average pooling (GAP)~\cite{swin, rest, halonet, cnnvsvit}. \cls offers attention maps for free, however of low quality unless in a self-supervised setting~\cite{dino}, which is not well studied. Few works that attempt to rectify this in the supervised setting include a spatial entropy loss~\cite{spatialentropy}, shape distillation from convolutional networks~\cite{intriguing} and skipping computation of self-attention~\cite{skip}.

We attempt to address these limitations and study pooling in convolutional networks, vision transformers, supervised and self-supervised alike. We derive a simple, attention-based, universal pooling mechanism, improving both performance and attention maps.

\section{Method}
\label{sec:method}

We develop a generic pooling framework that encompasses many simple or more complex pooling methods, iterative or not, attention-based or not. We then examine a number of methods as instantiations of this framework. Finally, we discuss their properties and make particular choices in designing our solution.

%------------------------------------------------------------------------------

\subsection{A generic pooling framework}
\label{sec:frame}

\paragraph{Preliminaries}

Let $\vX \in \real^{d \times W \times H}$ be the $3$-dimensional \emph{feature tensor} obtained from the last layer of a network for a given input image, where $d$ is the number of feature channels and $W, H$ are the width and height. We represent the image by the \emph{feature} matrix $X \in \real^{d \times p}$ by flattening the spatial dimensions of $\vX$, where $p \defn W \times H$ is the number of spatial locations. Let $\vx_i \in \real^p$ denote the $i$-th row of $X$, that is, corresponding to the $2$-dimensional feature map in channel $i$, and $\vx_{\all j} \in \real^d$ denote the $j$-th column of $X$, that is, the feature vector of spatial location $j$.

By $\vone_n \in \real^n$, we denote the all-ones vector. Given an $m \times n$ matrix $A \ge 0$, by $\eta_1(A) \defn \diag(A \vone_n)^{-1} A$ we denote row-wise $\ell_1$-normalization; similarly, $\eta_2(A) \defn A \diag(\vone_m\tran A)^{-1}$ for column-wise.

\paragraph{Pooling process}

The objective of pooling is to represent the image by one or more vectors, obtained by interaction with $X$, either in a single step or by an iterative process. We denote the pooling process by function $\pi: \real^{d \times p} \to \real^{d' \times k}$ and the output vectors by matrix $U = \pi(X) \in \real^{d' \times k}$, where $d'$ is the number of dimensions, possibly $d' = d$, and $k$ is the number of vectors. In the most common case of a single vector, $k = 1$, we denote $U$ by $\vu \in \real^{d'}$. We discuss here the general iterative process; single-step pooling is the special case where the number of iterations is $1$.

\paragraph{Initialization}

We define $X^0 \defn X$ and make a particular choice for $U^0 \in \real^{d^0 \times k}$, where $d^0 \defn d$. The latter may depend on the input $X$, in which case it is itself a simple form of pooling or not; for example, it may be random or a learnable parameter over the entire training set.

\paragraph{Pairwise interaction}

Given $U^t$ and $X^t$ at iteration $t$, we define the \emph{query} and \emph{key} matrices
\begin{align}
	Q &= \phi_Q^t(U^t) \in \real^{n^t \times k}  \label{eq:query} \\
	K &= \phi_K^t(X^t) \in \real^{n^t \times p}. \label{eq:key}
\end{align}
Here, functions $\phi_Q^t: \real^{d^t \times k} \to \real^{n^t \times k}$ and $\phi_K^t: \real^{d^t \times p} \to \real^{n^t \times p}$ may be the identity, linear or non-linear mappings to a space of the same ($n^t = d^t$) or different dimensions. We let $K, Q$ interact pairwise by defining the $p \times k$ matrix $S(K, Q) \defn ((s(\vk_{\all i}, \vq_{\all j}))_{i=1}^p)_{j=1}^k$, where $s: \real^n \times \real^n \to \real$ for any $n$ is a similarity function. For example, $s$ can be dot product, cosine similarity, or a decreasing function of some distance. In the case of dot product, $s(\vx, \vy) \defn \vx\tran \vy$ for $\vx, \vy \in \real^d$, it follows that $S(K, Q) = K\tran Q \in \real^{p \times k}$.

\paragraph{Attention}

We then define the \emph{attention} matrix
\begin{align}
	A = h(S(K, Q)) \in \real^{p \times k}.
\label{eq:attn}
\end{align}
Here, $h: \real^{p \times k} \to [0,1]^{p \times k}$ is a nonlinear function that may be elementwise, for instance $\relu$ or $\exp$, normalization over rows or columns of $S(K, Q)$, or it may yield a form of correspondence or assignment between the columns of $K$ and $Q$, possibly optimizing a cost function.

%------------------------------------------------------------------------------
\begin{table*}
\centering
\scriptsize
\setlength{\tabcolsep}{2.5pt}
\begin{tabular}{cccccccccccccc} \toprule
\#         & \Th{Method}          & \Th{Cat} & \Th{Iter} & $k$      & $U^0$           & $\phi_Q(U)$             & $\phi_K(X)$                  & $s(\vx, \vy)$       & $A$                               & $\phi_V(X)$             & $f(x)$                     & $\phi_X(X)$                 & $\phi_U(Z)$                    \\ \midrule
\mr{5}{1}  & GAP~\cite{resnet}    & \ch      & \hc{}     & \hc{$1$} &                 &                         &                              &                     & $\vone_p/p$                       & \hc{$X$}                & $f_{-1}(x)$                &                             & \hc{$Z$}                       \\
           & $\max$~\cite{mac}    &          & \hc{}     & \hc{$1$} &                 &                         &                              &                     & $\vone_p$                         & \hc{$X$}                & $f_{-\infty}(x)$           &                             & \hc{$Z$}                       \\
           & GeM~\cite{gem}       &          & \hc{}     & \hc{$1$} &                 &                         &                              &                     & $\vone_p/p$                       & \hc{$X$}                & \hc{$f_{\lrn{\alpha}}(x)$} &                             & \hc{$Z$}                       \\
           & LSE~\cite{lse}       & \ch      & \hc{}     & \hc{$1$} &                 &                         &                              &                     & $\vone_p/p$                       & \hc{$X$}                & $e^{\lrn{r}x}$             &                             & \hc{$Z$}                       \\
           & HOW~\cite{how}       &          & \hc{}     & \hc{$1$} &                 &                         &                              &                     & $\diag(X\tran X)$                 & $\fc(\dt{\avg}_3(X))$   & $f_{-1}(x)$                &                             & \hc{$Z$}                       \\ \midrule
\mr{3}{2}  & OTK~\cite{otk}       & \ch      &           & $\hp{k}$ & $\lrn{U}$       & $U$                     & $X$                          & $-\|\vx-\vy\|^2$    & $\dt{\sink}(e^{S/\epsilon})$      & $\dt{\psi}(X)$          & $f_{-1}(x)$                &                             & $Z$                            \\
           & $k$-means            &          & \ch       & $\hp{k}$ & random          & $U$                     & $X$                          & $-\|\vx-\vy\|^2$    & $\eta_2(\dt{\arg\max}_1(S))$      & $X$                     & $f_{-1}(x)$                & $X$                         & $Z$                            \\
           & Slot~\cite{slot}$^*$ & \ch      & \ch       & $\hp{k}$ & $\lrn{U}$       & \hc{$\lrn{W_Q} U$}      & \hc{$\lrn{W_K} X$}           & \hc{$\vx\tran \vy$} & \hc{$\vsigma_2(S/\sqrt{d})$}      & $\lrn{W_V} X$           & $f_{-1}(x)$                & $X$                         & $\lrn{\mlp}(\lrn{\gru}(Z))$    \\ \midrule
\mr{2}{3}  & SE~\cite{se}         & \ch      & \hc{}     & \hc{$1$} & \hc{$\pi_A(X)$} & $\sigma(\lrn{\mlp}(U))$ &                              &                     &                                   & $\diag(\vq) X$          &                            & $V$                         &                                \\
           & CBAM~\cite{cbam}$^*$ & \ch      & \hc{}     & \hc{$1$} & \hc{$\pi_A(X)$} & $\sigma(\lrn{\mlp}(U))$ & $X$                          & \hc{$\vx\tran \vy$} & $\sigma(\lrn{\conv_7}(S))$        & $\diag(\vq) X$          &                            & $V \diag(\va)$              &                                \\ \midrule
\mr{2}{4}  & ViT~\cite{vit}$^*$   & \ch      & \ch       & $1$      & $\lrn{U}$       & $\dt{g}_m(\lrn{W_Q} U)$ & $\dt{g}_m(\lrn{W_K} X)$      & \hc{$\vx\tran \vy$} & $\vsigma_2(S_i/\sqrt{d})_{i=1}^m$ & $\dt{g}_m(\lrn{W_V} X)$ & $f_{-1}(x)$                & $\lrn{\mlp}(\lrn{\msa}(X))$ & $\lrn{\mlp}(\dt{g}_m^{-1}(Z))$ \\
           & CaiT~\cite{cait}$^*$ & \ch      & \ch       & $1$      & $\lrn{U}$       & $\dt{g}_m(\lrn{W_Q} U)$ & $\dt{g}_m(\lrn{W_K} X)$      & \hc{$\vx\tran \vy$} & $\vsigma_2(S_i/\sqrt{d})_{i=1}^m$ & $\dt{g}_m(\lrn{W_V} X)$ & $f_{-1}(x)$                & $X$                         & $\lrn{\mlp}(\dt{g}_m^{-1}(Z))$ \\ \midrule \we
5          & \Ours                & \ch      &           & $1$      & $\pi_A(X)$      & $\lrn{W_Q} U$           & $\lrn{W_K} X$                & $\vx\tran \vy$      & $\vsigma_2(S/\sqrt{d})$           & $X - \min X$            & $f_{\hp{\alpha}}(x)$       &                             & $Z$                            \\ \bottomrule
\end{tabular}
\vspace{3pt}
\caption{A landscape of pooling methods. \Th{Cat}: used in category-level tasks; \Th{Iter}: iterative; *: simplified. $\pi_A$: GAP; $\sigma$: sigmoid; $\vsigma_2$: softmax over columns; $\eta_2$: column normalization; $g_m$: partitioning in $m$ groups (see appendix). \hl[LightCyan]{Cyan}: ours; \hl{gray}: common choices with ours; \lrn{green}: learnable; \hp{red}: hyperparameter; \dt{blue}: detailed in the appendix.}
\label{tab:land}
\end{table*}
%------------------------------------------------------------------------------

\paragraph{Attention-weighted pooling}

We define the \emph{value} matrix
\begin{align}
	V &= \phi_V^t(X^t) \in \real^{n^t \times p}.
\label{eq:value}
\end{align}
Here, function $\phi_V^t: \real^{d^t \times p} \to \real^{n^t \times p}$ plays a similar role with $\phi_Q^t, \phi_K^t$. \emph{Attention-weighted pooling} is defined by
\begin{align}
	Z = f^{-1}(f(V) A) \in \real^{n^t \times k}.
\label{eq:pool}
\end{align}
Here, $f: \real \to \real$ is a nonlinear elementwise function that determines the pooling operation, for instance, average or max-pooling. The product $f(V) A$ defines $k$ linear combinations over the columns of $f(V)$, that is, the features at different spatial locations. If the columns of $A$ are $\ell_1$-normalized, then those are convex combinations. Thus, matrix $A$ defines the weights of an averaging operation.

\paragraph{Output} Finally, we define the output matrices corresponding to image features and pooling,
\begin{align}
	X^{t+1} &= \phi_X^t(X^t) \in \real^{d^{t+1} \times p}  \label{eq:feat-next} \\
	U^{t+1} &= \phi_U^t(Z)   \in \real^{d^{t+1} \times k}. \label{eq:pool-next}
\end{align}
Functions $\phi_X^t: \real^{n^t \times p} \to \real^{d^{t+1} \times p}$ and $\phi_U^t: \real^{n^t \times k} \to \real^{d^{t+1} \times k}$ play a similar role with $\phi_Q^t, \phi_K^t, \phi_V^t$ but also determine the dimensionality $d^{t+1}$ for the next iteration.

At this point, we may iterate by returning to the ``pairwise interaction'' step, or terminate, yielding $U^{t+1}$ as $U$ with $d' = d^{t+1}$. Non-iterative methods do not use $\phi_X^t$.

%------------------------------------------------------------------------------

\subsection{A pooling landscape}
\label{sec:land}

\autoref{tab:land} examines %, in groups,
a number of pooling methods as instantiations of our framework. The objective is to get insight into their basic properties. How this table was obtained is detailed in the appendix.

\emph{Group 1} consists of simple methods with $k=1$ that are not attention-based and have been studied in category-level tasks~\cite{resnet,lse} or mostly in instance-level tasks~\cite{mac, gem, how}. Here, the attention is a vector $\va \in \real^p$ and either is uniform or depends directly on $X$, by pooling over channels~\cite{how}. Most important is the choice of pooling operation by function $f$. Log-sum-exp~\cite{lse} arises with $f(x) = e^{rx}$ with learnable scale $r$. For the rest, we define $f = f_\alpha$, where
\begin{align}
	f_\alpha(x) \defn \left\{
	\begin{array}{ll}
		x^{\frac{1-\alpha}{2}}, & \mif \alpha \ne 1, \\
		\ln x,                  & \mif \alpha = 1.
	\end{array} \right.
\label{eq:alpha}
\end{align}
As studied by Amari~\cite{amari2007integration}, function $f_\alpha$ is defined for $x \ge 0$ ($\alpha \ne 1$) or $x>0$ ($\alpha=1$). It reduces to the maximum, quadratic mean (RMS), arithmetic mean, geometric mean, harmonic mean, and minimum for $\alpha = -\infty,-3,-1,1,3,+\infty$, respectively. It has been proposed as a transition from average to max-pooling~\cite{theor} and is known as GeM~\cite{gem}, with $\gamma = (1-\alpha)/2 > 1$ being a learnable parameter.

\emph{Group 2} incorporates iterative methods with $k > 1$, including standard $k$-means, the soft-clustering variant Slot Attention~\cite{slot} and optimal transport between $U$ and $X$~\cite{otk}. The latter is not formally iterative according to our framework, but the Sinkhorn algorithm is iterative internally.

\emph{Group 3} refers to methods introduced as modules within the architecture rather than pooling mechanisms~\cite{se,cbam}. An interesting aspect is initialization of $U^0$ by \emph{global average pooling} (GAP) on $X$:
\begin{align}
	\pi_A(X) \defn X \vone_p / p = \frac{1}{p} \sum_{j=1}^p \vx_{\all j} \in \real^d,
\label{eq:gap}
\end{align}
where $\vone_p \in \real^p$ is the all-ones vector. Channel attention ($\phi_Q(U)$) and spatial attention ($A$) in CBAM~\cite{cbam} are based on a few layers followed by sigmoid, playing the role of a binary classifier (\eg foreground/background); whereas, transformer-based attention uses directly the query and softmax normalization, respectively. Although not evident in the original formulation, we show in the appendix that there is pairwise interaction.

\emph{Group 4} refers to vision transformers~\cite{vit,cait}, which we reformulate in two separate streams, one for the \cls token, $U$, and another for the patch tokens, $X$. We observe that, what happens to the \cls token throughout the entire encoder, is an iterative pooling process. Moreover, although $U$ is just one vector, multi-head attention splits it into $m$ subvectors, where $m$ is the number of heads. Thus, $m$ is similar to $k$ in $k$-means. The difference of CaiT~\cite{cait} from ViT~\cite{vit} is that this iteration happens only in the last couple of layers, with the patch embeddings $X$ being fixed.

%------------------------------------------------------------------------------
\begin{figure}
\centering
\fig[1.01]{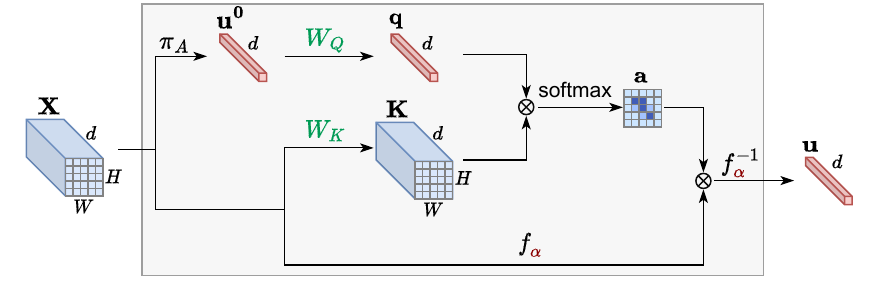}
\caption{\emph{Overview of \Ours}. Given an input tensor $\vX \in \real^{d \times W \times H}$ flattened into $X \in \real^{d \times p}$ with $p \defn W \times H$ patches, one stream forms the initial representation $\vu^0 = \pi_A(X) \in \real^d$~\eq{sp-init} by \emph{global average pooling} (GAP), mapped by $\lrn{W_Q} \in \real^{d \times d}$~\eq{sp-query} to form the \emph{query} vector $\vq \in \real^d$. Another stream maps $X$ by $\lrn{W_K} \in \real^{d \times d}$~\eq{sp-key} to form the \emph{key} $K \in \real^{d \times p}$, shown as tensor $\vK$. Then, $\vq$ and $K$ interact to generate the attention map $\va \in \real^p$~\eq{sp-att}. Finally, the pooled representation $\vu \in \real^d$ is a generalized weighted average of $X$ with $\va$ determining the weights and scalar function $f_\alpha$ determining the pooling operation~\eq{sp}.}
\label{fig:overview}
\end{figure}
%------------------------------------------------------------------------------

\subsection{\Ours}
\label{sec:ours}

\emph{Group 5} of \autoref{tab:land} is our method, \Ours. A schematic overview is given in \autoref{fig:overview}.

\paragraph{Pooling process}

We are striving for a simple design. While pooling into $k > 1$ vectors would yield a more discriminative representation, either these would have to be concatenated, as is the case of multi-head attention, or a particular similarity kernel would be needed beyond dot product, which we consider to be beyond the scope of this work. We rather argue that it is the task of the encoder to learn a single vector representation of objects, even if those are composed of different parts. This argument is stronger when pre-training is performed on images mostly depicting one object, like \imagenet.

We observe in \autoref{tab:land} that only methods explicitly pooling into $k > 1$ vectors or implicitly using $m > 1$ heads are iterative. We explain why in the next paragraph. Following this insight, we perform pooling in a single step.

In summary, our solution is limited to a single vector $\vu \in \real^d$ for pooling, that is, $k = 1$, and is non-iterative.

\paragraph{Initialization}

We observe in \autoref{tab:land} that single-step attention-based methods in Group 3 initialize $\vu^0$ by GAP. We hypothesize that, since attention is based on pairwise similarities, it is essential that $\vu^0$ is chosen such that its similarities with $X$ are maximized on average, which would help to better discriminate between foreground (high similarity) and background (low similarity). Indeed, for $s(\vx, \vy) = -\|\vx-\vy\|^2$, the sum of squared Euclidean distances of each column $\vx_{\all i}$ of $X$ to $\vu \in \real^d$
\begin{align}
	J(\vu) = \frac{1}{2} \sum_{i=1}^p \|\vx_{\all i} - \vu\|^2
\label{eq:cost}
\end{align}
is a convex distortion measure with unique minimum the average of vectors $\{\vx_{\all i}\}$
\begin{align}
	\vu^* \defn \arg\min_{\vu \in \real^d} J(\vu) =
		\frac{1}{p} \sum_{i=1}^p \vx_{\all i} = \pi_A(X),
\label{eq:opt}
\end{align}
which can be found in closed form. By contrast, for $k > 1$ vectors, distortion can only be minimized iteratively, \eg by $k$-means.
% A more general result is given by Amari~\cite{amari2007integration}, whereby the $\alpha$-integration of a number of probability distributions is optimal under the $\alpha$-divergence measure.
We therefore choose:
\begin{align}
	\vu^0 = \pi_A(X) = X \vone_p / p.
\label{eq:sp-init}
\end{align}

\paragraph{Pairwise interaction, attention}

We follow the attention mechanism of transformers, in its simplest possible form. In particular, we use a single head, $m = 1$, like Slot Attention~\cite{slot} (which however uses $k$ vectors). We find that the query and key mappings are essential in learning where to attend as a separate task from learning the representation for the given task at hand. In particular, we use linear mappings $\phi_Q, \phi_K$ with learnable parameters $W_Q, W_K \in \real^{d \times d}$ respectively:
\begin{align}
	\vq &= \phi_Q(\vu^0) = W_Q \vu^0 \in \real^d             \label{eq:sp-query} \\
	K   &= \phi_K(X)     = W_K X     \in \real^{d \times p}. \label{eq:sp-key}
\end{align}
As in transformers, we define pairwise similarities as dot product, that is, $S(K, \vq) = K\tran \vq \in \real^{p \times k}$, and attention as scaled softmax over columns (spatial locations), that is, $h(S) \defn \vsigma_2(S / \sqrt{d})$:
\begin{align}
	\va = \vsigma_2 \left( K\tran \vq / \sqrt{d} \right) \in \real^p,
\label{eq:sp-att}
\end{align}
where $\vsigma_2(S) \defn \eta_2(\exp(S))$ and $\exp$ is taken elementwise.

\paragraph{Attention-weighted pooling}

As shown in \autoref{tab:land}, the average pooling operation ($f = f_{-1}$) is by far the most common. However, the more general function $f_\alpha$~\eq{alpha} has shown improved performance in instance-level tasks~\cite{gem}. For $\alpha < -1$ ($\gamma > 1$) in particular, it yields an intermediate operation between average and max-pooling. The latter is clearly beneficial when feature maps are sparse, because it better preserves the non-zero elements.

We adopt $f = f_\alpha$ for its genericity: the only operation that is not included as a special case in \autoref{tab:land} is log-sum-exp~\cite{lse}. This choice assumes $X \ge 0$. This is common in networks ending in $\relu$, like ResNet~\cite{resnet}, which is also what makes feature maps sparse. However, vision transformers and modern convolutional networks like ConvNeXt~\cite{convnext} do not end in $\relu$; hence $X$ has negative elements and is not necessarily sparse. We therefore define
\begin{align}
	V = \phi_V(X) = X - \min X \in \real^{d \times p},
\label{eq:sp-val}
\end{align}
where the minimum is taken over all elements of $X$, such that $f_\alpha$ operates only on non-negative numbers.

We also define $\vu = \phi_U(\vz) = \vz$ and the output dimension is $d' = d$. Thus, the mappings $\phi_V, \phi_U$ are parameter-free. The argument is that, for average pooling for example ($f = f_{-1}$ in~\eq{pool}), any linear layers before or after pooling would commute with pooling, thus they would form part of the encoder rather than the pooling process. Moreover, \autoref{tab:land} shows that $\phi_U$ is non-identity only for iterative methods.

In summary, we define \Ours (\our) as
\begin{align}
	\vu = \pi_\our(X) \defn f_\alpha^{-1}(f_\alpha(V) \va) \in \real^d,
\label{eq:sp}
\end{align}
where $V \in \real^{d \times p}$ is the value~\eq{sp-val} and $\va \in \real^p$ is the attention map~\eq{sp-att}. Parameter $\alpha$ is learned in GeM~\cite{gem}, but we find that treating it as a hyperparameter better controls the quality of the attention maps.

\section{Experiments}
\label{sec:exp}

\subsection{Datasets, networks and evaluation protocols}
\label{sec:setup}

%------------------------------------------------------------------------------

\paragraph{Supervised pre-training}

We train ResNet-18, ResNet-50~\cite{resnet}, ConvNeXt-S~\cite{convnext}, ViT-S and ViT-B~\cite{vit} for \emph{image classification} on ImageNet-1k. For the analysis \autoref{sec:analysis} and ablation \autoref{sec:ablation}, we train ResNet-18 on the first 20\% of training examples per class of ImageNet-1k~\cite{imagenet} (called ImageNet-20\%) for 100 epochs. For the benchmark of \autoref{sec:bench}, we train ResNet-50 for 100 and 200 epochs, ConvNeXt-S and ViT-S for 100 and 300 epochs and ViT-B for 100 epochs, all on the 100\% of ImageNet-1k. We evaluate on the full validation set in all cases and measure top-1 classification accuracy. The baseline is the default per network, \ie GAP for convolutional networks and \cls token for transformers.

\paragraph{Self-supervised pre-training}

On the 100\% of ImageNet-1k, we train DINO~\cite{dino} with ResNet-50, ConvNeXt-S and ViT-S for 100 epochs. We evaluate on the validation set by $k$-NN and \emph{linear probing} on the training set. For \emph{linear probing}, we train a linear classifier on top of features as in DINO~\cite{dino}. For $k$-NN~\cite{instancediscrimination}, we freeze the model and extract features, then use a $k$-nearest neighbor classifier with $k = 10$.

\paragraph{Downstream tasks}

We fine-tune supervised and self-supervised ViT-S on CIFAR-10~\cite{cifar}, CIFAR-100~\cite{cifar} and Oxford Flowers~\cite{oxford_flowers} for \emph{image classification}, measuring top-1 classification accuracy.
We perform \emph{object localization} without fine-tuning using supervised and self-supervised ViT-S on CUB~\cite{cub} and ImageNet-1k, measuring MaxBoxAccV2~\cite{wsol}. We perform \emph{object discovery} without fine-tuning using self-supervised ViT-S with DINO-SEG~\cite{dino} and LOST~\cite{lost} on VOC07~\cite{voc}, VOC12~\cite{voc} and COCO~\cite{coco}, measuring CorLoc~\cite{corloc}.
We validate \emph{robustness} against background changes using ViT-S on ImageNet-9~\cite{xiao2021noise} and its variations. We use the linear head and linear probe for supervised and self-supervised ViT-S, respectively, measuring top-1 classification accuracy.

In the appendix, we provide implementation details, more benchmarks, ablations and visualizations.

%For HOW we use a kernel of size 3 and we do not perform dimensionality reduction. For LSE we initialize $r=10$ . For GeM we use a kernel of size 7 and initialize $gamma=2$.
\subsection{Experimental Analysis}
\label{sec:analysis}

% %------------------------------------------------------------------------------
% \begin{figure}
% \caption{\iavr{VISUALIZATION}} 
% \label{fig:vis-maps}
% \end{figure}
% %------------------------------------------------------------------------------

%------------------------------------------------------------------------------
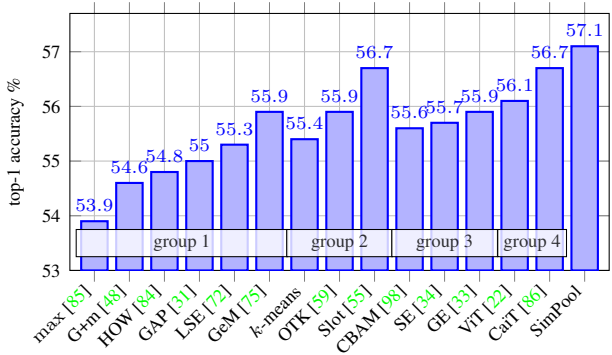
\begin{figure}
\centering
\begin{tikzpicture}[
	group/.style={draw,fill=white,fill opacity=.8,inner sep=2pt},
]
\begin{axis}[
	font=\scriptsize,
	width=1.05\linewidth,
	height=0.6\linewidth,
	ylabel={top-1 accuracy \%},
	ybar,
	ymin=53,
	ymax=57.7,
	enlarge x limits=0.05,
	x tick label style={rotate=45,anchor=east},
	xtick={0, 1, 2, 3, 4, 5, 6, 7, 8, 9, 10, 11, 12, 13, 14},
	xticklabels={$\max$~\cite{mac},G+m~\cite{gapmax},HOW~\cite{how},GAP~\cite{resnet},LSE~\cite{lse}, GeM~\cite{gem}, $k$-means, OTK~\cite{otk},Slot~\cite{slot},CBAM~\cite{cbam},SE~\cite{se},GE~\cite{gatherexcite},ViT~\cite{vit},CaiT~\cite{cait},\ours},
	nodes near coords,
	nodes near coords align=vertical,
	]
	
	\addplot coordinates {
		(0, 53.9) 
		(1, 54.6)
		(2, 54.8) 
		(3, 55.0)
		(4, 55.3)
        (5, 55.9)
  	(6, 55.4) 
		(7, 55.9) 
		(8, 56.7) 
		(9, 55.6) 
		(10, 55.7) 
		(11, 55.9) 
		(12, 56.1) 
		(13, 56.7)
		(14, 57.1)
		%(GeM\cite{gem},54.2)
		%(LSE\cite{lse},55.3)
		%(HOW\cite{how},54.8)
		%(OTK\cite{otk},55.8)
		%(Slot\cite{slot},56.6)
		%(SE\cite{se},55.7)
		%(CBAM\cite{cbam},55.6)
		%(GE\cite{gatherexcite},55.9)
		%(ViT\cite{vit},56.1)
		%(CaiT\cite{cait},56.6)
		%(\ours,57.1)
	};
	\node[group,minimum width=80] at (axis cs:2.5, 53.5) {group 1};
	\node[group,minimum width=40] at (axis cs:7.0, 53.5) {group 2};
	\node[group,minimum width=40] at (axis cs:10,  53.5) {group 3};
	\node[group,minimum width=26] at (axis cs:12.5,53.5) {group 4};
\end{axis}
\end{tikzpicture}
\caption{\emph{Image classification} on ImageNet-20. Supervised training of ResNet-18 for 100 epochs.}
\label{fig:tourn}
\end{figure}
%------------------------------------------------------------------------------

\autoref{fig:tourn} evaluates different methods in groups following \autoref{tab:land}, regardless of their original design for (a) pooling or not, (b) different tasks, \eg instance-level or category-level, (c) different networks, \eg convolutional or transformers.

\emph{Group 1} consists of simple pooling methods with: (a) no parameters: GAP~\cite{nin}, max~\cite{mac}, GAP+$\max$~\cite{gapmax}; and (b) scalar parameter: GeM~\cite{gem} and LSE~\cite{lse}. HOW~\cite{how} is the only method to use (parameter-free) attention. GeM is performing the best, with LSE following second. These methods are inferior to those in other groups. 

\emph{Group 2} incorporates methods with $k > 1$ vectors. We set $k=3$ and take the maximum of the $3$ logits per class. OTK and Slot use attention. Slot attention~\cite{slot} works best, outperforming $k$-means by 1.3\%.

\emph{Group 3} refers to parametric attention-based methods, weighting features based on their importance for the task: CBAM~\cite{cbam}, Squeeze-Excitation~\cite{se} and Gather-Excite~\cite{gatherexcite}. While originally designed as components within the architecture, we adapt them to pooling by GAP at the end. Gather-Excite~\cite{gatherexcite} performs best.

\emph{Group 4} refers to parametric attention-based methods found in vision transformers. ViT~\cite{vit} refers to multi-head self-attention learnable \cls and four heads, which we incorporate as a single layer at the end of the model. CaiT~\cite{cait} is the same but using only cross-attention between \cls and patch embeddings. CaiT performs the best. 

\Ours outperforms all other methods. Seeing this experiment as a tournament, we select the best performing method of each group and qualify it for the benchmark of \autoref{sec:bench}.

\subsection{Benchmark}
\label{sec:bench}

\paragraph{Image Classification}

%------------------------------------------------------------------------------
\begin{table}
\small
\centering
\setlength{\tabcolsep}{3pt}
\begin{tabular}{lccccccccc}
\toprule
\Th{Method} & \Th{Ep} & \mc{2}{\Th{ResNet-50}} & \mc{2}{\Th{ConvNext-S}} & \mc{2}{\Th{ViT-S}} & \mc{2}{\Th{ViT-B}} \\
%\cmidrule{3-8}
%& & \mc{2}{\Th{Accuracy}} & \mc{2}{\Th{Accuracy}} & \mc{2}{\Th{Accuracy}} \\
\midrule
Baseline                 & 100 & \mc{2}{77.4}   & \mc{2}{81.1}   & \mc{2}{72.7} & \mc{2}{74.1} \\
CaiT \cite{cait}         & 100 & \mc{2}{77.3}   & \mc{2}{81.2}   & \mc{2}{72.6} & \mc{2}{-} \\
Slot \cite{slot}         & 100 & \mc{2}{77.3}   & \mc{2}{80.9} & \mc{2}{72.9} & \mc{2}{-} \\
GE \cite{gatherexcite}   & 100 & \mc{2}{77.6}   & \mc{2}{81.3}   & \mc{2}{72.6} & \mc{2}{-} \\
\rowcolor{LightCyan}
\Ours                    & 100 &  \mc{2}{\bf{78.0}}   & \mc{2}{\bf{81.7}}   & \mc{2}{\bf{74.3}} & \mc{2}{\bf{75.1}} \\
\midrule
Baseline                 & 300 & \mc{2}{78.1$^{\dag}$} &  \mc{2}{83.1}   &  \mc{2}{77.9} & \mc{2}{-} \\
\rowcolor{LightCyan}
\Ours                    & 300 & \mc{2}{\bf{78.7$^{\dag}$}} &  \mc{2}{\bf{83.5}}  &  \mc{2}{\bf{78.7}} & \mc{2}{-} \\
%\midrule
%& & \Th{$k$-NN} &\Th{Prob} & \Th{$k$-NN} &\Th{Prob} & \Th{$k$-NN} &\Th{Prob} \\
%\midrule
%Baseline                 & 100 & 61.8 & 61.1 & 59.3 & 63.9 & 68.9 & 71.5 \\
%\rowcolor{LightCyan}
%\Ours                    & 100 & \bf{63.7}  & \bf{61.9} & \bf{68.8} & \bf{72.2} & \bf{69.8} & ~\bf{72.8} \\
%\midrule
%Baseline                 & 300 & - & - & - & - & 72.2 & - \\
%\rowcolor{LightCyan}
%\Ours                    & 300 & \bf{-}  & \bf{-} & \bf{-} & \bf{-} & \bf{72.6} & \bf{-} \\
\bottomrule
\end{tabular}
\vspace{3pt}
\caption{\emph{Image classification} top-1 accuracy (\%) on ImageNet-1k. Supervised pre-training for 100 and 300 epochs. Best competitors selected per group from \autoref{fig:tourn}. Baseline: GAP for convolutional, \cls for transformers; \Th{ep}: epochs; $^{\dag}$: 200 epochs.}
\label{tab:Super100percImageNet}
\end{table}
%------------------------------------------------------------------------------

\autoref{tab:Super100percImageNet} compares \Ours with baseline and tournament winners per group of \autoref{sec:analysis} on supervised pre-training for classification. For 100 epochs, \Ours outperforms all methods, consistently improving the baseline by 0.6\% using convolutional networks, 1.6\% using ViT-S and 1.0\% using ViT-B. Gather-Excite~\cite{gatherexcite} improves over the baseline only on convolutional networks, while Slot~\cite{slot} only on ViT-S. CaiT improves over the baseline only for ConvNeXt-S. By contrast, \Ours improves everywhere. For more than 100 epochs, \Ours improves the baseline by 0.5\% using ResNet-50, 0.4\% using ConvNeXt-S and 0.8\% using ViT-S.

%------------------------------------------------------------------------------
\begin{table}
\small
\centering
\setlength{\tabcolsep}{3pt}
\begin{tabular}{lccccccc}
\toprule
\mr{2}{\Th{Method}} &  \mr{2}{\Th{Ep}} & \mc{2}{\Th{ResNet-50}} & \mc{2}{\Th{ConvNeXt-S}} & \mc{2}{\Th{ViT-S}} \\ \cmidrule{3-8}
& & \Th{$k$-NN} &\Th{Prob} & \Th{$k$-NN} &\Th{Prob} & \Th{$k$-NN} &\Th{Prob} \\
\midrule
Baseline                 & 100 & 61.8 & 63.0 & 65.1 & 68.2 & 68.9 & 71.5 \\
\rowcolor{LightCyan}
\Ours                    & 100 & \bf{63.8}  & \bf{64.4} & \bf{68.8} & \bf{72.2} & \bf{69.8} & \bf{72.8} \\
\bottomrule
\end{tabular}
\vspace{3pt}
\caption{\emph{Image classification} top-1 accuracy (\%) on ImageNet-1k. Self-supervised pre-training with DINO~\cite{dino} for 100 epochs. Baseline: GAP for convolutional, \cls for transformers.}
\label{tab:Self100percImageNet}
\end{table}
%------------------------------------------------------------------------------

\autoref{tab:Self100percImageNet} evaluates self-supervised pre-training for 100 epochs. \Ours improves over the baseline by 2.0\% $k$-NN and 1.4\% linear probing on ResNet-50; 3.7\% $k$-NN and 4.0\% linear probing on ConvNeXt-S; and 0.9\% $k$-NN and 1.3\% linear probing on ViT-S.

% resnet-18 -> 11.2
% resnet-50 -> 25.6
% vit-t -> 6.6
% vit-s -> 22.1
% convnext-s -> 50.0
% convnext-t -> 29.0

%------------------------------------------------------------------------------

%------------------------------------------------------------------------------
\begin{table}
\small
\centering
\setlength{\tabcolsep}{2.5pt}
\begin{tabular}{lcccccc} \toprule
\mr{2}{\Th{Method}} & \mc{3}{\Th{Supervised}} & \mc{3}{\Th{Self-Supervised}} \\ \cmidrule{2-7}
& \Th{CF-10} & \Th{CF-100} & \Th{FL} & \Th{CF-10} & \Th{CF-100} & \Th{FL} \\ \midrule
Baseline & 98.1 & 86.0 & 97.1 & 98.7 & 89.8 & 98.3 \\
\rowcolor{LightCyan}
\Ours & \bf{98.4} & \bf{86.2} & \bf{97.4} &\bf{98.9} & \bf{89.9} & \bf{98.4} \\
\bottomrule
\end{tabular}
\vspace{3pt}
\caption{\emph{Image classification} accuracy (\%), fine-tuning for classification for 1000 epochs. ViT-S pre-trained on ImageNet-1k for 100 epochs. Self-supervision with DINO~\cite{dino}. \Th{CF-10}: CIFAR-10~\cite{cifar}, \Th{CF-100}: CIFAR-100~\cite{cifar}, \Th{FL}: Flowers\cite{oxford_flowers}.}
\label{tab:finetuning}
\end{table}
%------------------------------------------------------------------------------

\paragraph{Fine-tuning for classification}

\autoref{tab:finetuning} evaluates fine-tuning for classification on different datasets of a supervised and a self-supervised ViT-S. \Ours brings small improvement over the baseline in all cases.

%------------------------------------------------------------------------------
\begin{table}
\small
\centering
\setlength{\tabcolsep}{3pt}
\begin{tabular}{lcccc} \toprule
\mr{2}{\Th{Method}} & \mc{2}{\Th{Supervised}} & \mc{2}{\Th{Self-Supervised}} \\ \cmidrule{2-5}
& \Th{CUB} & \Th{ImageNet} & \Th{CUB} & \Th{ImageNet} \\ \midrule
Baseline & 63.1 & 53.6 & 82.7 & 62.0   \\
\rowcolor{LightCyan}
\ours & \bf{77.9} & \bf{64.4} & \bf{86.1} & \bf{66.1}  \\
\midrule
Baseline@20 & 62.4 & 50.5 & 65.5 & 52.5 \\
\rowcolor{LightCyan}
\ours@20 & \bf{74.0} & \bf{62.6} & \bf{72.5} & \bf{58.7} \\
\bottomrule
\end{tabular}
\vspace{3pt}
\caption{\emph{Localization accuracy} MaxBoxAccV2 on CUB test and ImageNet-1k validation set. ViT-S pre-trained on ImageNet-1k for 100 epochs. Self-supervision with DINO~\cite{dino}. @20: at epoch 20.}
\label{tab:maxboxacc}
\end{table}
%------------------------------------------------------------------------------

\paragraph{Object localization}

Accurate localization can have a significant impact on classification accuracy, particularly under multiple objects, complex scenes and background clutter. \autoref{tab:maxboxacc} evaluates localization accuracy under both supervision settings. \Ours significantly improves the baseline by up to 7\% MaxBoxAccV2 when self-supervised and up to 14\% when supervised. In the latter case, the gain is already up to 12\% at epoch 20.

%------------------------------------------------------------------------------
\begin{table}
\small
\centering
\setlength{\tabcolsep}{1pt}
\begin{tabular}{lccc|ccc} \toprule
\mr{2}{\Th{Method}} & \mc{3}{\Th{DINO-seg}~\cite{lost, dino}} & \mc{3}{\Th{LOST}~\cite{lost}} \\ \cmidrule{2-7}
& \Th{VOC07} & \Th{VOC12} & \Th{COCO} & \Th{VOC07} & \Th{VOC12} & \Th{COCO} \\ \midrule
Baseline & 30.8 & 31.0 & 36.7 & 55.5 & 59.4 & 46.6  \\
\rowcolor{LightCyan}
\ours & \bf{53.2} & \bf{56.2} & \bf{43.4} & \bf{59.8} & \bf{65.0} & \bf{49.4} \\
\midrule
Baseline@20 & 14.9 & 14.8 & 19.9 & 50.7 & 56.6 & 40.9 \\
\rowcolor{LightCyan}
\ours@20 & \bf{49.2} & \bf{54.8} & \bf{37.9} & \bf{53.9} & \bf{58.8} & \bf{46.1} \\
\bottomrule
\end{tabular}
\vspace{3pt}
\caption{\emph{Object discovery} CorLoc. ViT-S pre-trained on ImageNet-1k for 100 epochs. Self-supervision with DINO~\cite{dino}. @20: at epoch 20.}
\label{tab:corloc_lost}
\end{table}
%------------------------------------------------------------------------------

\paragraph{Unsupervised object discovery}

\autoref{tab:corloc_lost} studies LOST \cite{lost}, which uses the raw features of a vision transformer pre-trained using DINO~\cite{dino} for unsupervised single-object discovery, as well as the baseline DINO-seg~\cite{lost, dino}, which uses the attention maps instead. \Ours significantly outperforms the baseline on all datasets by up to 25.2\% CorLoc for DINO-seg and 5.6\% for LOST on VOC12. Again, the gain is significant already at the first 20 epochs.

%------------------------------------------------------------------------------
\begin{table}
\small
\centering
\setlength{\tabcolsep}{3.5pt}
\begin{tabular}{lcccccccc} \toprule
\Th{Method} &  OF & MS & MR & MN & NF & OBB & OBT & IN-9 \\ \midrule
\mc{9}{\Th{Supervised}} \\ \midrule
Baseline &  66.4 & 79.1 & 67.4 & 65.5 & 37.2 & 12.9 & 15.2 & 92.0 \\
\rowcolor{LightCyan}
\Ours &  \bf{71.8} & \bf{80.2} & \bf{69.3} & \bf{67.3} & \bf{42.8} & \bf{15.2} & \bf{15.6} & \bf{92.9} \\\midrule
\mc{9}{\Th{Self-supervised + Linear probing}} \\ \midrule
Baseline & \bf{87.3} & 87.9 & 78.5 & 76.7 & 47.9 & \bf{20.0} & \bf{16.9} & 95.3 \\
\rowcolor{LightCyan}
\Ours &  \bf{87.3} & \bf{88.1} & \bf{80.6} & \bf{78.7} & \bf{48.2} & 17.8 & 16.7 & \bf{95.6} \\\bottomrule
\end{tabular}
\vspace{3pt}
\caption{\emph{Background robustness} on IN-9~\cite{xiao2021noise} and its variations; more details in the appendix. ViT-S pre-trained on ImageNet-1k for 100 epochs. Self-supervision with DINO~\cite{dino}.}
\label{tab:BGChalsimpool}
\end{table}
%------------------------------------------------------------------------------

\paragraph{Background changes}

We evaluate robustness to the background changes using IN-9~\cite{xiao2021noise} dataset. \autoref{tab:BGChalsimpool} shows that \Ours improves over the baseline under both supervision settings with only 2 out of 8 exceptions under DINO~\cite{dino} pre-training. The latter is justified, given that none of the foreground objects or masks are present in these settings.

%------------------------------------------------------------------------------
\begin{table}[h]
\small
\centering
\setlength{\tabcolsep}{2pt}
\begin{tabular}{lcc|cc|cc|cc}
\toprule
& \mc{2}{\Th{ResNet-18}}  & \mc{2}{\Th{ResNet-50}} & \mc{2}{\Th{ConvNeXt-S}} & \mc{2}{\Th{ViT-S}} \\
& \Th{\#par} & \Th{flo} & \Th{\#par} & \Th{flo} & \Th{\#par} & \Th{flo} & \Th{\#par} & \Th{flo} \\
\midrule
Baseline   & 11.7 & 1.82 & 25.6 & 4.13 & 50.2 & 8.68 & 22.1 & 4.24 \\
CaiT    & 18.0 & 1.85 & 75.9 & 4.60 & 57.3 & 8.75 & 23.8 & 4.29 \\
Slot    & 14.6 & 1.87 & 71.7 & 4.89 & 56.7 & 8.79 & 23.7 & 4.30 \\
GE   & 11.7 & 1.83 & 26.1 & 4.15 & 50.3 & 8.69 & 22.1 & 4.25 \\
\rowcolor{LightCyan}
SimPool & 12.2 & 1.84 & 33.9 & 4.34 & 51.4 & 8.71 & 22.3 & 4.26 \\
\bottomrule
\end{tabular}
\vspace{3pt}
\caption{\emph{Computation resources} on Imagenet-1k, with $d=512$ (ResNet-18), $2048$ (ResNet-50), $768$ (ConvNeXt), $384$ (ViT-S). \Th{\#par}: number of parameters, in millions; \Th{flo}: GFLOPS.}
\label{tab:ParamsSuper100percImageNet}
\end{table}
%------------------------------------------------------------------------------

\paragraph{Computation resources} 

\autoref{tab:ParamsSuper100percImageNet} shows the number of parameters and floating point operations per second for the best competitors of \autoref{fig:tourn}. Resources depend on the embedding dimension $d$. \Ours is higher than the baseline but not the highest.

%------------------------------------------------------------------------------
\begin{table}[t]
\small
\centering
\setlength{\tabcolsep}{3pt}
\begin{tabular}{lccccc} \toprule
\Th{Network}          & \Th{Pooling}  & \Th{Depth} &\Th{Init} & \Th{Accuracy} & \Th{\#params} \\ \midrule
\textsc{Base}         & GAP           & 12 & 12 & 73.3  & 22.1M \\ \midrule
\textsc{Base}         & \mr{6}{\cls}  & 12 & 0 & 72.7  & 22.1M \\
$\textsc{Base} + 1$   &               & 13 & 0 & 73.2  & 23.8M \\
$\textsc{Base} + 2$   &               & 14 & 0 & 73.7  & 25.6M \\
$\textsc{Base} + 3$   &               & 15 & 0 & 73.8  & 27.4M \\
$\textsc{Base} + 4$   &               & 16 & 0 & 73.9  & 29.2M \\
\rowcolor{LightCyan}
$\textsc{Base} + 5$   &               & 17 & 0 & \bf{74.6}  & 30.9M \\ \midrule
\rowcolor{LightCyan}
\textsc{Base}         & \mr{4}{\ours} & 12 & 12 & \bf{74.3}  & 22.3M\\
$\textsc{Base} - 1$   &               & 11 & 11 & 73.9  & 20.6M\\
$\textsc{Base} - 2$   &               & 10 & 10 & 73.6  & 18.7M\\
$\textsc{Base} - 3$   &               & 9 & 9 & 72.5  & 17.0M\\
\bottomrule
\end{tabular}
\vspace{3pt}
\caption{\emph{Trade-off between performance and parameters}. Supervised pre-training of ViT-S on \imagenet for 100 epochs. \Th{Init}: Initial layer of pooling token. \textsc{Base}: original network. \textsc{Base}$+b$ (\textsc{Base}$-b$): $b$ blocks added to (removed from) the network.}
\label{tab:param}
\end{table}
%------------------------------------------------------------------------------

\paragraph{Performance \vs parameters}

\autoref{tab:param} aims to answer the question of how much the performance improvement of \Ours is due to parameters of the query and key mappings. Interestingly, ViT-S works better with GAP than the default \cls. \Ours adds 0.2M parameters to the network. For fair comparison, we remove blocks from the network (\Th{Base}) when using \Ours and add blocks when using $\cls$. We find that, to exceed the accuracy of \Th{Base} \Ours, \Th{Base} $\cls$ needs 5 extra blocks, \ie, 9M more parameters. Equally interestingly, removing 3 blocks from \Th{Base} \Ours is only slightly worse than \Th{Base} \cls, having 5M fewer parameters.
\subsection{Ablation study}
\label{sec:ablation}

\begin{table}
\small
\centering
\setlength{\tabcolsep}{1.5pt}
\begin{tabular}{lccc|cccccccc} 
% \mc{4}{(a)} & \mc{7}{(b)}  \\ 
\toprule
\Th{Ablation} & \Th{Options} & \Th{Acc} & \hspace{6pt} & \hspace{6pt} & \mc{3}{\Th{Linear}} &  \mc{3}{\Th{LN}} & \Th{Acc} \\ \midrule
\mr{2}{$h(S)$} & $\vsigma_2(S_i/\sqrt{d})_{i=1}^m$ & 56.6 & & & $Q$    & $K$        & $V$   & $Q$    & $K$ & $V$   &    \\ \cmidrule{5-12}
 & $\eta_2(\vsigma_1(S/\sqrt{d}))$ & 55.6 & & &\checkmark & \checkmark & \checkmark & \checkmark & \checkmark & \checkmark & \bf{57.0}  \\\midrule
\mr{2}{\Th{Layers}}   & 3 & 56.8 & & & \win{\checkmark} & \win{\checkmark} &  & \checkmark & \checkmark & \checkmark & \bf{56.6}  \\
  & 5 & 55.9 & & & \checkmark &  &  & \checkmark & \checkmark & \checkmark & 56.5  \\\cmidrule{1-3}
 \mr{2}{\Th{Iter}}     & 3 & 56.5 & & & & \checkmark &  & \checkmark & \checkmark & \checkmark & 56.4  \\
   & 5 & 56.4 & & & &  &  & \checkmark & \checkmark & \checkmark & 55.6  \\ \midrule
\mr{2}{$U^0$} & $\gp{U}$ & 56.3 & & & \checkmark & \checkmark &  & \checkmark & \checkmark &  & 56.3  \\
 & $\diag(X\tran X)$ & 56.6 & & & \checkmark & \checkmark &  & \checkmark &  &  & 56.0  \\\cmidrule{1-3}
\mr{2}{$s(\vx, \vy)$} & $-\|\vx-\vy\|^2$ & 56.5 & & & \checkmark & \checkmark &  &  & \checkmark &  & 56.2  \\
  & cosine & 56.3 & & \us{} & \us{\checkmark} & \us{\checkmark} & \us{} & \us{} & \us{\win{\checkmark}} & \us{\win{\checkmark}} & \us{\bf{56.6}}  \\ \cmidrule{1-3}
\mr{2}{$\hp{k}$ ($\max$)} & 2 & 56.5 & & & \checkmark & \checkmark &  &  &  & \checkmark & 56.4  \\
 & 5 & 56.4 & & & \checkmark & \checkmark &  & \checkmark &  & \checkmark & 56.2 \\ \cmidrule{5-12}
\mr{2}{$\hp{k}$ (concat)}  & 2 & 56.5 & & & &  &  &  &  \win{\checkmark} &  \win{\checkmark} & 56.2  \\
  & 5 & 55.9 & & & \win{\checkmark} & \win{\checkmark} &  &  &  &  & 54.4  \\ \cmidrule{1-3}
$\phi_Q$, $\phi_K$ & $\gp{W_Q}=\gp{W_K}$ & 56.4 & & & &  &  &  &  &  & 54.5  \\ \midrule
 \us{\ours} & \us{} & \us{\bf{57.1}} & & & \mc{3}{\Th{GAP}}  & & & & 55.0  \\ \bottomrule
\end{tabular}
\vspace{3pt}
\caption{\Ours ablation on ImageNet-20\% using ResNet-18 trained for 100 epochs. Ablation of (left) design; (right) linear and LayerNorm (LN)~\cite{layernorm} layers. $q, k, v$: query, key, value. $\vsigma_2(S_i/\sqrt{d})_{i=1}^m$: same as our default, but with multi-head attenion, $m=4$ heads; $\hp{k}$ ($\max$): maximum taken over output logits; $\hp{k}$ (concat): concatenation and projection to the same output dimensions $d'$. \lrn{Green}: learnable parameter; \win{blue}: winning choice per group of experiments; \hl[LightCyan]{Cyan}: Our chosen default. Using pooling operation $f = f_\alpha$~\eq{alpha} (left); $f = f_{-1}$ (right).}
\label{tab:ablationall}
\end{table}

We ablate the design and components of \Ours. More ablations are found in the appendix. In particular, for function $f_\alpha$~\eq{alpha}, we set $\gamma=2$ for convolutional networks and $\gamma=1.25$ for transformers by default, where $\gamma = (1-\alpha)/2$ is a hyperparameter.

\paragraph{Design}

In \autoref{tab:ablationall} (left), we ablate (a) the attention function $h$~\eq{attn}; (b) the number of iterations with shared parameters at every iteration (\Th{Layers}) or not (\Th{Iter}); (c) the initialization $U^0$; (d) the pairwise similarity function $s$; (e) the number $\hp{k}$ of pooled vectors, obtained by $k$-means instead of GAP. We also consider queries and keys sharing the same mapping, $\gp{W_Q}=\gp{W_K}$. We observe that multi-head, few iterations and initialization by $\diag(X\tran X)$ perform slightly worse, without adding any extra parameters, while setting $\gp{W_Q}=\gp{W_K}$ performs slightly worse, having 50\% less parameters.

% with maximum taken over output logits or concatenation and projection to the same output dimensions $d'$

\paragraph{Linear and LayerNorm layers}

In \autoref{tab:ablationall} (right), we systematically ablate linear and LayerNorm (LN)~\cite{layernorm} layers on query $q$, key $k$ and value $v$. We strive for performance and quality while at the same time having a small number of components and parameters. In this sense, we choose the setup that includes linear layers on $q, k$ and LN on $k, v$, yielding 56.6 accuracy. We observe that having linear and LN layers everywhere performs best under classification accuracy. However, this setup 
% yields 52.56 MaxBoxAccV2 compared with 53.76 of our chosen default, 
has attention maps of lower quality and more parameters.

\section{Conclusion}
\label{sec:conclusion}

We have introduced \Ours, a simple, attention-based pooling mechanism that acts at the very last step of either convolutional or transformer encoders, delivering highly superior quantitative results on several benchmarks and downstream tasks. In addition, \Ours delivers decent attention maps in both convolutional and transformer networks under both supervision and self-supervision with remarkable improvement in delineating object boundaries for supervised transformers. Despite this progress, we believe that investigating why the standard \cls-based attention fails under supervision deserves further study.

\paragraph{Acknowledgements}

This work was supported by the Hellenic Foundation for Research and Innovation (HFRI) under the BiCUBES project (grant: 03943). It was also supported by the RAMONES and iToBos EU Horizon 2020 projects, under grants 101017808 and 965221, respectively. NTUA thanks NVIDIA for the support with the donation of GPU hardware.

%We thank Shashanka Venkataramanan for his valuable contribution to certain experiments. This work was supported by computational time granted from GRNET in the Greek HPC facility ARIS under projects PR009017, PR011004 and PR012047 and by the HPC resources of GENCI-IDRIS in France under the 2021 grant AD011012884. NTUA thanks NVIDIA for the support with the donation of GPU hardware. This work has been supported by RAMONES and iToBos projects, funded by the EU Horizon 2020 research and innovation programme, under grants 101017808 and 965221, respectively.

%------------------------------------------------------------------------------
\clearpage
%------------------------------------------------------------------------------
%%%%%%%%% REFERENCES
{\small
\bibliographystyle{ieee_fullname}
\bibliography{egbib}
}
\clearpage
%------------------------------------------------------------------------------
%%%%%%%%% APPENDIX
%\title{Keep it \Ours: Who said supervised transformers suffer from attention deficit? \\ {\emph{Supplementary material}}}
%\renewcommand{\teaser}{}
%------------------------------------------------------------------------------
%\maketitle
%------------------------------------------------------------------------------
%\setcounter{page}{1}
%\iccvrulercount=1

% Start of appendix
\appendix

% Reset counters
\setcounter{equation}{0}
\setcounter{figure}{0}
\setcounter{table}{0}
\setcounter{section}{0}

% Adjust numbering format for equations, tables, and figures
\renewcommand{\thesection}{\Alph{section}}
\renewcommand{\theequation}{\thesection\arabic{equation}}
\renewcommand{\thetable}{\thesection\arabic{table}}
\renewcommand{\thefigure}{\thesection\arabic{figure}}

\section{Extended related work}
\label{sec:more-related}

Spatial pooling of visual input is the process by which spatial resolution is reduced to $1 \times 1$, such that the input is mapped to a single vector. This process can be gradual and interleaved with mapping to a feature space, because any feature space is amenable to smoothing or downsampling. The objective is robustness to deformation while preserving important visual information.

Via a similarity function, \eg dot product, the vector representation of an image can be used for efficient matching to class representations for category-level tasks or to the representation of another image for instance-level tasks. One may obtain more than one vectors per image as a representation, but this requires a particular kernel for matching.

\paragraph{Background}

The study of receptive fields in neuroscience~\cite{receptive} lead to the development of 2D \emph{Gabor filters}~\cite{gabor} as a model of the first processing layer in the visual cortex. Visual descriptors based on filter banks in the frequency domain~\cite{gist} and orientation histograms~\cite{sift, hog} can be seen as efficient implementations of the same idea. Apart from mapping to a new space---that of filter responses or orientation bins---they involve a form of smoothing, at least in some orientation, and weighted local spatial pooling.

\emph{Textons}~\cite{texton1} can be seen as a second layer, originally studied in the context of texture discrimination~\cite{texture} and segmentation~\cite{texton1, texton2} and taking the form of multidimensional histograms on Gabor filter responses. The \emph{bag of words} model~\cite{bow-r, bow-c} is based on the same idea, as a histogram on other visual descriptors. Again, apart from mapping to a new space---that of textons or visual words---they involve local or global spatial pooling.

\emph{Histograms} and every step of building visual features can be seen as a form of nonlinear coding followed by pooling~\cite{code-pool}. \emph{Coding} is maybe the most important factor. For example, a high-dimensional mapping before pooling, optionally followed by dimension reduction after pooling, can reduce interference between elements~\cite{fisher, vlad, emk}. \emph{Weighting} of individual elements is also important in attending important regions~\cite{spp, recept-learn, auto-pool} and in preventing certain elements from dominating others~\cite{burst, gmp, demo}.

The \emph{pooling operation} itself is any symmetric (permutation-invariant) set function, which can be expressed in the form $F(X) = g\left(\sum_{x \in X} f(x)\right)$~\cite{deep-sets}. The most common is average and maximum~\cite{hmax, theor, code-pool}.

Common ways to obtain a representation of \emph{multiple vectors} are using a spatial partition~\cite{spp} or a partition in the feature space~\cite{he, asmk}.

%------------------------------------------------------------------------------

\paragraph{Convolutional networks}

Following findings of neuroscience, early convolutional networks~\cite{neo, lenet} are based on learnable \emph{convolutional layers} interleaved with fixed \emph{spatial pooling layers} that downsample, which is an instance of the coding-pooling framework. The same design remains until today~\cite{alexnet, vgg, resnet, convnext}. Again, apart from mapping to a new space, convolutional layers involve a form of weighted local pooling. Again, the operation in pooling layers is commonly average~\cite{lenet} or maximum~\cite{hmax, alexnet}.

Early networks end in a fully-connected layer over a feature tensor of low resolution~\cite{lenet, alexnet, vgg}. This evolved into spatial pooling, \eg \emph{global average pooling} (GAP) for classification~\cite{nin, resnet}, regional pooling for detection~\cite{frcnn}, or global maximum followed by a pairwise loss~\cite{mac} for instance-level tasks. This is beneficial for downstream tasks and interpretability~\cite{cam}.

The spatial pooling operation at the end of the network is widely studied in instance level-tasks~\cite{spoc, mac, gem}, giving rise to forms of \emph{spatial attention}~\cite{crow, delf, delg, how, solar}, In category-level tasks, it is more common to study \emph{feature re-weighting} as components of the architecture~\cite{se, cbam, gatherexcite}. The two are closely related because \eg the weighted average is element-wise weighting followed by sum. Most modern pooling operations are learnable.

% \cite{bilinear, compact-bilinear}
% GAP is proposed in \cite{theor}, gap+max in \cite{gapmax} and an early attention pooling in \cite{gatedatt} -> deep sets.
% BlurPool \cite{blurpool}, i-sqrt cov\cite{isqrtcov}, PSWE \cite{pswe}, Kernel Pooling\cite{kernelpooling}, WELDON \cite{weldon}.

Pooling can be \emph{spatial}~\cite{gatherexcite, delf, delg, how, solar}, \emph{over channels}~\cite{se}, or both \cite{crow, cbam}. CBAM~\cite{cbam} is particularly related to our work in the sense that it includes global average pooling followed by a form of spatial attention, although the latter is not evident in its original formulation and although CBAM is designed as a feature re-weighting rather than pooling mechanism.

One may obtain a representation of \emph{multiple vectors} \eg by some form of clustering~\cite{diffuse} or optimal transport~\cite{otk}.

%------------------------------------------------------------------------------

\paragraph{Vision transformers}

Pairwise interactions between features are forms of \emph{self-attention} that can be seen as alternatives to convolution or forms of pooling. They have commonly been designed as architectural components of convolutional networks, again over the spatial~\cite{nln, aa_net, san, n3_net} or the channel dimensions~\cite{a2_net, eca}. Originating in language models~\cite{transformer}, \emph{vision transformers}~\cite{vit} streamlined these approaches and became the dominant competitors of convolutional networks.

Transformers commonly downsample only at the input, forming spatial \emph{patch tokens}. Pooling is based on a learnable \cls (``classification'') token, which, beginning at the input space, undergoes the same self-attention operation with patch tokens and eventually provides a global image representation. That is, the network ends in global weighted average pooling, using as weights the attention of \cls over the patch tokens. Pooling is still gradual, since \cls interacts with patch tokens throughout the network depth.

Several variants of transformers often bring back ideas from convolutional networks, including spatial hierarchy~\cite{swin}, relative position encoding~\cite{rethinking, levit}, re-introducing convolution~\cite{cvt, convit}, re-introducing pooling layers~\cite{pit, swin, pvt, pvtv2}, or simple pooling instead of attention~\cite{metaformer}. In this sense, downsampling may occur inside the transformer, \eg for classification~\cite{pit, swin} or detection~\cite{pvt, pvtv2}.

Few works that have studied anything other than \cls for pooling in transformers are mostly limited to GAP~\cite{swin, rest, halonet, cnnvsvit}. \cls offers attention maps for free, but those are typically of low quality unless in a self-supervised setting~\cite{dino}, which is not well studied. Few works that attempt to rectify this in the supervised setting include a spatial entropy loss~\cite{spatialentropy}, shape distillation from convolutional networks~\cite{intriguing} and skipping computation of self-attention, observing that the quality of self-attention is still good at intermediate layers~\cite{skip}. It has also been found beneficial to inject the \cls token only at the last few layers~\cite{cait}.

We are thus motivated to question why the pooling operation at the end of the network needs to be different in convolutional networks and vision transformers and why pooling with a \cls token needs to be performed across the network depth. We study pooling in both kinds of networks, in supervised and self-supervised settings alike. We derive a simple, attention-based, universal pooling mechanism that applies equally to all cases, improving both performance and the quality of attention maps.

\section{More on the method}
\label{sec:more-method}

In \autoref{sec:a-frame}, we summarize the generalized pooling framework of \autoref{sec:frame}. We then detail how to formulate methods studied in \autoref{sec:land} as instances of our pooling framework so as to obtain \autoref{tab:land}, examining them in groups as in \autoref{sec:land}. Finally, we summarize \Ours in \autoref{sec:a-group5}.

\paragraph{Notation}

By $\id$ we denote the identity mapping. Given $n \in \nat$, we define $[n] \defn \{1, \dots, n\}$. By $\ind_A$ we denote the indicator function of set $A$, by $\delta_{ij}$ the Kronecker delta and by $[P]$ the Iverson bracket of statement $P$. By $A \circ B$ we denote the Hadamard product of matrices $A, B$ and by $A^{\circ n}$ the Hadamard $n$-th power of $A$. We recall that by $\eta_1, \eta_2$ we denote the row-wise and column-wise $\ell_1$-normalization of a matrix, respectively, while $\vsigma_2$ is column-wise softmax.

%------------------------------------------------------------------------------
\begin{algorithm}
\DontPrintSemicolon
\SetFuncSty{textsc}
\SetDataSty{emph}
\newcommand{\commentsty}[1]{{\color{Purple}#1}}
\SetCommentSty{commentsty}
\SetKwComment{Comment}{$\triangleright$ }{}
\SetKwInOut{Input}{input}
\SetKwInOut{Option}{option}
\SetKwInOut{Output}{output}
\SetKwFunction{Init}{init}

\Input{ $p$: \#patches, $d$: dimension}
\Input{ $X \in \real^{d \times p}$: features}
\Option{ $k$: \#pooled vectors}
\Option{ $\Init$: pooling initialization}
\Option{ $T$: \#iterations}
\Option{ $\{\phi_Q^t\}, \{\phi_K^t\}$: query, key mappings}
\Option{ $s$: pairwise similarity function}
\Option{ $h$: attention function}
\Option{ $\{\phi_V^t\}$: value mapping}
\Option{ $f$: pooling function}
\Option{ $\{\phi_X^t\}, \{\phi_U^t\}$: output mappings}
\Output{ $d'$: output dimension}
\Output{ $U \in \real^{d' \times k}$: pooled vectors}
\BlankLine

$d^0 ~\gets d$ \Comment*{input dimension}
$X^0 \gets X \in \real^{d^0 \times k}$ \Comment*{initialize features}
$U^0 \gets \Init(X) \in \real^{d^0 \times k}$ \Comment*{initialize pooling}
\For{$t = 0, \dots, T-1$}
{
	$Q \gets \phi_Q^t(U^t) \in \real^{n^t \times k}$ \Comment*{query~\eq{query}}
	$K \gets \phi_K^t(X^t) \in \real^{n^t \times p}$ \Comment*{key~\eq{key}}
	$S \gets \vzero_{p \times k}$ \Comment*{pairwise similarity}
	\For{$i \in [p]$, $j \in [k]$}{
		$s_{ij} \gets s(\vk_{\all i}, \vq_{\all j})$ \label{alg:frame-sim}
	}
	$A \gets h(S) \in \real^{p \times k}$ \Comment*{attention~\eq{attn}}
	$V \gets \phi_V^t(X^t) \in \real^{n^t \times p}$ \Comment*{value~\eq{value}}
	$Z \gets f^{-1}(f(V) A) \in \real^{n^t \times k}$ \Comment*{pooling~\eq{pool}}
	$X^{t+1} \gets \phi_X^t(X^t) \in \real^{d^{t+1} \times p}$ \Comment*{update feat.~\eq{feat-next}}
	$U^{t+1} \gets \phi_U^t(Z)   \in \real^{d^{t+1} \times k}$ \Comment*{update pool.~\eq{pool-next}}
}
$d' \gets d^T$ \Comment*{output dimension}
$U \gets U^T$ \Comment*{pooled vectors}

\caption{Our generalized pooling framework.}
\label{alg:frame}
\end{algorithm}
%------------------------------------------------------------------------------

\subsection{Pooling framework summary}
\label{sec:a-frame}

Our generalized pooling framework is summarized in \autoref{alg:frame}. As \emph{input}, it takes the features $X \in \real^{d \times p}$, representing $p$ patch embeddings of dimension $d$. As \emph{output}, it returns the pooled vectors $U \in \real^{d' \times k}$, that is, $k$ vectors of dimension $d'$. As \emph{options}, it takes the number $k$ of vectors to pool; the pooling initialization function \textsc{init}; the number $T$ of iterations; the query and key mappings $\{\phi_Q^t\}, \{\phi_K^t\}$; the pairwise similarity function $s$; the attention function $h$; the value mapping $\{\phi_V^t\}$; the pooling function $f$; and the output mappings $\{\phi_X^t\}, \{\phi_U^t\}$.

The mappings and dimensions within iterations may be different at each iteration, and all optional functions may be learnable. As such, the algorithm is general enough to incorporate any deep neural network. However, the focus is on pooling, as is evident by the pairwise similarity between queries (pooled vectors) and keys (features) in line~\ref{alg:frame-sim}, which is a form of \emph{cross-attention}.

%------------------------------------------------------------------------------
\newtcolorbox{conclude}[1][]{%
	breakable,
	top=6pt,
	bottom=6pt,
	left=6pt,
	right=6pt,
	boxrule=.6pt,
	sharp corners,
	colframe=black,
% 	colback=white,
  #1
}
%------------------------------------------------------------------------------

\subsection{Group 1: Simple methods with $k = 1$}
\label{sec:a-group1}

These methods are non-iterative, there are no query $Q$, key $K$, similarity matrix $S$ or function $h$, and the attention is a vector $\va \in \real^p$ that is either fixed or a function directly of $X$. With the exception of HOW~\cite{how}, the value matrix is $V = X$, that is, $\phi_V = \id$, and we are pooling into vector $\vu = \vz \in \real^d$, that is, $\phi_U = \id$. Then,~\eq{pool} takes the form
\begin{align}
	\vu = f^{-1}(f(X) \va) \in \real^d,
\label{eq:a-pool}
\end{align}
and we focus on instantiating it to identify function $f$ and attention vector $\va$. With the exception of LSE~\cite{lse}, function $f$ is $f_\alpha$~\eq{alpha} and we seek to identify $\alpha$.

%------------------------------------------------------------------------------

\paragraph{Global average pooling (GAP)~\cite{nin, resnet}}

According to~\eq{gap},
\begin{align}
	\pi_A(X) \defn \frac{1}{p} \sum_{j=1}^p \vx_{\all j} =
		X \vone_p / p = f_{-1}^{-1}(f_{-1}(X) \va),
\label{eq:a-gap}
\end{align}
where $f_{-1}(x) = x^{\frac{1-(-1)}{2}} = x$, thus $f_{-1} = \id$, and $\va = \vone_p / p$.

%------------------------------------------------------------------------------

\paragraph{Max pooling~\cite{mac}}

Assuming $X \ge 0$,
\begin{align}
	\pi_{\max}(X) &\defn \max_{j \in [p]} \vx_{\all j}
		= \lim_{\gamma \to \infty}
			\left( \sum_{j=1}^p \vx_{\all j}^\gamma \right)^\frac{1}{\gamma}
				\label{eq:a-max-1} \\
		&= \lim_{\gamma \to \infty} \left( X^\gamma \vone_p \right)^\frac{1}{\gamma}
		= f_{-\infty}^{-1}(f_{-\infty}(X) \va), \label{eq:a-max-2}
\end{align}
where all operations are taken element-wise and $\va = \vone_p$.

%------------------------------------------------------------------------------

\paragraph{Generalized mean (GeM)~\cite{gem}}

Assuming $X \ge 0$,
\begin{align}
	\pi_{\gem}(X) &\defn
		\left( \frac{1}{p} \sum_{j=1}^p \vx_{\all j}^\gamma \right)^\frac{1}{\gamma}
			\label{eq:a-gem-1} \\
		&= \left( X^\gamma \vone_p / p \right)^\frac{1}{\gamma}
		= f_\alpha^{-1}(f_\alpha(X) \va), \label{eq:a-gem-2}
\end{align}
where all operations are taken element-wise, $\gamma = (1 - \alpha) / 2$ is a learnable parameter and $\va = \vone_p / p$.

\emph{\Ours has the same pooling function but is based on an attention mechanism.}

%------------------------------------------------------------------------------

\paragraph{Log-sum-exp (LSE)~\cite{lse}}

\begin{align}
	\pi_{\lse}(X) &\defn
		\frac{1}{r} \log \left( \frac{1}{p} \sum_{j=1}^p \exp(r \vx_{\all j}) \right)
			\label{eq:a-lse-1} \\
		&= f^{-1}(f(X) \va), \label{eq:a-lse-2}
\end{align}
where all operations are taken element-wise, $r$ is a learnable scale parameter, $f(x) = e^{rx}$ and $\va = \vone_p / p$.

%------------------------------------------------------------------------------

\paragraph{HOW~\cite{how}}

The attention value of each feature $\vx_{\all j}$ is its norm $\|\vx_{\all j}\|$. That is,
\begin{align}
	\va &= (\|\vx_{\all 1}\|, \dots, \|\vx_{\all p}\|)\tran
		= (X^{\circ 2})\tran \vone_d     \label{eq:a-how-att-1} \\
		&= \diag(X\tran X) \in \real^p,  \label{eq:a-how-att-2}
\end{align}
obtained by pooling over channels. The value matrix is
\begin{align}
	V = \phi_V(X) = \fc(\avg_3(X)) \in \real^{d' \times p},
\label{eq:a-how-val}
\end{align}
where $\avg_3$ is $3 \times 3$ local average pooling, $\fc$ is a fixed fully-connected ($1 \times 1$ convolutional) layer incorporating centering, PCA dimension reduction and whitening according to the statistics of the local features of the training set and $d' < d$ is the output dimension. Then,
\begin{align}
	\vz &= \sum_{j=1}^p a_j \vv_{\all j} = V \va
		= f_{-1}^{-1}(f_{-1}(V) \va) \in \real^{d'},
\label{eq:a-how}
\end{align}
where $f_{-1} = \id$ as in GAP. Finally, the output is $\vu = \eta^2(\vz)$, where the mapping $\phi_U = \eta^2$ is $\ell_2$-normalization.

%------------------------------------------------------------------------------

\subsection{Group 2: Iterative methods with $k > 1$}
\label{sec:a-group2}

We examine three methods, which, given $X \in \real^{d \times p}$ and $k < p$, seek $U \in \real^{d \times k}$ by iteratively optimizing a kind of assignment between columns of $X$ and $U$. The latter are called references~\cite{otk}, centroids~\cite{lloyd}, or slots~\cite{slot}. Assignment can be soft~\cite{otk, slot} or hard~\cite{lloyd}. It can be an assignment of columns of $X$ to columns of $U$~\cite{lloyd, slot} or both ways~\cite{otk}. The algorithm may contain learnable components~\cite{otk, slot} or not~\cite{lloyd}.

%------------------------------------------------------------------------------

\paragraph{Optimal transport kernel embedding (OTK)~\cite{otk}}

Pooling is based on a learnable parameter $U \in \real^{d \times k}$. We define the $p \times k$ \emph{cost} matrix $C = (c_{ij})$ consisting of the pairwise squared Euclidean distances between columns of $X$ and $U$, \ie, $c_{ij} = \norm{\vx_{\all i} - \vu_{\all j}}^2$. We seek a $p \times k$ non-negative \emph{transportation plan} matrix $P \in \cP$ representing a joint probability distribution over features of $X$ and $U$ with uniform marginals:
\begin{align}
	\cP \defn \{ P \in \real_+^{p \times k} :
		P \vone_k = \vone_p/p, P\tran \vone_p = \vone_k/k \}.
\label{eq:a-plan}
\end{align}
The objective is to minimize the expected, under $P$, pairwise cost with entropic regularization
\begin{align}
	P^* \defn \arg\min_{P \in \cP} \inner{P, C} - \epsilon H(P),
\label{eq:a-ot}
\end{align}
where $H(P) = -\vone_p\tran (P \circ \log P) \vone_k$ is the entropy of $P$, $\inner{\cdot,\cdot}$ is the Frobenius inner product and $\epsilon > 0$ controls the sparsity of $P$. The optimal solution is $P^* = \sink(e^{-C / \epsilon})$, where exponentiation is element-wise and $\sink$ is the Sinkhorn-Knopp algorithm~\cite{sinkhorn}, which iteratively $\ell_1$-normalizes rows and columns of a matrix until convergence~\cite{cuturi}. Finally, pooling is defined as
\begin{align}
	U = \psi(X) P^* \in \real^{d' \times k},
\label{eq:a-otk}
\end{align}
where $\psi(X) \in \real^{d' \times p}$ and $\psi: \real^d \to \real^{d'}$ is a Nystr\"om approximation of a kernel embedding in $\real^d$, \eg a Gaussian kernel~\cite{otk}, which applies column-wise to $X \in \real^{d \times p}$.

\begin{conclude}
	We conclude that OTK~\cite{otk} is a instance of our pooling framework with learnable $U_0 = U \in \real^{d \times k}$, query/key mappings $\phi_Q = \phi_K = \id$, pairwise similarity function $s(\vx, \vy) = -\norm{\vx - \vy}^2$, attention matrix $A = h(S) = \sink(e^{S / \epsilon}) \in \real^{p \times k}$, value mapping $\phi_V = \psi$, average pooling function $f = f_{-1}$ and output mapping $\phi_U = \id$.
\end{conclude}

Although OTK is not formally iterative in our framework, $\sink$ internally iterates indeed to find a soft-assignment between the features of $X$ and $U$.

%------------------------------------------------------------------------------

\paragraph{$k$-means~\cite{lloyd}}

$k$-means aims to find a $d \times k$ matrix $U$ minimizing the sum of squared Euclidean distances of each column $\vx_{\all i}$ of $X$ to its nearest column $\vu_{\all j}$ of $U$:
\begin{align}
	J(U) \defn \sum_{i=1}^p \min_{j \in [k]} \norm{\vx_{\all i} - \vu_{\all j}}^2.
\label{eq:a-cost}
\end{align}
Observe that~\eq{cost} is the special case $k = 1$, where the unique minimum $\vu^* = \pi_A(X)$ is found in closed form~\eq{opt}. For $k > 1$, the distortion measure $J$ is non-convex and we are only looking for a local minimum.

The standard $k$-means algorithm is initialized by a $d \times k$ matrix $U^0$ whose columns are $k$ of the columns of $X$ sampled at random and represent a set of $k$ \emph{centroids} in $\real^d$. Given $U^t$ at iteration $t$, we define the $p \times k$ \emph{distance} matrix $D = (d_{ij})$ consisting of the pairwise squared Euclidean distances between columns of $X$ and $U^t$, \ie, $d_{ij} = \norm{\vx_{\all i} - \vu^t_{\all j}}^2$. For $i \in [p]$, feature $\vx_{\all i}$ is \emph{assigned} to the nearest centroid $\vu^t_{\all j}$ with index
\begin{align}
	c_i = \arg\min_{j \in [k]} d_{ij},
\label{eq:a-mean-assign}
\end{align}
where ties are resolved to the lowest index. Then, at iteration $t+1$, centroid $\vu^t_{\all j}$ is \emph{updated} as the mean of features $\vx_{\all i}$ assigned to it, \ie, for which $c_i = j$:
\begin{align}
	\vu^{t+1}_{\all j} =
		\frac{1}{\sum_{i=1}^p \delta_{c_i j}}
		\sum_{i=1}^p \delta_{c_i j} \vx_{\all i}.
\label{eq:a-mean-update}
\end{align}

Let $\arg\min_1(D)$ be the $p \times k$ matrix $M = (m_{ij})$ with
\begin{align}
	m_{ij} = \delta_{c_i j} =
		\left[ j = \textstyle{\arg\min_{j' \in [k]}} d_{ij'} \right].
\label{eq:a-arg-min}
\end{align}
That is, each row $\vd_i \in \real^k$ of $D$ yields a row $\vm_i \in \{0,1\}^k$ of $M$ that is an one-hot vector indicating the minimal element over $\vd_i$. Define operator $\arg\max_1$ accordingly. Then,~\eq{a-mean-update} can be written in matrix form as
\begin{align}
	U^{t+1} = X \eta_2(\textstyle{\arg\max_1}(-D)) \in \real^{d \times k}.
\label{eq:a-mean}
\end{align}

\begin{conclude}
	We conclude that $k$-means is an iterative instance of our pooling framework with the columns of $U^0 \in \real^{d \times k}$ sampled at random from the columns of $X$, query/key mappings $\phi_Q = \phi_K = \id$, pairwise similarity function $s(\vx, \vy) = -\norm{\vx - \vy}^2$, attention matrix $A = h(S) = \eta_2(\arg\max_1(S)) \in \real^{p \times k}$, value mapping $\phi_V = \id$, average pooling function $f = f_{-1}$ and output mappings $\phi_X = \phi_U = \id$.
\end{conclude}

%------------------------------------------------------------------------------

\paragraph{Slot attention~\cite{slot}}

Pooling is initialized by a random $d' \times k$ matrix $U^0$ sampled from a normal distribution $\normal(\mu, \sigma^2)$ with shared, learnable mean $\mu \in \real^{d'}$ and standard deviation $\sigma \in \real^{d'}$. Given $U^t$ at iteration $t$, define the query $Q = W_Q \LN(U^t) \in \real^{n \times k}$ and key $K = W_K \LN(X) \in \real^{n \times p}$, where $\LN$ is LayerNorm~\cite{layernorm} and $n$ is a common dimension. An attention matrix is defined as
\begin{align}
	A = \eta_1(\vsigma_2(K\tran Q / \sqrt{n})) \in \real^{p \times k}.
\label{eq:a-slot-att}
\end{align}
Then, with value $V = W_V \LN(X) \in \real^{n \times p}$, pooling is defined as the weighted average
\begin{align}
	Z = V A \in \real^{n \times k}.
\label{eq:a-slot-pool}
\end{align}
Finally, $U^t$ is updated according to
\begin{align}
	G &= \gru(Z) \in \real^{d' \times k}  \label{eq:a-slot-out-1} \\
	U^{t+1} &= G + \mlp(\LN(G)) \in \real^{d' \times k},  \label{eq:a-slot-out-2}
\end{align}
where $\gru$ is a \emph{gated recurrent unit}~\cite{gru} and $\mlp$ a multi-layer perceptron with ReLU activation and a residual connection~\cite{slot}.

We now simplify the above formulation by removing LayerNorm and residual connections.

\begin{conclude}
	We conclude that slot attention~\cite{slot} is an iterative instance of our pooling framework with $U^0$ a random $d' \times k$ matrix sampled from $\normal(\mu, \sigma^2)$ with learnable parameters $\mu, \sigma \in \real^{d'}$, query mapping $\phi_Q(U) = W_Q U \in \real^{n \times k}$, key mapping $\phi_K(X) = W_K X \in \real^{n \times p}$, pairwise similarity function $s(\vx, \vy) = \vx\tran \vy$, attention matrix $A = h(S) = \eta_1(\vsigma_2(S / \sqrt{n})) \in \real^{p \times k}$, value mapping $\phi_V(X) = W_V X \in \real^{n \times p}$, average pooling function $f = f_{-1}$, output mapping $\phi_U(Z) = \mlp(\gru(Z)) \in \real^{d' \times k}$ and output dimension $d'$.
\end{conclude}

\emph{\Ours is similar in its attention mechanism, but is non-iterative with $k=1$ and initialized by GAP.}

%------------------------------------------------------------------------------

\subsection{Group 3: Feature re-weighting, $k = 1$}
\label{sec:a-group3}

We examine two methods, originally proposed as components of the architecture, which use attention mechanisms to re-weight features in the channel or the spatial dimension. We modify them by placing at the end of the network, followed by GAP. We thus reveal that they serve as attention-based pooling. This includes pairwise interaction, although this was not evident in their original formulation.

\paragraph{Squeeze-and-excitation block (SE)~\cite{se}}

The \emph{squeeze} operation aims to mitigate the limited receptive field of convolutional networks, especially in the lower layers. It uses global average pooling over the spatial dimension,
\begin{align}
	\vu^0 = \pi_A(X) \in \real^d.
\label{eq:a-se-init}
\end{align}

Then, the \emph{excitation} operation aims at capturing channel-wise dependencies and involves two steps. In the first step, a learnable gating mechanism forms a vector
\begin{align}
	\vq = \sigma(\mlp(\vu^0)) \in \real^d,
\label{eq:a-se-query}
\end{align}
where $\sigma$ is the sigmoid function and $\mlp$ concists of two linear layers with ReLU activation in-between and forming a bottlenect of hidden dimension $d/r$. This vector expresses an importance of each channel that is not mutually exclusive. The second step re-scales each channel (row) of $X$ by the corresponding element of $\vq$,
\begin{align}
	V = \diag(\vq) X \in \real^{d \times p}.
\label{eq:se-value}
\end{align}
The output $X' = V \in \real^{d \times p}$ is a new tensor of the same shape as $X$, which can be used in the next layer. In this sense, the entire process is considered a block to be used within the architecture of convolutional networks at several layers. This yields a new family of networks, called \emph{squeeze-and-excitation networks} (SENet).

However, we can also see it as a pooling process if we perform it at the end of a network, followed by GAP:
\begin{align}
	\vz = \pi_A(V) = \diag(\vq) X \vone_p/p \in \real^d,
\label{eq:se-pool}
\end{align}

\begin{conclude}
	We conclude that this modified SE block is a non-iterative instance of our pooling framework with $\vu^0 = \pi_A(X) \in \real^d$, query mapping $\phi_Q(\vu) = \sigma(\mlp(\vu)) \in \real^d$, no key $K$, similarity matrix $S$ of function $h$, uniform spatial attention $\va = \vone_p/p$, value mapping $\phi_V(X) = \diag(\vq) X \in \real^{d \times p}$ and average pooling function $f = f_{-1}$.
\end{conclude}

The original design does not use $\va$ or $\vz$; instead, it has an output mapping $\phi_X(X) = V = \diag(\vq) X \in \real^{d \times p}$. Thus, it can be used iteratively along with other mappings of $X$ to form a modified network architecture.

%------------------------------------------------------------------------------

\paragraph{Convolutional block attention module (CBAM)~\cite{cbam}}

This is an extension of SE~\cite{se} that acts on both the channel and spatial dimension in similar ways. \emph{Channel attention} is similar to SE: It involves (a) global average and maximum pooling of $X$ over the spatial dimension,
\begin{align}
	U^0 = (\pi_A(X) \ \ \pi_{\max}(X)) \in \real^{d \times 2};
\label{eq:a-cbam-init}
\end{align}
(b) a learnable gating mechanism forming vector
\begin{align}
	\vq = \sigma(\mlp(U^0) \vone_2/2) \in \real^d,
\label{eq:a-cbam-query}
\end{align}
which is defined as in SE~\cite{se} but includes averaging over the two columns before $\sigma$; and (c) re-scaling channels (rows) of $X$ by $\vq$,
\begin{align}
	V = \diag(\vq) X \in \real^{d \times p}.
\label{eq:a-cbam-value}
\end{align}

\emph{Spatial attention} performs a similar operation in the spatial dimension: (a) global average and maximum pooling of $V$ over the channel dimension,
\begin{align}
	S = (\pi_A(V\tran) \ \ \pi_{\max}(V\tran)) \in \real^{p \times 2};
\label{eq:a-cbam-sim}
\end{align}
(b) a learnable gating mechanism forming vector
\begin{align}
	\va = \sigma(\conv_7(S)) \in \real^p,
\label{eq:a-cbam-att}
\end{align}
where $\conv_7$ is a a convolutional layer with kernel size $7 \times 7$; and (c) re-scaling features (columns) of $V$ by $\va$,
\begin{align}
	X' = V \diag(\va) \in \real^{d \times p}.
\label{eq:a-cbam-out}
\end{align}

The output $X'$ is a new tensor of the same shape as $X$, which can be used in the next layer. In this sense, CBAM is a block to be used within the architecture, like SE~\cite{se}. However, we can also see it as a \emph{pooling process} if we perform it at the end of a network, followed by GAP:
\begin{align}
	\vz = \pi_A(X') = V \diag(\va) \vone_p / p  = V \va \iavr{/ p} \in \real^d.
\label{eq:a-cbam-pool}
\end{align}
We also \emph{simplify} CBAM by removing max-pooling from both attention mechanisms and keeping average pooling only. Then,~\eq{a-cbam-sim} takes the form
\begin{align}
	\Vs &= \pi_A(V\tran) = V\tran \vone_d / d = (\diag(\vq) X)\tran \vone_d / d
			\label{eq:a-cbam-pair-1} \\
		&= X\tran \vq \iavr{/ d} \in \real^p.
			\label{eq:a-cbam-pair-2}
\end{align}
This reveals \emph{pairwise interaction} by dot-product similarity between $\vq$ as query and $X$ as key. It was not evident in the original formulation, because dot product was split into element-wise product followed by sum.

\begin{conclude}
	We conclude that this modified CBAM module is a non-iterative instance of our pooling framework with $\vu^0 = \pi_A(X) \in \real^d$, query mapping $\phi_Q(\vu) = \sigma(\mlp(\vu)) / d \in \real^d$, key mapping $\phi_K = \id$, pairwise similarity function $s(\vx, \vy) = \vx\tran \vy$, spatial attention $\va = h(\Vs) = \sigma(\conv_7(\Vs)) / p \in \real^p$, value mapping $\phi_V(X) = \diag(\vq) X \in \real^{d \times p}$, average pooling function $f = f_{-1}$ and output mapping $\phi_U = \id$.
\end{conclude}

The original design does not use $\vz$; instead, it has an output mapping $\phi_X(X) = V \diag(\va) = \diag(\vq) X \diag(\va) \in \real^{d \times p}$. Thus, it can be used iteratively along with other mappings of $X$ to form a modified network architecture.

\emph{\Ours is similar in that $\vu^0 = \pi_A(X)$ but otherwise its attention mechanism is different: there is no channel attention while in spatial attention there are learnable query/key mappings and competition between spatial locations.}

%------------------------------------------------------------------------------

\subsection{Group 4: Transformers}
\label{sec:a-group4}

We re-formulate the standard ViT~\cite{vit} in two streams, where one performs pooling and the other feature mapping. We thus show that the pooling stream is an iterative instance of our framework, where iterations are blocks. We then examine the variant CaiT~\cite{cait}, which is closer to \Ours in that pooling takes place in the upper few layers with the features being fixed.

%------------------------------------------------------------------------------

\paragraph{Vision transformer (ViT)~\cite{vit}}

The transformer encoder \emph{tokenizes} the input image, \ie, it splits the image into $p$ non-overlapping \emph{patches} and maps them to patch token embeddings of dimension $d$ through a linear mapping. It then concatenates a learnable $\cls$ token embedding, also of dimension $d$, and adds a learnable \emph{position embedding} of dimension $d$ to all tokens. It is thus initialized as
\begin{align}
	F^0 &= (\vu^0 \ \ X^0) \in \real^{d \times (p+1)},
\label{eq:a-vit-init}
\end{align}
where $\vu^0 \in \real^d$ is the initial $\cls$ token embedding and $X^0 \in \real^{d \times p}$ contains the initial patch embeddings.

The encoder contains a sequence of \emph{blocks}. Given token embeddings $F^t = (\vu^t \ \ X^t) \in \real^{d \times (p+1)}$ as input, a block performs the following operations:
\begin{align}
	G^t &= F^t + \msa(\LN(F^t)) \in \real^{d \times (p+1)}  \label{eq:a-vit-block-1} \\
	F^{t+1} &= G^t + \mlp(\LN(G^t)) \in \real^{d \times (p+1)},  \label{eq:a-vit-block-2}
\end{align}
where $\LN$ is LayerNorm~\cite{layernorm} and $\mlp$ is a network of two affine layers with a ReLU activation in-between, applied to all tokens independently. Finally, at the end of block $T - 1$, the image is pooled into vector $\vu = \LN(\vu^T)$.

Given $F^t \in \real^{d \times (p+1)}$, the \emph{multi-head self-attention} (\msa) operation uses three linear mappings to form the query $Q = W_Q F^t$, key $K = W_K F^t$ and value $V = W_V F^t$, all in $\real^{d \times (p+1)}$. It then splits each of the three into $m$ submatrices, each of size $d/m \times (p+1)$, where $m$ is the number of \emph{heads}.

Given a stacked matrix $A = (A_1; \dots; A_m) \in \real^{d \times n}$, where $A_i \in \real^{d/m \times n}$ for $i \in [m]$, we denote splitting as
\begin{align}
	\cA = g_m(A) = \{A_1, \dots, A_m\} \subset \real^{d/m \times n}.
\label{eq:a-vit-split}
\end{align}
Thus, with $\cQ = g_m(Q) = \{Q_i\}$, $\cK = g_m(K) = \{K_i\}$, $\cV = g_m(V) = \{V_i\}$, self-attention is defined as
\begin{align}
	A_i &= \vsigma_2 \left( K_i\tran Q_i / \sqrt{d'} \right)
		\in \real^{(p+1) \times (p+1)} \label{eq:a-vit-self-att} \\
	Z_i &= V_i A_i \in \real^{d' \times (p+1)}, \label{eq:a-vit-self-pool}
\end{align}
for $i \in [m]$, where $d' = d/m$. Finally, given $\cZ = \{Z_i\}$, submatrices are grouped back and an output linear mapping yields the output of \msa:
\begin{align}
	U = W_U g_m^{-1}(\cZ) \in \real^{d \times (p+1)}.
\label{eq:a-vit-group}
\end{align}

Here, we decompose the above formulation into two parallel streams. The first operates on the \cls token embedding $\vu^t \in \real^d$, initialized by learnable parameter $\vu^0 \in \real^d$ and iteratively performing pooling. The second operates on the patch embeddings $X^t \in \real^{d \times p}$, initialized by $X^0 \in \real^{d \times p}$ as obtained by tokenization and iteratively performing feature extraction. We focus on the first one.

Given $\vu^t \in \real^d$, $X^t \in \real^{d \times p}$ at iteration $t$, we form the query $\cQ = g_m(W_Q \LN(\vu^t))$, key $\cK = g_m(W_K \LN(X^t))$ and value $\cV = g_m(W_V \LN(X^t))$. \emph{Cross-attention} between $\cQ$ and $\cK, \cV$ follows for $i \in [m]$:
\begin{align}
	\va_i &= \vsigma_2 \left( K_i\tran \vq_i / \sqrt{d'} \right)
		\in \real^p \label{eq:a-vit-str-att} \\
	\vz_i &= V_i \va_i \in \real^{d'}. \label{eq:a-vit-str-pool}
\end{align}
Finally, denoting $\cZ = \{\vz_1, \dots, \vz_m\}$, the \cls token embedding at iteration $t+1$ is given by
\begin{align}
	\vg^t &= \vu^t + W_U g_m^{-1}(\cZ) \in \real^d  \label{eq:a-vit-str-block-1} \\
	\vu^{t+1} &= \vg^t + \mlp(\LN(\vg^t)) \in \real^d.  \label{eq:a-vit-str-block-2}
\end{align}

We now simplify the above formulation by removing LayerNorm and residual connections. We also remove the dependence of self-attention of patch embeddings on the \cls token.

\begin{conclude}
	We conclude that ViT~\cite{vit} is an iterative instance of our pooling framework with learnable $\vu^0 \in \real^d$, query mapping $\phi_Q(\vu) = g_m(W_Q \vu) \subset \real^{d'}$ with $d' = d/m$, key mapping $\phi_K(X) = g_m(W_K X) \subset \real^{d' \times p}$, pairwise similarity function $s(\vx, \vy) = \vx\tran \vy$, spatial attention $\cA = h(\cS) = \{\vsigma_2(\Vs_i / \sqrt{d'})\}_{i=1}^m \subset \real^p$, value mapping $\phi_V(X) = g_m(W_V X) \subset \real^{d' \times p}$, average pooling function $f = f_{-1}$ and output mappings $\phi_X(X) = \mlp(\msa(X)) \in \real^{d \times p}$ and $\phi_U(\cZ) = \mlp(W_U g_m^{-1}(\cZ)) \in \real^d$.
\end{conclude}

Although $k = 1$, splitting into $m$ submatrices and operating on them independently is the same as defining $m$ query vectors in $\real^d$ via the block-diagonal matrix
\begin{align}
	Q = \left( \begin{array}{ccc}
		\vq_1  & \dots  & \vzero \\
		\vdots & \ddots & \vdots \\
		\vzero & \dots  & \vq_m  \\
	\end{array} \right) \in \real^{d \times m}. \label{eq:a-vit-sub}
\end{align}
$Q$ interacts with $K$ by dot product, essentially operating in $m$ orthogonal subspaces. This gives rise to an attention matrix $A \in \real^{p \times m}$ containing $\va_i$~\eq{a-vit-str-att} as columns and a pooled matrix $Z \in \real^{d \times m}$ containing $\vz_i$~\eq{a-vit-str-pool} as columns.

Thus, the $m$ heads in multi-head attention bear similarities to the $k$ pooled vectors in our formulation. The fact that transformer blocks act as iterations strengthens our observation that methods with $k > 1$ are iterative. However, because of linear maps at every stage, there is no correspondence between heads across iterations.

%------------------------------------------------------------------------------

\paragraph{Class-attention in image transformers (CaiT)~\cite{cait}}

This work proposes two modifications in the architecture of ViT~\cite{vit}. The first is that the encoder consists of two stages. In stage one, patch embeddings are processed alone, without a \cls token. In stage two, a learnable \cls token is introduced that interacts with patch embeddings with cross-attention, while the patch embeddings remain fixed. The second modification is that it introduces two learnable diagonal matrices $\Lambda_G^t, \Lambda_X^t \in \real^{d \times d}$ at each iteration (block) $t$ and uses them to re-weight features along the channel dimension.

Thus, stage one is specified by a modification of \eq{a-vit-block-1},~\eq{a-vit-block-2} as follows:
\begin{align}
	G^t &= X^t + \Lambda_G^t \msa(\LN(X^t)) \in \real^{d \times p}
		\label{eq:a-cait-block-1} \\
	X^{t+1} &= G^t + \Lambda_X^t \mlp(\LN(G^t)) \in \real^{d \times p}.
		\label{eq:a-cait-block-2}
\end{align}
This is similar to~\cite{se, cbam}, only here the parameters are learnable rather than obtained by GAP. Similarly, stage two is specified by a modification of \eq{a-vit-str-att}-\eq{a-vit-str-block-2}. Typically, stage two consists only of a few (1-3) iterations.

\begin{conclude}
	We conclude that a simplified version of stage two of CaiT~\cite{cait} is an iterative instance of our pooling framework with the same options as ViT~\cite{vit} except for the output mapping $\phi_X = \id$.
\end{conclude}

\emph{\Ours is similar in that there are again two stages, but stage one is the entire encoder, while stage two is a single non-iterative cross-attention operation between features and their GAP, using function $f_\alpha$ for pooling.}

Slot attention~\cite{slot} is also similar to stage two of CaiT, performing few iterations of cross-attention between features and slots with $\phi_X = \id$, but with a single head, $k > 1$ and different mapping functions.

%------------------------------------------------------------------------------
\begin{algorithm}
\DontPrintSemicolon
\SetFuncSty{textsc}
\SetDataSty{emph}
\newcommand{\commentsty}[1]{{\color{Purple}#1}}
\SetCommentSty{commentsty}
\SetKwComment{Comment}{$\triangleright$ }{}
\SetKwInOut{Input}{input}
\SetKwInOut{Output}{output}

\Input{ $d$: dimension, $p$: patches}
\Input{ features $X \in \real^{d \times p}$}
\Output{ pooled vector $\vu \in \real^d$}
\BlankLine

$\vu^0 \gets X \vone_p / p \in \real^d$ \Comment*{initialization~\eq{sp-init}}
$X \gets \LN(X) \in \real^{d \times p}$ \Comment*{LayerNorm~\cite{layernorm}}
$\vq ~\gets \lrn{W_Q} \vu^0 \in \real^d$ \Comment*{query~\eq{sp-query}}
$K \gets \lrn{W_K} X \in \real^{d \times p}$ \Comment*{key~\eq{sp-key}}
$\va ~\gets \vsigma_2 (K\tran \vq / \sqrt{d}) \in \real^p$ \Comment*{attention~\eq{sp-att}}
$V \gets X - \min X \in \real^{d \times p}$ \Comment*{value~\eq{sp-val}}
$\vu ~\gets f_\alpha^{-1}(f_\alpha(V) \va) \in \real^d$ \Comment*{pooling~\eq{alpha},\eq{sp}}

\caption{\Ours. \lrn{Green}: learnable.}
\label{alg:simpool}
\end{algorithm}
%------------------------------------------------------------------------------

\subsection{\Ours}
\label{sec:a-group5}

\Ours is summarized in \autoref{alg:simpool}. We are given a \emph{feature matrix} $X \in \real^{d \times p}$, resulting from flattening of tensor $\vX \in \real^{d \times W \times H}$ into $p = W \times H$ patches. We form the initial representation $\vu^0 = \pi_A(X) \in \real^d$~\eq{sp-init} by \emph{global average pooling} (GAP), which is then mapped by $\lrn{W_Q} \in \real^{d \times d}$~\eq{sp-query} to form the \emph{query} vector $\vq \in \real^d$. After applying LayerNorm~\cite{layernorm}, $X' = \LN(X)$, we map $X'$ by $\lrn{W_K} \in \real^{d \times d}$~\eq{sp-key} to form the \emph{key} $K \in \real^{d \times p}$. Then, $\vq$ and $K$ interact to generate the attention map $\va \in \real^p$~\eq{sp-att}. Finally, the pooled representation $\vu \in \real^d$ is a generalized weighted average of the \emph{value} $V = X' - \min X' \in \real^{d \times p}$ with $\va$ determining the weights and scalar function $f_\alpha$~\eq{alpha} determining the pooling operation~\eq{sp}.

The addition to what presented in the paper is LayerNorm after obtaining $\vu^0$ and before $K, V$. That is,~\eq{sp-key} and~\eq{sp-val} are modified as
\begin{align}
	K &= \phi_K(X) = W_K \LN(X) \in \real^{d \times p}. \label{eq:sp-key-ln} \\
	V &= \phi_V(X) = \LN(X) - \min \LN(X) \in \real^{d \times p}. \label{eq:sp-val-ln}
\end{align}
As shown in \autoref{tab:ablationall}, it is our choice in terms of simplicity, performance, and attention map quality to apply LayerNorm to key and value and linear layers to query and key. The learnable parameters are $\lrn{W_Q}$ and $\lrn{W_K}$.

\begin{conclude}
	In summary, \Ours is a non-iterative instance of our pooling framework with $k = 1$, $\vu^0 = \pi_A(X) \in \real^d$, query mapping $\phi_Q(\vu) = W_Q \vu \in \real^d$, key mapping $\phi_K(X) = W_K \LN(X) \in \real^{d \times p}$, pairwise similarity function $s(\vx, \vy) = \vx\tran \vy$, spatial attention $\va = h(\Vs) = \vsigma_2(\Vs / \sqrt{d}) \in \real^p$, value mapping $\phi_V(X) = \LN(X) - \min \LN(X) \in \real^{d \times p}$, average pooling function $f = f_\alpha$ and output mapping $\phi_U = \id$.
\end{conclude}

\section{More experiments}
\label{sec:more-exp}

\subsection{More datasets, networks \& evaluation protocols}
\label{sec:more-data}

\paragraph{Downstream tasks}

For \emph{image classification}, we use CIFAR-10~\cite{cifar}, CIFAR-100~\cite{cifar} and Oxford Flowers~\cite{oxford_flowers}. CIFAR-10 consists of 60000 images in 10 classes, with 6000 iamges per class. CIFAR-100 is just like CIFAR-10, except it has 100 classes containing 600 images each. Oxford Flowers consists of 102 flower categories containing between 40 and 258 images each.

For \emph{semantic segmentation}, we fine-tune a linear layer of a self-supervised ViT-S on ADE20K~\cite{ade20k}, measuring mIoU, mAcc, and aAcc. The training set consists of 20k images and the validation set of 2k images in 150 classes.

For \emph{background changes}, we use the linear head and linear probe of a supervised and self-supervised ViT-S, respectively, measuring top-1 classification accuracy on \imagenet-9~\cite{xiao2021noise} (IN-9) dataset. IN-9 contains nine coarse-grained classes with seven variations of both background and foreground.

For \emph{image retrieval}, we extract features from a self-supervised ResNet-50 and ViT-S and evaluate them on $\cR$Oxford and $\cR$Paris~\cite{radenovic2018revisiting}, measuring mAP. These are the revisited Oxford~\cite{fast-spatial} and Paris~\cite{Philbin08} datasets, comprising 5,062 and 6,412 images collected from Flickr by searching for Oxford and Paris landmarks respectively.

For \emph{fine-grained classification}, we extract features from a supervised and self-supervised ResNet-50 and ViT-S and evaluate them on Caltech-UCSD Birds (CUB200) \cite{cub}, Stanford Cars (CARS196) \cite{cars}, In-Shop Clothing Retrieval (In-Shop) \cite{in_shop} and Stanford Online Products (SOP) \cite{sop}, measuring Revall@$k$. Dataset statistics are summarized in \autoref{tab:fine-grained-datasets}.

For \emph{unsupervised object discovery}, we use VOC07~\cite{voc07} trainval, VOC12~\cite{voc12} trainval and COCO 20K \cite{coco, coco20k}. The latter is a subset of COCO2014 trainval dataset~\cite{coco}, comprising 19,817 randomly selected images. VOC07 comprises 9,963 images depicting 24,640 annotated objects. VOC12 comprises 11,530 images depicting 27,450 annotated objects.

\paragraph{Ablation} 

For the ablation of \autoref{sec:more-ablation}, we train supervised ResNet-18 and ViT-T for \emph{image classification} on ImageNet-20\% and \imagenet respectively.

\begin{table}
\centering
\setlength{\tabcolsep}{3pt}
\small
\begin{tabular}{lcccc} \toprule
	\Th{Dataset}        & \Th{CUB200} & \Th{Cars196} & \Th{SOP} & \Th{In-Shop} \\ \midrule
	Objects             & birds                  & cars                     & furniture                 & clothes                         \\
	\# classes          & $200$                  & $196$                    & $22,634$                   & $7,982$                         \\
	\# train images  & $5,894$                & $8,092$                  & $60,026$                   & $26,356$                        \\
	\# test images   & $5,894$                & $8,093$                  & $60,027$                   & $26,356$                        \\ \bottomrule
\end{tabular}
\vspace{3pt}
\caption{\emph{Statistics and settings} for the four fine-grained classification datasets.}
\label{tab:fine-grained-datasets}
\end{table}

\subsection{Implementation details}
\label{sec:more-setup}

\paragraph{Analysis}

We train ResNet-18 on ImageNet-20\% for 100 epochs following the ResNet-50 recipe of~\cite{timm}, but with learning rate $0.1$. We train on 4 GPUs with a global batch size of $4 \times 128 = 512$, using SGD~\cite{sgd} with momentum. We incorporate pooling methods as a layer at the end of the model.

\emph{Group 1}. For HOW~\cite{how}, we use a kernel of size 3 and do not perform dimension reduction. For LSE~\cite{lse}, we initialize the scale as $r = 10$. For GeM~\cite{gem}, we use a kernel of size 7 and initialize the exponent as $p = 2$.

\emph{Group 2}. For $k$-means, OTK~\cite{otk} and slot attention~\cite{slot}, we set $k = 3$ vectors and take the maximum of the three logits per class. For convergence, we set tolerance $t = 0.01$ and iterations $T = 5$ for $k$-means. We set the iterations to $T = 3$ for OTK and slot attention.

\emph{Group 3}. For CBAM~\cite{cbam}, we use a kernel of size 7. For SE~\cite{se} and GE~\cite{gatherexcite}, we follow the implementation of~\cite{timm}.

\emph{Group 4}. For ViT~\cite{vit} and CaiT~\cite{cait} we use $m = 4$ heads. For CaiT we set the iterations to $T = 1$, as this performs best.

\paragraph{Benchmark}

For \emph{supervised} pre-training, we train ResNet-50 for 100 and 200 epochs, ConvNeXt-S and ViT-S for 100 and 300 epochs and ViT-B for 100 epochs on \imagenet. For ResNet-50 we follow~\cite{timm}, using SGD with momentum with learning rate $0.4$. We train on 8 GPUs with global batch size $8 \times 128 = 1024$. For ConvNeXt-S we follow~\cite{convnext}, using AdamW \cite{adamw} with learning rate $0.004$. We use 8 GPUs with an aggregation factor of 4 (backpropagating every 4 iterations), thus with global batch size $8 \times 4 \times 256 = 4096$. For ViT-S we follow~\cite{timm}, using AdamW with learning rate $5 \times 10^{-4}$. We train on 8 GPUs with global batch size $8 \times 74 = 592$. For the 300 epoch experiments, we follow the same setup as for 100.

For \emph{self-supervised} pre-training, we train ResNet-50, ConvNeXt-S and ViT-S with DINO~\cite{dino} on \imagenet for 100 and 300 epochs, following~\cite{dino} and using 6 local crops. For ResNet-50, we train on 8 GPUs with global batch size $8 \times 160 = 1280$. We use learning rate $0.3$, minimum learning rate $0.0048$, global crop scale $[0.14, 1.0]$ and local crop scale $[0.05, 0.14]$. For ConvNeXt-S, we train on 8 GPUs with global batch size $8 \times 60 = 480$. We use learning rate $0.001$, minimum learning rate $2 \times 10^{-6}$, global crop scale $[0.14, 1.0]$ and local crop scale $[0.05, 0.14]$. As far as we know, we are the first to integrate DINO into ConvNeXt-S. For ViT-S, we train on 8 GPUs with global batch size $8 \times 100 = 800$. We use LARS \cite{lars} with learning rate $5 \times 10^{-4}$, minimum learning rate of $1 \times 10^{-5}$, global crop scale $[0.25, 1.0]$ and local crop scale $[0.05, 0.25]$. For the 300 epoch experiments, we follow the same setup as for 100. For linear probing, we follow~\cite{dino}, using 4 GPUs with global batch size $4 \times 256 = 1024$.

\paragraph{Downstream tasks}

For \emph{image classification}, we fine-tune supervised and self-supervised ViT-S on CIFAR-10, CIFAR-100 and Oxford Flowers, following~\cite{ibot}. We use a learning of $7.5 \times 10^{-6}$. We train on 8 GPUS for 1000 epochs with a global batch size of $8 \times 96 = 768$.

For \emph{object localization}, we use the supervised and self-supervised ViT-S on CUB and \imagenet, without finetuning. We follow~\cite{wsol} and we use the MaxBoxAccV2 metric. For the baseline, we use the mean attention map over all heads of the \cls token to generate the bounding boxes. For \Ours, we use the attention map $\va$~\eq{sp-att}.

For \emph{unsupervised object discovery}, we use the self-supervised ViT-S on VOC07~\cite{voc07} trainval, VOC12~\cite{voc12} trainval and COCO 20K~\cite{coco, coco20k}, without finetuning. We adopt LOST~\cite{lost} and DINO-seg~\cite{lost, dino} to extract bounding boxes. For both methods, we follow the best default choices~\cite{lost}. LOST operates on features. We use the the \emph{keys} of the last self-attention layer for the baseline and the \emph{keys} $K$~\eq{sp-key} for \Ours. DINO-seg operates on attention maps. We use the attention map of the head that achieves the best results following~\cite{lost}, \ie head 4, for the baseline and the attention map $\va$~\eq{sp-att} for \Ours.

For \emph{semantic segmentation}, we use the self-supervised ViT-S on ADE20K~\cite{ade20k}. To evaluate the quality of the learned representation, we only fine-tune a linear layer on top of the fixed patch features, without multi-scale training or testing and with the same hyper-parameters as in iBOT~\cite{ibot}. We follow the setup of~\cite{swin}, \ie, we train for 160,000 iterations with $512 \times 512$ images. We use AdamW~\cite{adamw} optimizer with initial learning rate $3 \times 10^{-5}$, poly-scheduling and weight decay of 0.05. We train on 4 GPUS with a global batch size of $4 \times 4 = 16$.

For \emph{computation resources}, GFLOPS are calculated on the input size of $224 \times 224$, on a single NVIDIA A100 40GB GPU.

% Throughput is measured with a batch size of 128 on a single NVIDIA A100 40GB GPU, averaged over 100 iterations.
\subsection{More benchmarks}
\label{sec:more-bench}

%------------------------------------------------------------------------------
\begin{table}
\small
\centering
\setlength{\tabcolsep}{3pt}
\begin{tabular}{lccc}
\toprule
\mr{2}{\Th{Method}} &  \mr{2}{\Th{Epochs}} & \mc{2}{\Th{ViT-S}} \\ \cmidrule{3-4}
& & \Th{$k$-NN} & \Th{Prob}  \\
\midrule
Baseline                 & 300 & 72.2 & 74.3 \\
\rowcolor{LightCyan}
\Ours                    & 300 & \bf{72.6} & \bf{75.0} \\
\bottomrule
\end{tabular}
\vspace{3pt}
\caption{\emph{Image classification} top-1 accuracy (\%) on ImageNet-1k. Self-supervised pre-training with DINO~\cite{dino} for 300 epochs. Baseline: GAP for convolutional, \cls for transformers.}
\label{tab:Self100percImageNet300ep}
\end{table}
%------------------------------------------------------------------------------

\paragraph{Self-supervised pre-training}

On the 100\% of ImageNet-1k, we train ViT-S with DINO~\cite{dino} for 300 epochs. \autoref{tab:Self100percImageNet300ep} shows that \Ours improves over the baseline by 0.4\% $k$-NN and 0.7\% linear probing.

%------------------------------------------------------------------------------
\begin{table}
\small
\centering
\setlength{\tabcolsep}{3pt}
\begin{tabular}{lccc}
\toprule
\Th{Method} & \Th{mIoU} &\Th{mAcc} & \Th{aAcc}  \\ \midrule
Baseline  &  26.4  &   34.0  & 71.6 \\
\rowcolor{LightCyan}
\Ours    &  \bf{27.9}  & \bf{35.7} & \bf{72.6} \\
\bottomrule
\end{tabular}
\vspace{3pt}
\caption{\emph{Semantic segmentation} on ADE20K~\cite{ade20k}. ViT-S pre-trained on ImageNet-1k for 100 epochs. Self-supervision with DINO~\cite{dino}.}
\label{tab:ade20k}
\end{table}
%------------------------------------------------------------------------------

\paragraph{Semantic segmentation}

We evaluate semantic segmentation on ADE20K~\cite{ade20k} under self-supervised pre-training. To evaluate the quality of the learned representation, we only fine-tune a linear layer on top of the fixed patch features, as in iBOT~\cite{ibot}. \autoref{tab:ade20k} shows that \Ours increases all scores by more than 1\% over the baseline. These results testify the improved quality of the learned representations when pre-training with \Ours.

%------------------------------------------------------------------------------

\paragraph{Background changes}

Deep neural networks often rely on the image background, which can limit their ability to generalize well. To achieve better performance, these models must be able to cope with changes in the background and prioritize the foreground. To evaluate \Ours robustness to the background changes, we use the \imagenet-9~\cite{xiao2021noise} (IN-9) dataset. In four of these datasets, \ie, Only-FG (OF), Mixed-Same (MS), Mixed-Rand (MR), and Mixed-Next (MN), the background is modified. The three other datasets feature masked foregrounds, \ie, No-FG (NF), Only-BG-B (OBB), and Only-BG-T (OBT).

%------------------------------------------------------------------------------
\begin{table}
\small
\centering
\setlength{\tabcolsep}{3pt}
\begin{tabular}{lccc|cc} \toprule
\mr{2}{\Th{Network}} & \mr{2}{\Th{Method}} & \mc{2}{\Th{$\cR$Oxford}} & \mc{2}{\Th{$\cR$Paris}} \\ \cmidrule{3-6}
                  &           & \Th{Medium}  & \Th{Hard}   & \Th{Medium}  & \Th{Hard}   \\ \midrule
% \mc{5}{Supervised}  \\ \midrule
% \mr{2}{ViT-S}     & Baseline    & 23.05        & \bf{6.55}   & \bf{55.07}   & \bf{27.09}  \\
%                   & \ours       & \bf{23.68}   & 5.58        & 54.23        & 26.89       \\ \midrule
% \mc{5}{DINO}        \\ \midrule
 & Baseline  & 27.2        & 7.9        & 47.3        & 19.0       \\
\rowcolor{LightCyan}
\mr{-2}{ResNet-50}  &  \ours     & \bf{29.7}   & \bf{8.7}   & \bf{51.6}   & \bf{23.0}  \\ \midrule
     & Baseline  & 29.4       & 10.0       & 54.6        & 26.2       \\
\rowcolor{LightCyan}
\mr{-2}{ViT-S}    & \ours     & \bf{32.1}   & \bf{10.6}  & \bf{56.5}   & \bf{27.3}  \\
\bottomrule
\end{tabular}
\vspace{3pt}
\caption{\emph{Image retrieval} mAP (\%) without fine-tuning on $\cR$Oxford and $\cR$Paris~\cite{radenovic2018revisiting}. Self-supervised pre-training with DINO~\cite{dino} on \imagenet for 100 epochs.}
\label{tab:100percDownStreamRox}
\end{table}
%------------------------------------------------------------------------------

\paragraph{Image retrieval without fine-tuning}

While classification accuracy indicates ability of a model to recognize objects of the same classes as those it was trained for, it does not necessarily reflect its ability to capture the visual similarity between images, when tested on a dataset from a different distribution. Here, we evaluate this property of visual features using ResNet-50 and ViT-S; for particular object retrieval without fine-tuning on $\cR$Oxford and $\cR$Paris~\cite{radenovic2018revisiting}. In \autoref{tab:100percDownStreamRox}, we observe that \Ours is very effective, improving the retrieval performance of both models on all datasets and evaluation protocols over the baseline.

%------------------------------------------------------------------------------
\begin{table*}
\small
\centering
\setlength{\tabcolsep}{5pt}
\begin{tabular}{lcccc|ccc|ccc|ccc} \toprule
\mr{2}{\Th{network}} & \mr{2}{\Th{Method}} & \mc{3}{CUB200} & \mc{3}{CARS196} & \mc{3}{SOP} & \mc{3}{\Th{In-Shop}} \\ \cmidrule{3-14}
& & R@1 & 2 & 4 & R@1 & 2 & 4 & R@1 & 10 & 100 & R@1 & 10 & 20 \\ \midrule
\mc{14}{\Th{Supervised}} \\ \midrule
 & Baseline &  42.7 & 55.2 & 67.7 & 42.3 & 54.2 & 65.7 & 48.3 & 63.2 & 71.8 & \bf{27.6} & \bf{49.9} & \bf{56.5} \\
\rowcolor{LightCyan}
\mr{-2}{ResNet-50} & \ours & \bf{43.0} & \bf{55.2} & \bf{67.9} & \bf{43.8} & \bf{56.2} & \bf{67.4} & \bf{48.7} & \bf{64.1} & \bf{72.9} & 27.0 & 49.9 & 56.5 \\\midrule
 & Baseline &  55.8 & 68.3 & 78.3 & 38.2 & 50.3 & 61.8 & 54.1 & 69.2 & 81.6 & 30.9 & 56.5 & 63.2 \\
\rowcolor{LightCyan}
\mr{-2}{ViT-S} & \ours & \bf{56.8} & \bf{69.6} & \bf{79.2} & \bf{38.9} & \bf{50.7} & \bf{63.3} & \bf{54.2} & \bf{69.4} & \bf{81.9} & \bf{32.8} & \bf{57.6} & \bf{64.3}\\\midrule
\mc{14}{\Th{Self-supervised}} \\ \midrule
 & Baseline &  26.0 & 36.2 & 46.9 & \bf{34.1} & \bf{44.2} & \bf{55.0} & 51.2 & 65.3 & 76.5 & 37.1 & 58.4 & 64.1 \\
\rowcolor{LightCyan}
\mr{-2}{ResNet-50} & \ours & \bf{30.7} & \bf{40.9} & \bf{53.3} & 33.6 & 43.6 & 54.3 & \bf{52.1} & \bf{66.5} & \bf{77.2} &\bf{38.1} & \bf{60.0} & \bf{65.6} \\\midrule
 & Baseline & 56.7 & 69.4 & 80.5 & 37.5 & 47.5 & 58.4 & \bf{59.8} & \bf{74.4} & \bf{85.4} & 40.4 & 63.9 & 70.3 \\
\rowcolor{LightCyan}
\mr{-2}{ViT-S} & \ours & \bf{61.8} & \bf{74.4} & \bf{83.6} & \bf{37.6} & \bf{48.0} & \bf{58.4} & 59.5 & 73.9 & 85.0 & \bf{41.1} & \bf{64.3} & \bf{70.8} \\
\bottomrule
\end{tabular}
\vspace{3pt}
\caption{\emph{Fine-grained classification} Recall@$k$ (R@$k$, \%) without fine-tuning on four datasets, following the same protocol as~\cite{reality_check,attmask}. Models pre-trained on \imagenet for 100 epochs. Self-supervision with DINO~\cite{dino}.}
\label{tab:100percDownStreamMetric}
\end{table*}
%------------------------------------------------------------------------------

\paragraph{Fine-grained classification}

We evaluate fine-grained classification using ResNet-50 and ViT-S, both supervised and self-supervised, following~\cite{attmask}. We extract features from test set images and directly apply nearest neighbor search, measuring Recall@$k$. \autoref{tab:100percDownStreamMetric} shows that \Ours is superior to the baseline in most of the datasets, models and supervision settings, with the exception of ResNet-50 supervised on In-Shop, ResNet-50 self-supervised on Cars196 and ViT-S self-supervised on SOP (3 out of 16 cases). The improvement is roughly 1-2\% Recall@1 in most cases, and is most pronounced on self-supervised on CUB200, roughly 5\%.
\subsection{More ablations}
\label{sec:more-ablation}

%------------------------------------------------------------------------------
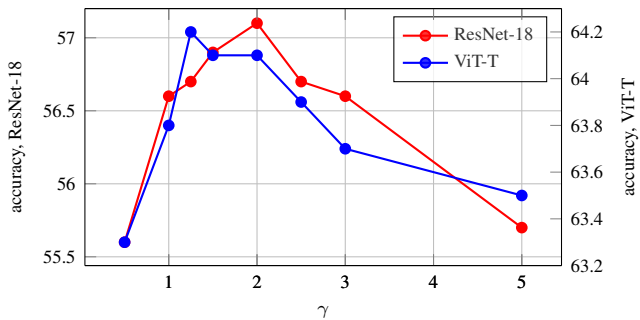
\begin{figure}
\centering
\pgfplotstableread{
	alpha  r18    vit
	0.5    55.6   63.3
	1.0    56.6   63.8
	1.25   56.7   64.2
	1.5    56.9   64.1
	2.0    57.1   64.1
	2.5    56.7   63.9
	3.0    56.6   63.7
	5.0    55.7   63.5
}{\pool}
\begin{tikzpicture}[
	sec/.style={blue, mark=*},
]
\begin{axis}[
	font=\scriptsize,
	width=.95\linewidth,
	height=0.6\linewidth,
	xlabel={$\gamma$},
	ylabel={accuracy, ResNet-18},
	axis y line*=left,
	ymax=57.2,
	legend pos=north east,
	hide/.style={mark=none,draw=none,legend image post style=sec},
]
\addplot[red,mark=*] table[x=alpha, y=r18] {\pool}; \leg{ResNet-18}
\addplot[hide] table[x=alpha, y=vit] {\pool}; \leg{ViT-T}
%\addplot[purple,dashed] coordinates {(0,55.0) (5.0,55.0)}; % Baseline line for ResNet-18
%\addplot[black,dashed] coordinates {(0,56.55) (5.0,56.55)}; % No alpha line for ResNet-18
\end{axis}
\begin{axis}[
	font=\scriptsize,
	width=.95\linewidth,
	height=0.6\linewidth,
	ylabel={accuracy, ViT-T},
	axis y line*=right,
	ymax=64.3,
	grid=none,
]
% \addplot[red,mark=*] table[x=alpha, y=r18] {\pool}; \leg{ResNet-18}
\addplot[blue,mark=*] table[x=alpha, y=vit] {\pool}; % \leg{ViT-T}
%\addplot[purple,dashed] coordinates {(0,62.6) (5.0,62.6)}; % Baseline line for ViT-T
%\addplot[black,dashed] coordinates {(0,63.8) (5.0,63.8)}; % No alpha line for ViT-T
\end{axis}
\end{tikzpicture}
\caption{\emph{Image classification} top-1 accuracy (\%) \vs \emph{exponent} $\gamma = (1-\alpha)/2$~\eq{sp} for ResNet-18 supervised on ImageNet-20\% and ViT-T supervised on ImageNet-1k, both for 100 epochs.}
\label{fig:alphas-plot}
\end{figure}
%------------------------------------------------------------------------------

\paragraph{Pooling parameter $\alpha$~\eq{sp}}

We ablate the effect of parameter $\alpha$ of the pooling function $f_\alpha$~\eq{sp} on the classification performance of \Ours using ResNet-18 on ImageNet-20\% and ViT-T on \imagenet for 100 epochs. We find learnable $\alpha$ (or $\gamma = (1-\alpha)/2$) to be inferior both in terms of performance and attention map quality. For ResNet-18 on ImageNet-20\%, it gives top-1 accuracy $56.0\%$. Clamping to $\gamma = 5$ gives $56.3\%$ and using a $10\times$ smaller learning rate gives $56.5\%$.

In \autoref{fig:alphas-plot}, we set exponent $\gamma$ to be a hyperparameter and observe that for both networks, values between $1$ and $3$ are relatively stable. Specifically, the best choice is $2$ for ResNet-18 and $1.25$ for ViT-T. Thus, we choose exponent $2$ for convolutional networks (ResNet-18, ResNet-50 and ConvNeXt-S) and $1.25$ for vision transformers (ViT-T, ViT-S and ViT-B).

\subsection{More visualizations}
\label{sec:more-vis}

\paragraph{Attention maps: ViT}

\begin{figure*}
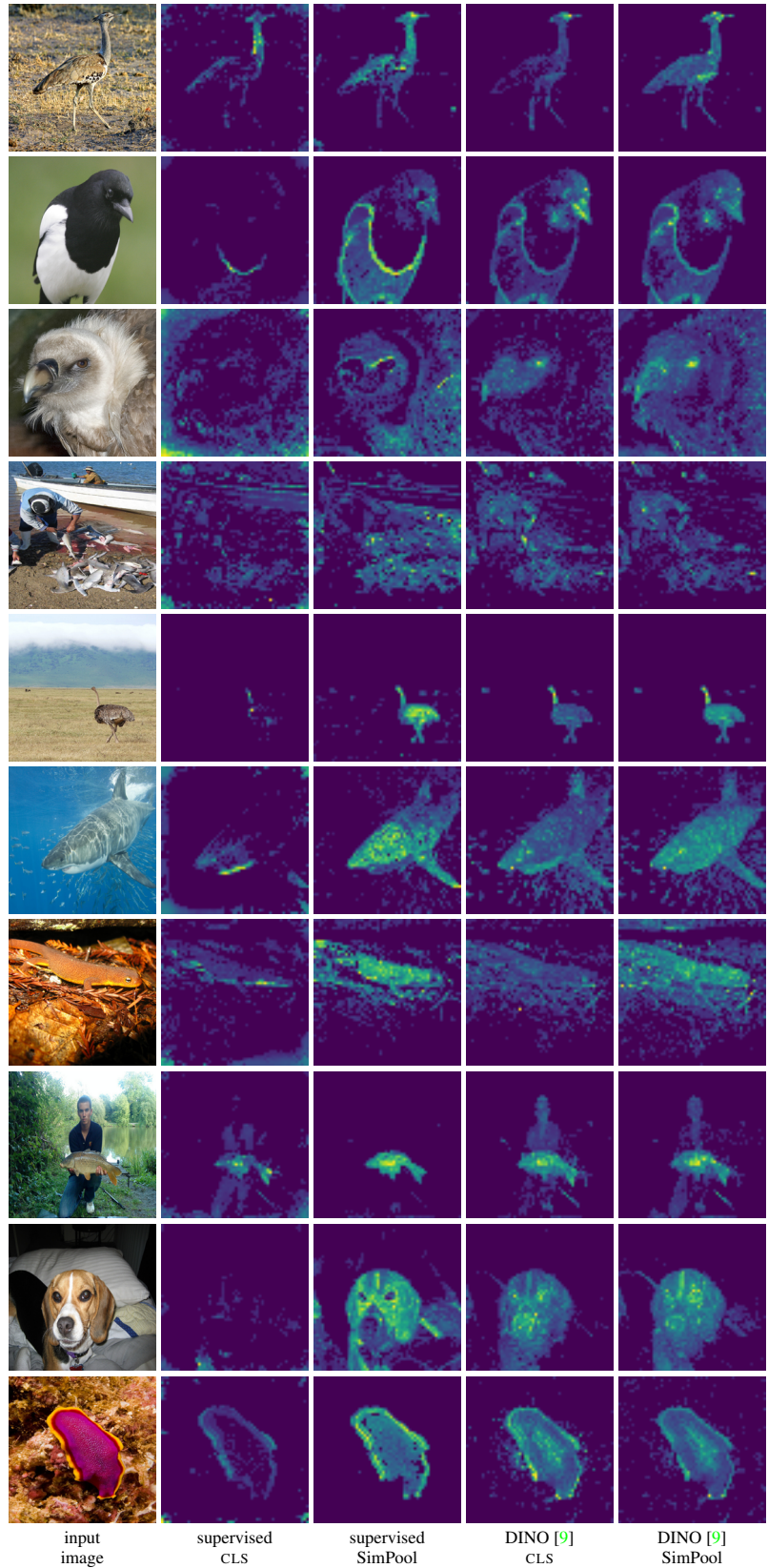

\scriptsize
\centering
\setlength{\tabcolsep}{1.2pt}
\newcommand{\sz}{.117}
\begin{tabular}{ccccc}

\fig[\sz]{attmaps/ILSVRC2012_val_00002311_orig.PNG} &
\fig[\sz]{attmaps/ILSVRC2012_val_00002311_supervised_official.PNG} &
\fig[\sz]{attmaps/ILSVRC2012_val_00002311_supervised_simpool.PNG} &
\fig[\sz]{attmaps/ILSVRC2012_val_00002311_dino_official.PNG} &
\fig[\sz]{attmaps/ILSVRC2012_val_00002311_dino_simpool.PNG} \\

\fig[\sz]{attmaps/ILSVRC2012_val_00003762_orig.PNG} &
\fig[\sz]{attmaps/ILSVRC2012_val_00003762_supervised_official.PNG} &
\fig[\sz]{attmaps/ILSVRC2012_val_00003762_supervised_simpool.PNG} &
\fig[\sz]{attmaps/ILSVRC2012_val_00003762_dino_official.PNG} &
\fig[\sz]{attmaps/ILSVRC2012_val_00003762_dino_simpool.PNG} \\

\fig[\sz]{attmaps/ILSVRC2012_val_00009119_orig.PNG} &
\fig[\sz]{attmaps/ILSVRC2012_val_00009119_supervised_official.PNG} &
\fig[\sz]{attmaps/ILSVRC2012_val_00009119_supervised_simpool.PNG} &
\fig[\sz]{attmaps/ILSVRC2012_val_00009119_dino_official.PNG} &
\fig[\sz]{attmaps/ILSVRC2012_val_00009119_dino_simpool.PNG} \\

\fig[\sz]{attmaps/ILSVRC2012_val_00012546_orig.PNG} &
\fig[\sz]{attmaps/ILSVRC2012_val_00012546_supervised_official.PNG} &
\fig[\sz]{attmaps/ILSVRC2012_val_00012546_supervised_simpool.PNG} &
\fig[\sz]{attmaps/ILSVRC2012_val_00012546_dino_official.PNG} &
\fig[\sz]{attmaps/ILSVRC2012_val_00012546_dino_simpool.PNG} \\

\fig[\sz]{attmaps/ILSVRC2012_val_00023778_orig.PNG} &
\fig[\sz]{attmaps/ILSVRC2012_val_00023778_supervised_official.PNG} &
\fig[\sz]{attmaps/ILSVRC2012_val_00023778_supervised_simpool.PNG} &
\fig[\sz]{attmaps/ILSVRC2012_val_00023778_dino_official.PNG} &
\fig[\sz]{attmaps/ILSVRC2012_val_00023778_dino_simpool.PNG} \\

\fig[\sz]{attmaps/ILSVRC2012_val_00025900_orig.PNG} &
\fig[\sz]{attmaps/ILSVRC2012_val_00025900_supervised_official.PNG} &
\fig[\sz]{attmaps/ILSVRC2012_val_00025900_supervised_simpool.PNG} &
\fig[\sz]{attmaps/ILSVRC2012_val_00025900_dino_official.PNG} &
\fig[\sz]{attmaps/ILSVRC2012_val_00025900_dino_simpool.PNG} \\

\fig[\sz]{attmaps/ILSVRC2012_val_00041540_orig.PNG} &
\fig[\sz]{attmaps/ILSVRC2012_val_00041540_supervised_official.PNG} &
\fig[\sz]{attmaps/ILSVRC2012_val_00041540_supervised_simpool.PNG} &
\fig[\sz]{attmaps/ILSVRC2012_val_00041540_dino_official.PNG} &
\fig[\sz]{attmaps/ILSVRC2012_val_00041540_dino_simpool.PNG} \\

\fig[\sz]{attmaps/ILSVRC2012_val_00048204_orig.PNG} &
\fig[\sz]{attmaps/ILSVRC2012_val_00048204_supervised_official.PNG} &
\fig[\sz]{attmaps/ILSVRC2012_val_00048204_supervised_simpool.PNG} &
\fig[\sz]{attmaps/ILSVRC2012_val_00048204_dino_official.PNG} &
\fig[\sz]{attmaps/ILSVRC2012_val_00048204_dino_simpool.PNG} \\

\fig[\sz]{attmaps/ILSVRC2012_val_00049604_orig.PNG} &
\fig[\sz]{attmaps/ILSVRC2012_val_00049604_supervised_official.PNG} &
\fig[\sz]{attmaps/ILSVRC2012_val_00049604_supervised_simpool.PNG} &
\fig[\sz]{attmaps/ILSVRC2012_val_00049604_dino_official.PNG} &
\fig[\sz]{attmaps/ILSVRC2012_val_00049604_dino_simpool.PNG} \\

\fig[\sz]{attmaps/ILSVRC2012_val_00037106_orig.PNG} &
\fig[\sz]{attmaps/ILSVRC2012_val_00037106_supervised_official.PNG} &
\fig[\sz]{attmaps/ILSVRC2012_val_00037106_supervised_simpool.PNG} &
\fig[\sz]{attmaps/ILSVRC2012_val_00037106_dino_official.PNG} &
\fig[\sz]{attmaps/ILSVRC2012_val_00037106_dino_simpool.PNG} \\

input &
supervised &
supervised &
DINO~\cite{dino} &
DINO~\cite{dino} \\

image &
\cls &
\Ours &
\cls &
\Ours \\

\end{tabular}
\vspace{3pt}
\caption{\emph{Attention maps} of ViT-S~\cite{vit} trained on ImageNet-1k for 100 epochs under supervision and  self-supervision with DINO~\cite{dino}. For ViT-S baseline, we use the mean attention map of the \cls token. For \Ours, we use the attention map $\va$~\eq{sp-att}. Input image resolution: $896 \times 896$; patches: $16 \times 16$; output attention map: $56 \times 56$.}
\label{fig:attention-maps}
\end{figure*}

\autoref{fig:attention-maps} shows attention maps of supervised and self-supervised ViT-S trained on \imagenet. The ViT-S baseline uses the \cls token for pooling by default. For \Ours, we remove the \cls stream entirely from the encoder and use the attention map $\va$~\eq{sp-att}.

We observe that under \emph{self-supervision}, the attention map quality of \Ours is on par with the baseline and in some cases the object of interest is slightly more pronounced, \eg, rows 1, 3, 6 and 7.

What is more impressive is \emph{supervised} training. In this case, the baseline has very low quality of attention maps, focusing only on part of the object of interest (\eg, rows 1, 2, 5, 6, 10), focusing on background more than self-supervised (\eg, rows 1, 4, 6, 7, 8), even missing the object of interest entirely (\eg, rows 3, 9). By contrast, the quality of attention maps of \Ours is superior even to self-supervised, attending more to the object surface and less background.

\paragraph{Segmentation masks}

\begin{figure*}
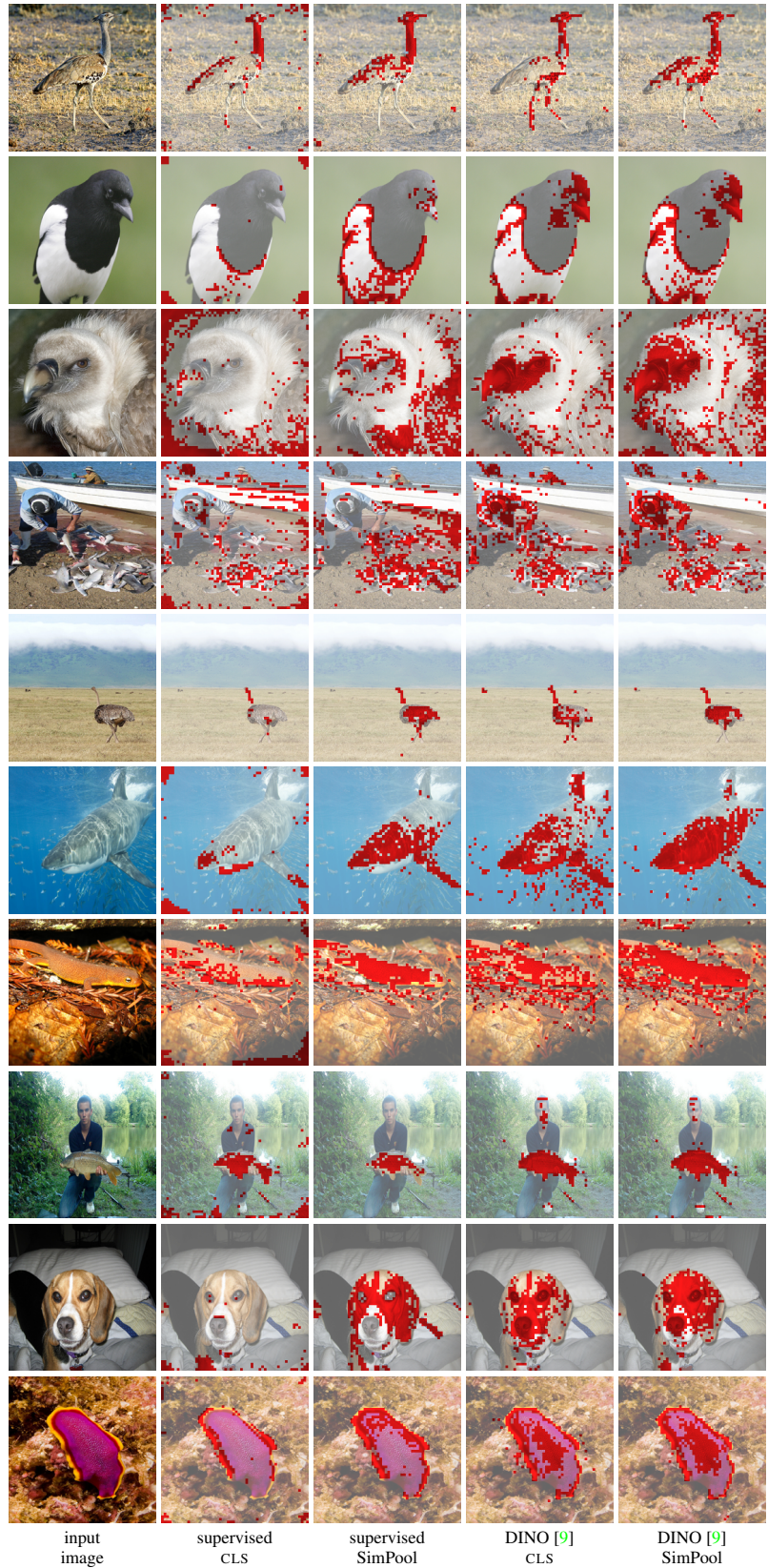

\scriptsize
\centering
\setlength{\tabcolsep}{1.2pt}
\newcommand{\sz}{.117}
\begin{tabular}{ccccc}

\fig[\sz]{attmaps/ILSVRC2012_val_00002311_orig.PNG} &
\fig[\sz]{segmaps/ILSVRC2012_val_00002311_overlay_supervised_official.PNG} &
\fig[\sz]{segmaps/ILSVRC2012_val_00002311_overlay_supervised_simpool.PNG} &
\fig[\sz]{segmaps/ILSVRC2012_val_00002311_overlay_dino_official.PNG} &
\fig[\sz]{segmaps/ILSVRC2012_val_00002311_overlay_dino_simpool.PNG} \\

\fig[\sz]{attmaps/ILSVRC2012_val_00003762_orig.PNG} &
\fig[\sz]{segmaps/ILSVRC2012_val_00003762_overlay_supervised_official.PNG} &
\fig[\sz]{segmaps/ILSVRC2012_val_00003762_overlay_supervised_simpool.PNG} &
\fig[\sz]{segmaps/ILSVRC2012_val_00003762_overlay_dino_official.PNG} &
\fig[\sz]{segmaps/ILSVRC2012_val_00003762_overlay_dino_simpool.PNG} \\

\fig[\sz]{attmaps/ILSVRC2012_val_00009119_orig.PNG} &
\fig[\sz]{segmaps/ILSVRC2012_val_00009119_overlay_supervised_official.PNG} &
\fig[\sz]{segmaps/ILSVRC2012_val_00009119_overlay_supervised_simpool.PNG} &
\fig[\sz]{segmaps/ILSVRC2012_val_00009119_overlay_dino_official.PNG} &
\fig[\sz]{segmaps/ILSVRC2012_val_00009119_overlay_dino_simpool.PNG} \\

\fig[\sz]{attmaps/ILSVRC2012_val_00012546_orig.PNG} &
\fig[\sz]{segmaps/ILSVRC2012_val_00012546_overlay_supervised_official.PNG} &
\fig[\sz]{segmaps/ILSVRC2012_val_00012546_overlay_supervised_simpool.PNG} &
\fig[\sz]{segmaps/ILSVRC2012_val_00012546_overlay_dino_official.PNG} &
\fig[\sz]{segmaps/ILSVRC2012_val_00012546_overlay_dino_simpool.PNG} \\

\fig[\sz]{attmaps/ILSVRC2012_val_00023778_orig.PNG} &
\fig[\sz]{segmaps/ILSVRC2012_val_00023778_overlay_supervised_official.PNG} &
\fig[\sz]{segmaps/ILSVRC2012_val_00023778_overlay_supervised_simpool.PNG} &
\fig[\sz]{segmaps/ILSVRC2012_val_00023778_overlay_dino_official.PNG} &
\fig[\sz]{segmaps/ILSVRC2012_val_00023778_overlay_dino_simpool.PNG} \\

\fig[\sz]{attmaps/ILSVRC2012_val_00025900_orig.PNG} &
\fig[\sz]{segmaps/ILSVRC2012_val_00025900_overlay_supervised_official.PNG} &
\fig[\sz]{segmaps/ILSVRC2012_val_00025900_overlay_supervised_simpool.PNG} &
\fig[\sz]{segmaps/ILSVRC2012_val_00025900_overlay_dino_official.PNG} &
\fig[\sz]{segmaps/ILSVRC2012_val_00025900_overlay_dino_simpool.PNG} \\

\fig[\sz]{attmaps/ILSVRC2012_val_00041540_orig.PNG} &
\fig[\sz]{segmaps/ILSVRC2012_val_00041540_overlay_supervised_official.PNG} &
\fig[\sz]{segmaps/ILSVRC2012_val_00041540_overlay_supervised_simpool.PNG} &
\fig[\sz]{segmaps/ILSVRC2012_val_00041540_overlay_dino_official.PNG} &
\fig[\sz]{segmaps/ILSVRC2012_val_00041540_overlay_dino_simpool.PNG} \\

\fig[\sz]{attmaps/ILSVRC2012_val_00048204_orig.PNG} &
\fig[\sz]{segmaps/ILSVRC2012_val_00048204_overlay_supervised_official.PNG} &
\fig[\sz]{segmaps/ILSVRC2012_val_00048204_overlay_supervised_simpool.PNG} &
\fig[\sz]{segmaps/ILSVRC2012_val_00048204_overlay_dino_official.PNG} &
\fig[\sz]{segmaps/ILSVRC2012_val_00048204_overlay_dino_simpool.PNG} \\

\fig[\sz]{attmaps/ILSVRC2012_val_00049604_orig.PNG} &
\fig[\sz]{segmaps/ILSVRC2012_val_00049604_overlay_supervised_official.PNG} &
\fig[\sz]{segmaps/ILSVRC2012_val_00049604_overlay_supervised_simpool.PNG} &
\fig[\sz]{segmaps/ILSVRC2012_val_00049604_overlay_dino_official.PNG} &
\fig[\sz]{segmaps/ILSVRC2012_val_00049604_overlay_dino_simpool.PNG} \\

\fig[\sz]{attmaps/ILSVRC2012_val_00037106_orig.PNG} &
\fig[\sz]{segmaps/ILSVRC2012_val_00037106_overlay_supervised_official.PNG} &
\fig[\sz]{segmaps/ILSVRC2012_val_00037106_overlay_supervised_simpool.PNG} &
\fig[\sz]{segmaps/ILSVRC2012_val_00037106_overlay_dino_official.PNG} &
\fig[\sz]{segmaps/ILSVRC2012_val_00037106_overlay_dino_simpool.PNG} \\

input &
supervised &
supervised &
DINO~\cite{dino} &
DINO~\cite{dino} \\

image &
\cls &
\Ours &
\cls &
\Ours \\

\end{tabular}
\vspace{3pt}
\caption{\emph{Segmentation masks} of ViT-S~\cite{vit} trained on ImageNet-1k for 100 epochs under supervision and self-supervision with DINO~\cite{dino}. For ViT-S baseline, we use the attention map of the \cls token. For \Ours, we use the attention map $\va$~\eq{sp-att}. Same as \autoref{fig:attention-maps}, with attention map value thresholded at 60\% of mass and mask overlaid on input image.}
\label{fig:segmentation-masks}
\end{figure*}

\autoref{fig:segmentation-masks} shows the same images for the same setting as in \autoref{fig:attention-maps}, but this time overlays segmenation masks on top input images, corresponding to more than 60\% mass of the attention map. Again, \Ours is on par with baseline when self-supervised, supervised baseline has poor quality and supervised \Ours is a lot better, although its superiority is not as evident as with the raw attention maps.

\paragraph{Object localization}

\begin{figure*}
\scriptsize
\centering
\setlength{\tabcolsep}{1.2pt}
\newcommand{\sz}{.117}
\begin{tabular}{ccccc}

\fig[\sz]{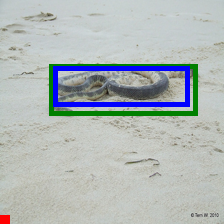} &
\fig[\sz]{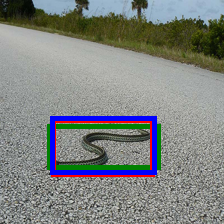} &
\fig[\sz]{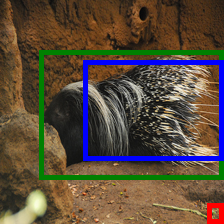} &
\fig[\sz]{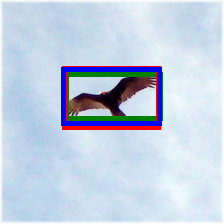} &
\fig[\sz]{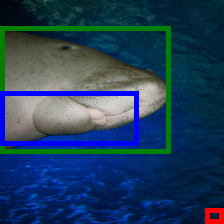} \\

\fig[\sz]{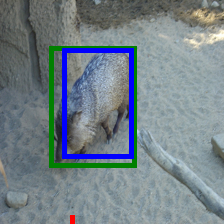} &
\fig[\sz]{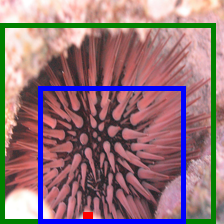} &
\fig[\sz]{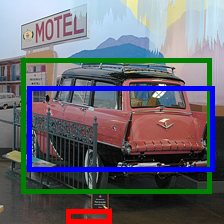} &
\fig[\sz]{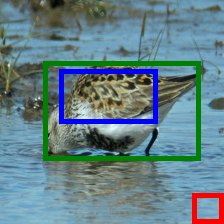} &
\fig[\sz]{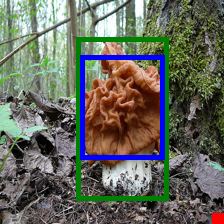} \\

\fig[\sz]{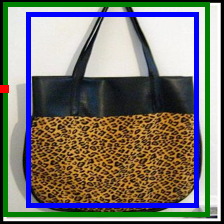} &
\fig[\sz]{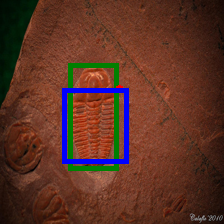} &
\fig[\sz]{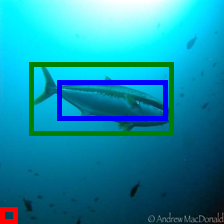} &
\fig[\sz]{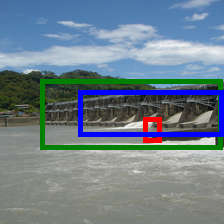} &
\fig[\sz]{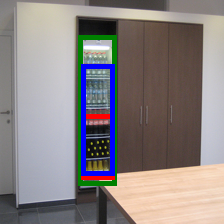} \\

\fig[\sz]{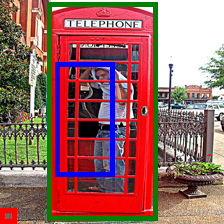} &
\fig[\sz]{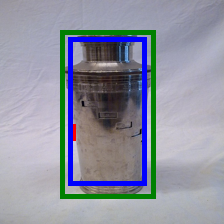} &
\fig[\sz]{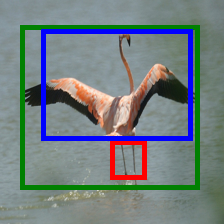} &
\fig[\sz]{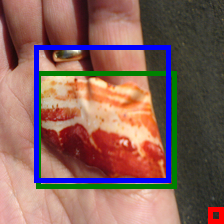} &
\fig[\sz]{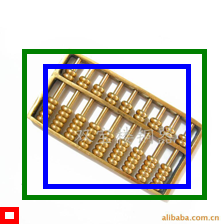} \\

\fig[\sz]{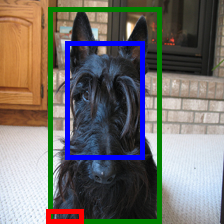} &
\fig[\sz]{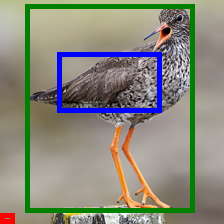} &
\fig[\sz]{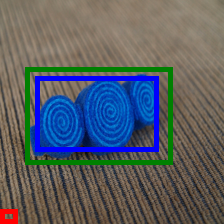} &
\fig[\sz]{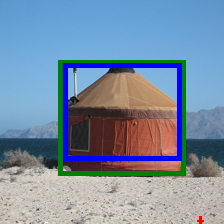} &
\fig[\sz]{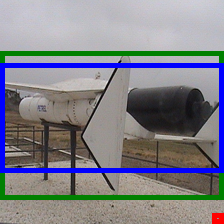} \\

\fig[\sz]{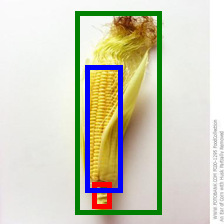} &
\fig[\sz]{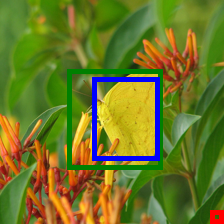} &
\fig[\sz]{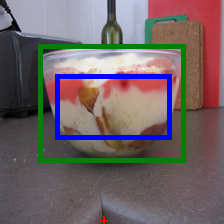} &
\fig[\sz]{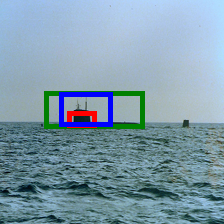} &
\fig[\sz]{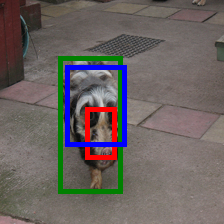} \\

\fig[\sz]{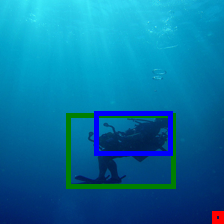} &
\fig[\sz]{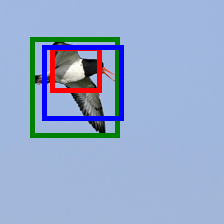} &
\fig[\sz]{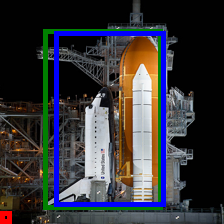} &
\fig[\sz]{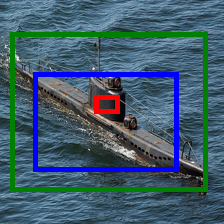} &
\fig[\sz]{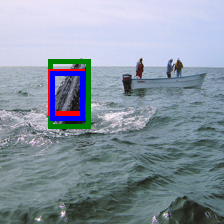} \\

\fig[\sz]{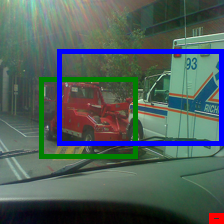} &
\fig[\sz]{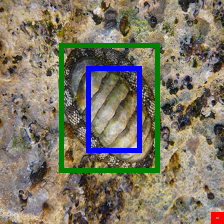} &
\fig[\sz]{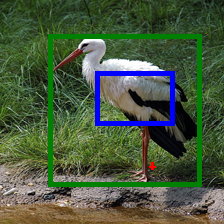} &
\fig[\sz]{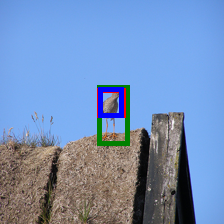} &
\fig[\sz]{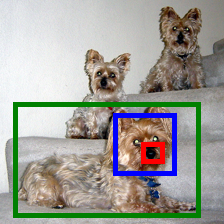} \\

\fig[\sz]{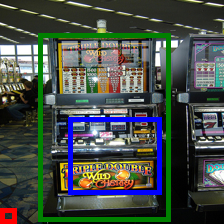} &
\fig[\sz]{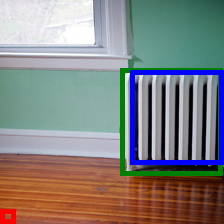} &
\fig[\sz]{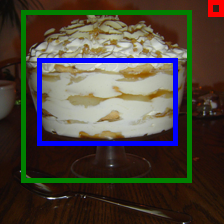} &
\fig[\sz]{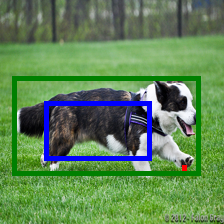} &
\fig[\sz]{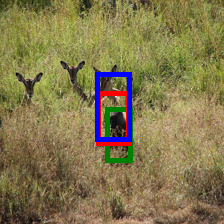} \\

\fig[\sz]{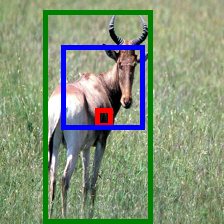} &
\fig[\sz]{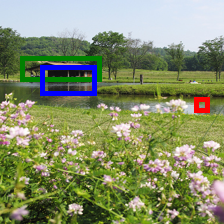} &
\fig[\sz]{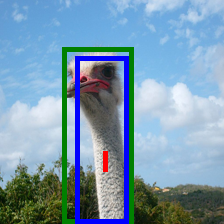} &
\fig[\sz]{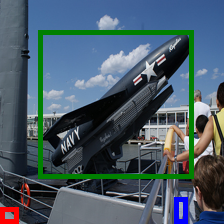} &
\fig[\sz]{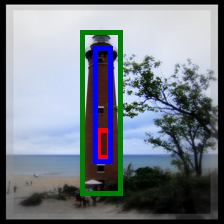} \\

\end{tabular}
\vspace{3pt}
\caption{\emph{Object localization} on \imagenet with ViT-S~\cite{vit} supervised pre-training on \imagenet-1k for 100 epochs. Bounding boxes obtained from experiment of \autoref{tab:maxboxacc}, following~\cite{wsol}. \gp{Green}: ground-truth bounding boxes; \red{red}: baseline, predicted by the attention map of the \cls token; \blue{blue}: predicted by \Ours, using the attention map $\va$~\eq{sp-att}.}
\label{fig:localization}
\end{figure*}

\autoref{fig:localization} visualizes object localization results, comparing bounding boxes of \Ours with the baseline. The results are obtained from the experiments of \autoref{tab:maxboxacc}, using ViT-S with supervised pre-training.
We observe that the baseline systematically fails to localize the objects accurately. On the other hand, \Ours allows reasonable localization of the object of interest just from the attention map, without any supervision other than the image-level label.

\paragraph{Attention maps: The effect of $\gamma$}

\begin{figure*}
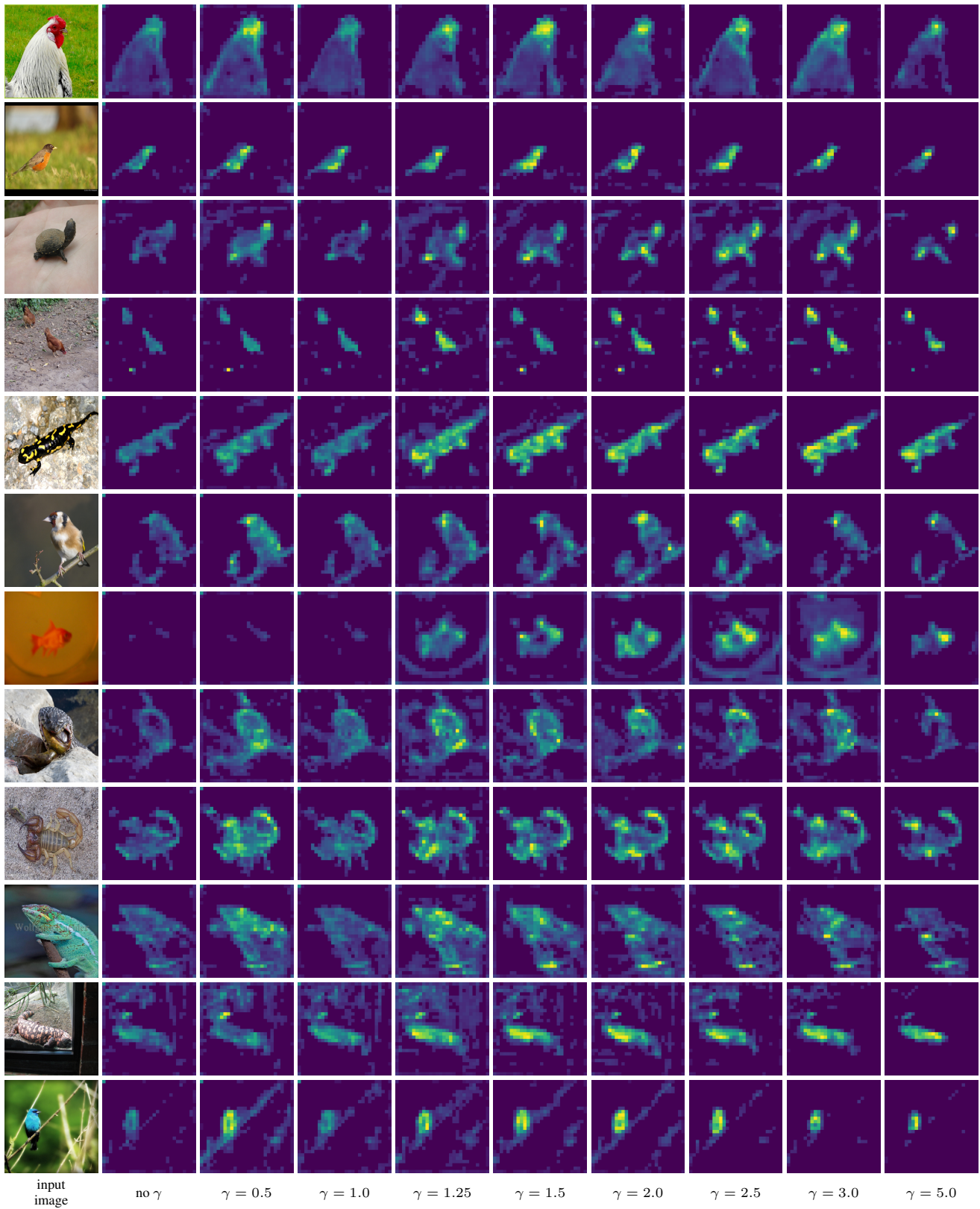

\scriptsize
\centering
\setlength{\tabcolsep}{1.2pt}
\newcommand{\sz}{.095}
\begin{tabular}{cccccccccc}

\fig[\sz]{resnet18alphaattmaps/ILSVRC2012_val_00004550_orig.PNG} &
\fig[\sz]{resnet18alphaattmaps/ILSVRC2012_val_00004550_attmap_28x28_no_blur_resnet18_simpool_no_alpha_ep100.PNG} &
\fig[\sz]{resnet18alphaattmaps/ILSVRC2012_val_00004550_attmap_28x28_no_blur_resnet18_simpool_alpha0.5_ep100.PNG} &
\fig[\sz]{resnet18alphaattmaps/ILSVRC2012_val_00004550_attmap_28x28_no_blur_resnet18_simpool_alpha1.0_ep100.PNG} &
\fig[\sz]{resnet18alphaattmaps/ILSVRC2012_val_00004550_attmap_28x28_no_blur_resnet18_simpool_alpha1.25_ep100.PNG} &
\fig[\sz]{resnet18alphaattmaps/ILSVRC2012_val_00004550_attmap_28x28_no_blur_resnet18_simpool_alpha1.5_ep100.PNG} &
\fig[\sz]{resnet18alphaattmaps/ILSVRC2012_val_00004550_attmap_28x28_no_blur_resnet18_simpool_alpha2.0_ep100.PNG} &
\fig[\sz]{resnet18alphaattmaps/ILSVRC2012_val_00004550_attmap_28x28_no_blur_resnet18_simpool_alpha2.5_ep100.PNG} &
\fig[\sz]{resnet18alphaattmaps/ILSVRC2012_val_00004550_attmap_28x28_no_blur_resnet18_simpool_alpha3.0_ep100.PNG} & 
\fig[\sz]{resnet18alphaattmaps/ILSVRC2012_val_00004550_attmap_28x28_no_blur_resnet18_simpool_alpha5.0_ep100.PNG} \\

%\fig[\sz]{resnet18alphaattmaps/ILSVRC2012_val_00004915_orig.PNG} &
%\fig[\sz]{resnet18alphaattmaps/ILSVRC2012_val_00004915_attmap_28x28_no_blur_resnet18_simpool_no_alpha_ep100.PNG} &
%\fig[\sz]{resnet18alphaattmaps/ILSVRC2012_val_00004915_attmap_28x28_no_blur_resnet18_simpool_alpha0.5_ep100.PNG} &
%\fig[\sz]{resnet18alphaattmaps/ILSVRC2012_val_00004915_attmap_28x28_no_blur_resnet18_simpool_alpha1.0_ep100.PNG} &
%\fig[\sz]{resnet18alphaattmaps/ILSVRC2012_val_00004915_attmap_28x28_no_blur_resnet18_simpool_alpha1.25_ep100.PNG} &
%\fig[\sz]{resnet18alphaattmaps/ILSVRC2012_val_00004915_attmap_28x28_no_blur_resnet18_simpool_alpha1.5_ep100.PNG} &
%\fig[\sz]{resnet18alphaattmaps/ILSVRC2012_val_00004915_attmap_28x28_no_blur_resnet18_simpool_alpha2.0_ep100.PNG} &
%\fig[\sz]{resnet18alphaattmaps/ILSVRC2012_val_00004915_attmap_28x28_no_blur_resnet18_simpool_alpha2.5_ep100.PNG} &
%\fig[\sz]{resnet18alphaattmaps/ILSVRC2012_val_00004915_attmap_28x28_no_blur_resnet18_simpool_alpha3.0_ep100.PNG} & 
%\fig[\sz]{resnet18alphaattmaps/ILSVRC2012_val_00004915_attmap_28x28_no_blur_resnet18_simpool_alpha5.0_ep100.PNG} \\

\fig[\sz]{resnet18alphaattmaps/ILSVRC2012_val_00005594_orig.PNG} &
\fig[\sz]{resnet18alphaattmaps/ILSVRC2012_val_00005594_attmap_28x28_no_blur_resnet18_simpool_no_alpha_ep100.PNG} &
\fig[\sz]{resnet18alphaattmaps/ILSVRC2012_val_00005594_attmap_28x28_no_blur_resnet18_simpool_alpha0.5_ep100.PNG} &
\fig[\sz]{resnet18alphaattmaps/ILSVRC2012_val_00005594_attmap_28x28_no_blur_resnet18_simpool_alpha1.0_ep100.PNG} &
\fig[\sz]{resnet18alphaattmaps/ILSVRC2012_val_00005594_attmap_28x28_no_blur_resnet18_simpool_alpha1.25_ep100.PNG} &
\fig[\sz]{resnet18alphaattmaps/ILSVRC2012_val_00005594_attmap_28x28_no_blur_resnet18_simpool_alpha1.5_ep100.PNG} &
\fig[\sz]{resnet18alphaattmaps/ILSVRC2012_val_00005594_attmap_28x28_no_blur_resnet18_simpool_alpha2.0_ep100.PNG} &
\fig[\sz]{resnet18alphaattmaps/ILSVRC2012_val_00005594_attmap_28x28_no_blur_resnet18_simpool_alpha2.5_ep100.PNG} &
\fig[\sz]{resnet18alphaattmaps/ILSVRC2012_val_00005594_attmap_28x28_no_blur_resnet18_simpool_alpha3.0_ep100.PNG} & 
\fig[\sz]{resnet18alphaattmaps/ILSVRC2012_val_00005594_attmap_28x28_no_blur_resnet18_simpool_alpha5.0_ep100.PNG} \\

\fig[\sz]{resnet18alphaattmaps/ILSVRC2012_val_00010528_orig.PNG} &
\fig[\sz]{resnet18alphaattmaps/ILSVRC2012_val_00010528_attmap_28x28_no_blur_resnet18_simpool_no_alpha_ep100.PNG} &
\fig[\sz]{resnet18alphaattmaps/ILSVRC2012_val_00010528_attmap_28x28_no_blur_resnet18_simpool_alpha0.5_ep100.PNG} &
\fig[\sz]{resnet18alphaattmaps/ILSVRC2012_val_00010528_attmap_28x28_no_blur_resnet18_simpool_alpha1.0_ep100.PNG} &
\fig[\sz]{resnet18alphaattmaps/ILSVRC2012_val_00010528_attmap_28x28_no_blur_resnet18_simpool_alpha1.25_ep100.PNG} &
\fig[\sz]{resnet18alphaattmaps/ILSVRC2012_val_00010528_attmap_28x28_no_blur_resnet18_simpool_alpha1.5_ep100.PNG} &
\fig[\sz]{resnet18alphaattmaps/ILSVRC2012_val_00010528_attmap_28x28_no_blur_resnet18_simpool_alpha2.0_ep100.PNG} &
\fig[\sz]{resnet18alphaattmaps/ILSVRC2012_val_00010528_attmap_28x28_no_blur_resnet18_simpool_alpha2.5_ep100.PNG} &
\fig[\sz]{resnet18alphaattmaps/ILSVRC2012_val_00010528_attmap_28x28_no_blur_resnet18_simpool_alpha3.0_ep100.PNG} & 
\fig[\sz]{resnet18alphaattmaps/ILSVRC2012_val_00010528_attmap_28x28_no_blur_resnet18_simpool_alpha5.0_ep100.PNG} \\

%\fig[\sz]{resnet18alphaattmaps/ILSVRC2012_val_00011053_orig.PNG} &
%\fig[\sz]{resnet18alphaattmaps/ILSVRC2012_val_00011053_attmap_28x28_no_blur_resnet18_simpool_no_alpha_ep100.PNG} &
%\fig[\sz]{resnet18alphaattmaps/ILSVRC2012_val_00011053_attmap_28x28_no_blur_resnet18_simpool_alpha0.5_ep100.PNG} &
%\fig[\sz]{resnet18alphaattmaps/ILSVRC2012_val_00011053_attmap_28x28_no_blur_resnet18_simpool_alpha1.0_ep100.PNG} &
%\fig[\sz]{resnet18alphaattmaps/ILSVRC2012_val_00011053_attmap_28x28_no_blur_resnet18_simpool_alpha1.25_ep100.PNG} &
%\fig[\sz]{resnet18alphaattmaps/ILSVRC2012_val_00011053_attmap_28x28_no_blur_resnet18_simpool_alpha1.5_ep100.PNG} &
%\fig[\sz]{resnet18alphaattmaps/ILSVRC2012_val_00011053_attmap_28x28_no_blur_resnet18_simpool_alpha2.0_ep100.PNG} &
%\fig[\sz]{resnet18alphaattmaps/ILSVRC2012_val_00011053_attmap_28x28_no_blur_resnet18_simpool_alpha2.5_ep100.PNG} &
%\fig[\sz]{resnet18alphaattmaps/ILSVRC2012_val_00011053_attmap_28x28_no_blur_resnet18_simpool_alpha3.0_ep100.PNG} & 
%\fig[\sz]{resnet18alphaattmaps/ILSVRC2012_val_00011053_attmap_28x28_no_blur_resnet18_simpool_alpha5.0_ep100.PNG} \\

\fig[\sz]{resnet18alphaattmaps/ILSVRC2012_val_00007729_orig.PNG} &
\fig[\sz]{resnet18alphaattmaps/ILSVRC2012_val_00007729_attmap_28x28_no_blur_resnet18_simpool_no_alpha_ep100.PNG} &
\fig[\sz]{resnet18alphaattmaps/ILSVRC2012_val_00007729_attmap_28x28_no_blur_resnet18_simpool_alpha0.5_ep100.PNG} &
\fig[\sz]{resnet18alphaattmaps/ILSVRC2012_val_00007729_attmap_28x28_no_blur_resnet18_simpool_alpha1.0_ep100.PNG} &
\fig[\sz]{resnet18alphaattmaps/ILSVRC2012_val_00007729_attmap_28x28_no_blur_resnet18_simpool_alpha1.25_ep100.PNG} &
\fig[\sz]{resnet18alphaattmaps/ILSVRC2012_val_00007729_attmap_28x28_no_blur_resnet18_simpool_alpha1.5_ep100.PNG} &
\fig[\sz]{resnet18alphaattmaps/ILSVRC2012_val_00007729_attmap_28x28_no_blur_resnet18_simpool_alpha2.0_ep100.PNG} &
\fig[\sz]{resnet18alphaattmaps/ILSVRC2012_val_00007729_attmap_28x28_no_blur_resnet18_simpool_alpha2.5_ep100.PNG} &
\fig[\sz]{resnet18alphaattmaps/ILSVRC2012_val_00007729_attmap_28x28_no_blur_resnet18_simpool_alpha3.0_ep100.PNG} & 
\fig[\sz]{resnet18alphaattmaps/ILSVRC2012_val_00007729_attmap_28x28_no_blur_resnet18_simpool_alpha5.0_ep100.PNG} \\

\fig[\sz]{resnet18alphaattmaps/ILSVRC2012_val_00011278_orig.PNG} &
\fig[\sz]{resnet18alphaattmaps/ILSVRC2012_val_00011278_attmap_28x28_no_blur_resnet18_simpool_no_alpha_ep100.PNG} &
\fig[\sz]{resnet18alphaattmaps/ILSVRC2012_val_00011278_attmap_28x28_no_blur_resnet18_simpool_alpha0.5_ep100.PNG} &
\fig[\sz]{resnet18alphaattmaps/ILSVRC2012_val_00011278_attmap_28x28_no_blur_resnet18_simpool_alpha1.0_ep100.PNG} &
\fig[\sz]{resnet18alphaattmaps/ILSVRC2012_val_00011278_attmap_28x28_no_blur_resnet18_simpool_alpha1.25_ep100.PNG} &
\fig[\sz]{resnet18alphaattmaps/ILSVRC2012_val_00011278_attmap_28x28_no_blur_resnet18_simpool_alpha1.5_ep100.PNG} &
\fig[\sz]{resnet18alphaattmaps/ILSVRC2012_val_00011278_attmap_28x28_no_blur_resnet18_simpool_alpha2.0_ep100.PNG} &
\fig[\sz]{resnet18alphaattmaps/ILSVRC2012_val_00011278_attmap_28x28_no_blur_resnet18_simpool_alpha2.5_ep100.PNG} &
\fig[\sz]{resnet18alphaattmaps/ILSVRC2012_val_00011278_attmap_28x28_no_blur_resnet18_simpool_alpha3.0_ep100.PNG} & 
\fig[\sz]{resnet18alphaattmaps/ILSVRC2012_val_00011278_attmap_28x28_no_blur_resnet18_simpool_alpha5.0_ep100.PNG} \\

\fig[\sz]{resnet18alphaattmaps/ILSVRC2012_val_00013407_orig.PNG} &
\fig[\sz]{resnet18alphaattmaps/ILSVRC2012_val_00013407_attmap_28x28_no_blur_resnet18_simpool_no_alpha_ep100.PNG} &
\fig[\sz]{resnet18alphaattmaps/ILSVRC2012_val_00013407_attmap_28x28_no_blur_resnet18_simpool_alpha0.5_ep100.PNG} &
\fig[\sz]{resnet18alphaattmaps/ILSVRC2012_val_00013407_attmap_28x28_no_blur_resnet18_simpool_alpha1.0_ep100.PNG} &
\fig[\sz]{resnet18alphaattmaps/ILSVRC2012_val_00013407_attmap_28x28_no_blur_resnet18_simpool_alpha1.25_ep100.PNG} &
\fig[\sz]{resnet18alphaattmaps/ILSVRC2012_val_00013407_attmap_28x28_no_blur_resnet18_simpool_alpha1.5_ep100.PNG} &
\fig[\sz]{resnet18alphaattmaps/ILSVRC2012_val_00013407_attmap_28x28_no_blur_resnet18_simpool_alpha2.0_ep100.PNG} &
\fig[\sz]{resnet18alphaattmaps/ILSVRC2012_val_00013407_attmap_28x28_no_blur_resnet18_simpool_alpha2.5_ep100.PNG} &
\fig[\sz]{resnet18alphaattmaps/ILSVRC2012_val_00013407_attmap_28x28_no_blur_resnet18_simpool_alpha3.0_ep100.PNG} & 
\fig[\sz]{resnet18alphaattmaps/ILSVRC2012_val_00013407_attmap_28x28_no_blur_resnet18_simpool_alpha5.0_ep100.PNG} \\

\fig[\sz]{resnet18alphaattmaps/ILSVRC2012_val_00013513_orig.PNG} &
\fig[\sz]{resnet18alphaattmaps/ILSVRC2012_val_00013513_attmap_28x28_no_blur_resnet18_simpool_no_alpha_ep100.PNG} &
\fig[\sz]{resnet18alphaattmaps/ILSVRC2012_val_00013513_attmap_28x28_no_blur_resnet18_simpool_alpha0.5_ep100.PNG} &
\fig[\sz]{resnet18alphaattmaps/ILSVRC2012_val_00013513_attmap_28x28_no_blur_resnet18_simpool_alpha1.0_ep100.PNG} &
\fig[\sz]{resnet18alphaattmaps/ILSVRC2012_val_00013513_attmap_28x28_no_blur_resnet18_simpool_alpha1.25_ep100.PNG} &
\fig[\sz]{resnet18alphaattmaps/ILSVRC2012_val_00013513_attmap_28x28_no_blur_resnet18_simpool_alpha1.5_ep100.PNG} &
\fig[\sz]{resnet18alphaattmaps/ILSVRC2012_val_00013513_attmap_28x28_no_blur_resnet18_simpool_alpha2.0_ep100.PNG} &
\fig[\sz]{resnet18alphaattmaps/ILSVRC2012_val_00013513_attmap_28x28_no_blur_resnet18_simpool_alpha2.5_ep100.PNG} &
\fig[\sz]{resnet18alphaattmaps/ILSVRC2012_val_00013513_attmap_28x28_no_blur_resnet18_simpool_alpha3.0_ep100.PNG} & 
\fig[\sz]{resnet18alphaattmaps/ILSVRC2012_val_00013513_attmap_28x28_no_blur_resnet18_simpool_alpha5.0_ep100.PNG} \\

\fig[\sz]{resnet18alphaattmaps/ILSVRC2012_val_00014227_orig.PNG} &
\fig[\sz]{resnet18alphaattmaps/ILSVRC2012_val_00014227_attmap_28x28_no_blur_resnet18_simpool_no_alpha_ep100.PNG} &
\fig[\sz]{resnet18alphaattmaps/ILSVRC2012_val_00014227_attmap_28x28_no_blur_resnet18_simpool_alpha0.5_ep100.PNG} &
\fig[\sz]{resnet18alphaattmaps/ILSVRC2012_val_00014227_attmap_28x28_no_blur_resnet18_simpool_alpha1.0_ep100.PNG} &
\fig[\sz]{resnet18alphaattmaps/ILSVRC2012_val_00014227_attmap_28x28_no_blur_resnet18_simpool_alpha1.25_ep100.PNG} &
\fig[\sz]{resnet18alphaattmaps/ILSVRC2012_val_00014227_attmap_28x28_no_blur_resnet18_simpool_alpha1.5_ep100.PNG} &
\fig[\sz]{resnet18alphaattmaps/ILSVRC2012_val_00014227_attmap_28x28_no_blur_resnet18_simpool_alpha2.0_ep100.PNG} &
\fig[\sz]{resnet18alphaattmaps/ILSVRC2012_val_00014227_attmap_28x28_no_blur_resnet18_simpool_alpha2.5_ep100.PNG} &
\fig[\sz]{resnet18alphaattmaps/ILSVRC2012_val_00014227_attmap_28x28_no_blur_resnet18_simpool_alpha3.0_ep100.PNG} & 
\fig[\sz]{resnet18alphaattmaps/ILSVRC2012_val_00014227_attmap_28x28_no_blur_resnet18_simpool_alpha5.0_ep100.PNG} \\

\fig[\sz]{resnet18alphaattmaps/ILSVRC2012_val_00016030_orig.PNG} &
\fig[\sz]{resnet18alphaattmaps/ILSVRC2012_val_00016030_attmap_28x28_no_blur_resnet18_simpool_no_alpha_ep100.PNG} &
\fig[\sz]{resnet18alphaattmaps/ILSVRC2012_val_00016030_attmap_28x28_no_blur_resnet18_simpool_alpha0.5_ep100.PNG} &
\fig[\sz]{resnet18alphaattmaps/ILSVRC2012_val_00016030_attmap_28x28_no_blur_resnet18_simpool_alpha1.0_ep100.PNG} &
\fig[\sz]{resnet18alphaattmaps/ILSVRC2012_val_00016030_attmap_28x28_no_blur_resnet18_simpool_alpha1.25_ep100.PNG} &
\fig[\sz]{resnet18alphaattmaps/ILSVRC2012_val_00016030_attmap_28x28_no_blur_resnet18_simpool_alpha1.5_ep100.PNG} &
\fig[\sz]{resnet18alphaattmaps/ILSVRC2012_val_00016030_attmap_28x28_no_blur_resnet18_simpool_alpha2.0_ep100.PNG} &
\fig[\sz]{resnet18alphaattmaps/ILSVRC2012_val_00016030_attmap_28x28_no_blur_resnet18_simpool_alpha2.5_ep100.PNG} &
\fig[\sz]{resnet18alphaattmaps/ILSVRC2012_val_00016030_attmap_28x28_no_blur_resnet18_simpool_alpha3.0_ep100.PNG} & 
\fig[\sz]{resnet18alphaattmaps/ILSVRC2012_val_00016030_attmap_28x28_no_blur_resnet18_simpool_alpha5.0_ep100.PNG} \\

\fig[\sz]{resnet18alphaattmaps/ILSVRC2012_val_00039106_orig.PNG} &
\fig[\sz]{resnet18alphaattmaps/ILSVRC2012_val_00039106_attmap_28x28_no_blur_resnet18_simpool_no_alpha_ep100.PNG} &
\fig[\sz]{resnet18alphaattmaps/ILSVRC2012_val_00039106_attmap_28x28_no_blur_resnet18_simpool_alpha0.5_ep100.PNG} &
\fig[\sz]{resnet18alphaattmaps/ILSVRC2012_val_00039106_attmap_28x28_no_blur_resnet18_simpool_alpha1.0_ep100.PNG} &
\fig[\sz]{resnet18alphaattmaps/ILSVRC2012_val_00039106_attmap_28x28_no_blur_resnet18_simpool_alpha1.25_ep100.PNG} &
\fig[\sz]{resnet18alphaattmaps/ILSVRC2012_val_00039106_attmap_28x28_no_blur_resnet18_simpool_alpha1.5_ep100.PNG} &
\fig[\sz]{resnet18alphaattmaps/ILSVRC2012_val_00039106_attmap_28x28_no_blur_resnet18_simpool_alpha2.0_ep100.PNG} &
\fig[\sz]{resnet18alphaattmaps/ILSVRC2012_val_00039106_attmap_28x28_no_blur_resnet18_simpool_alpha2.5_ep100.PNG} &
\fig[\sz]{resnet18alphaattmaps/ILSVRC2012_val_00039106_attmap_28x28_no_blur_resnet18_simpool_alpha3.0_ep100.PNG} & 
\fig[\sz]{resnet18alphaattmaps/ILSVRC2012_val_00039106_attmap_28x28_no_blur_resnet18_simpool_alpha5.0_ep100.PNG} \\

\fig[\sz]{resnet18alphaattmaps/ILSVRC2012_val_00046106_orig.PNG} &
\fig[\sz]{resnet18alphaattmaps/ILSVRC2012_val_00046106_attmap_28x28_no_blur_resnet18_simpool_no_alpha_ep100.PNG} &
\fig[\sz]{resnet18alphaattmaps/ILSVRC2012_val_00046106_attmap_28x28_no_blur_resnet18_simpool_alpha0.5_ep100.PNG} &
\fig[\sz]{resnet18alphaattmaps/ILSVRC2012_val_00046106_attmap_28x28_no_blur_resnet18_simpool_alpha1.0_ep100.PNG} &
\fig[\sz]{resnet18alphaattmaps/ILSVRC2012_val_00046106_attmap_28x28_no_blur_resnet18_simpool_alpha1.25_ep100.PNG} &
\fig[\sz]{resnet18alphaattmaps/ILSVRC2012_val_00046106_attmap_28x28_no_blur_resnet18_simpool_alpha1.5_ep100.PNG} &
\fig[\sz]{resnet18alphaattmaps/ILSVRC2012_val_00046106_attmap_28x28_no_blur_resnet18_simpool_alpha2.0_ep100.PNG} &
\fig[\sz]{resnet18alphaattmaps/ILSVRC2012_val_00046106_attmap_28x28_no_blur_resnet18_simpool_alpha2.5_ep100.PNG} &
\fig[\sz]{resnet18alphaattmaps/ILSVRC2012_val_00046106_attmap_28x28_no_blur_resnet18_simpool_alpha3.0_ep100.PNG} & 
\fig[\sz]{resnet18alphaattmaps/ILSVRC2012_val_00046106_attmap_28x28_no_blur_resnet18_simpool_alpha5.0_ep100.PNG} \\

\fig[\sz]{resnet18alphaattmaps/ILSVRC2012_val_00047272_orig.PNG} &
\fig[\sz]{resnet18alphaattmaps/ILSVRC2012_val_00047272_attmap_28x28_no_blur_resnet18_simpool_no_alpha_ep100.PNG} &
\fig[\sz]{resnet18alphaattmaps/ILSVRC2012_val_00047272_attmap_28x28_no_blur_resnet18_simpool_alpha0.5_ep100.PNG} &
\fig[\sz]{resnet18alphaattmaps/ILSVRC2012_val_00047272_attmap_28x28_no_blur_resnet18_simpool_alpha1.0_ep100.PNG} &
\fig[\sz]{resnet18alphaattmaps/ILSVRC2012_val_00047272_attmap_28x28_no_blur_resnet18_simpool_alpha1.25_ep100.PNG} &
\fig[\sz]{resnet18alphaattmaps/ILSVRC2012_val_00047272_attmap_28x28_no_blur_resnet18_simpool_alpha1.5_ep100.PNG} &
\fig[\sz]{resnet18alphaattmaps/ILSVRC2012_val_00047272_attmap_28x28_no_blur_resnet18_simpool_alpha2.0_ep100.PNG} &
\fig[\sz]{resnet18alphaattmaps/ILSVRC2012_val_00047272_attmap_28x28_no_blur_resnet18_simpool_alpha2.5_ep100.PNG} &
\fig[\sz]{resnet18alphaattmaps/ILSVRC2012_val_00047272_attmap_28x28_no_blur_resnet18_simpool_alpha3.0_ep100.PNG} & 
\fig[\sz]{resnet18alphaattmaps/ILSVRC2012_val_00047272_attmap_28x28_no_blur_resnet18_simpool_alpha5.0_ep100.PNG} \\

input &
\mr{2}{no $\gamma$} &
\mr{2}{$\gamma = 0.5$} &
\mr{2}{$\gamma = 1.0$} &
\mr{2}{$\gamma = 1.25$} &
\mr{2}{$\gamma = 1.5$} &
\mr{2}{$\gamma = 2.0$} &
\mr{2}{$\gamma = 2.5$} &
\mr{2}{$\gamma = 3.0$} &
\mr{2}{$\gamma = 5.0$} \\

image &
&
&
&
&
&
&
&
&
\\

\end{tabular}
\vspace{3pt}
\caption{\emph{The effect of $\gamma$}. Attention maps of ResNet-18~\cite{resnet} with \Ours using different values of $\gamma$ trained on ImageNet-20\% for 100 epochs under supervision. We use the attention map $\va$~\eq{sp-att}. Input image resolution: $896 \times 896$; output attention map: $28 \times 28$; no $\gamma$: using the average pooling operation $f_{-1}$ instead of $f_\alpha$~\eq{alpha}. We set $\gamma=2$ by default for convolutional networks.}
\label{fig:attention-maps-resnet-alpha}
\end{figure*}

\begin{figure*}
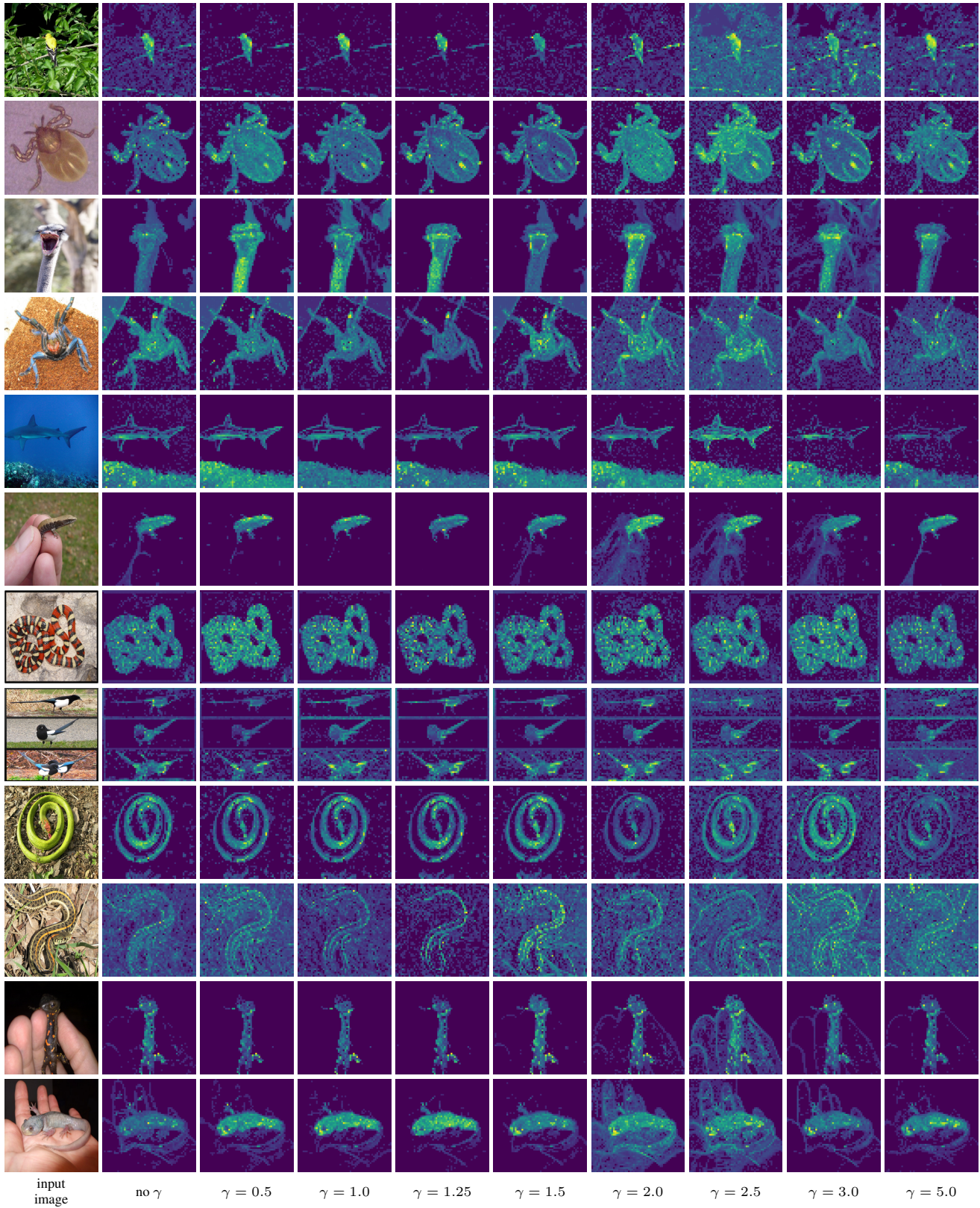

\scriptsize
\centering
\setlength{\tabcolsep}{1.2pt}
\newcommand{\sz}{.095}
\begin{tabular}{cccccccccc}

\fig[\sz]{vittalphaattmaps/ILSVRC2012_val_00000570_orig.PNG} &
\fig[\sz]{vittalphaattmaps/ILSVRC2012_val_00000570_attmap_vitt_gap_cls_supervised.PNG} &
\fig[\sz]{vittalphaattmaps/ILSVRC2012_val_00000570_attmap_vitt_gem0.5_supervised.PNG} &
\fig[\sz]{vittalphaattmaps/ILSVRC2012_val_00000570_attmap_vitt_gem1.0_supervised.PNG} &
\fig[\sz]{vittalphaattmaps/ILSVRC2012_val_00000570_attmap_vitt_gem1.25_supervised.PNG} &
\fig[\sz]{vittalphaattmaps/ILSVRC2012_val_00000570_attmap_vitt_gem1.5_supervised.PNG} &
\fig[\sz]{vittalphaattmaps/ILSVRC2012_val_00000570_attmap_vitt_gem2.0_supervised.PNG} &
\fig[\sz]{vittalphaattmaps/ILSVRC2012_val_00000570_attmap_vitt_gem2.5_supervised.PNG} &
\fig[\sz]{vittalphaattmaps/ILSVRC2012_val_00000570_attmap_vitt_gem3.0_supervised.PNG} & 
\fig[\sz]{vittalphaattmaps/ILSVRC2012_val_00000570_attmap_vitt_gem5.0_supervised.PNG} \\

\fig[\sz]{vittalphaattmaps/ILSVRC2012_val_00020115_orig.PNG} &
\fig[\sz]{vittalphaattmaps/ILSVRC2012_val_00020115_attmap_vitt_gap_cls_supervised.PNG} &
\fig[\sz]{vittalphaattmaps/ILSVRC2012_val_00020115_attmap_vitt_gem0.5_supervised.PNG} &
\fig[\sz]{vittalphaattmaps/ILSVRC2012_val_00020115_attmap_vitt_gem1.0_supervised.PNG} &
\fig[\sz]{vittalphaattmaps/ILSVRC2012_val_00020115_attmap_vitt_gem1.25_supervised.PNG} &
\fig[\sz]{vittalphaattmaps/ILSVRC2012_val_00020115_attmap_vitt_gem1.5_supervised.PNG} &
\fig[\sz]{vittalphaattmaps/ILSVRC2012_val_00020115_attmap_vitt_gem2.0_supervised.PNG} &
\fig[\sz]{vittalphaattmaps/ILSVRC2012_val_00020115_attmap_vitt_gem2.5_supervised.PNG} &
\fig[\sz]{vittalphaattmaps/ILSVRC2012_val_00020115_attmap_vitt_gem3.0_supervised.PNG} & 
\fig[\sz]{vittalphaattmaps/ILSVRC2012_val_00020115_attmap_vitt_gem5.0_supervised.PNG} \\

\fig[\sz]{vittalphaattmaps/ILSVRC2012_val_00023214_orig.PNG} &
\fig[\sz]{vittalphaattmaps/ILSVRC2012_val_00023214_attmap_vitt_gap_cls_supervised.PNG} &
\fig[\sz]{vittalphaattmaps/ILSVRC2012_val_00023214_attmap_vitt_gem0.5_supervised.PNG} &
\fig[\sz]{vittalphaattmaps/ILSVRC2012_val_00023214_attmap_vitt_gem1.0_supervised.PNG} &
\fig[\sz]{vittalphaattmaps/ILSVRC2012_val_00023214_attmap_vitt_gem1.25_supervised.PNG} &
\fig[\sz]{vittalphaattmaps/ILSVRC2012_val_00023214_attmap_vitt_gem1.5_supervised.PNG} &
\fig[\sz]{vittalphaattmaps/ILSVRC2012_val_00023214_attmap_vitt_gem2.0_supervised.PNG} &
\fig[\sz]{vittalphaattmaps/ILSVRC2012_val_00023214_attmap_vitt_gem2.5_supervised.PNG} &
\fig[\sz]{vittalphaattmaps/ILSVRC2012_val_00023214_attmap_vitt_gem3.0_supervised.PNG} & 
\fig[\sz]{vittalphaattmaps/ILSVRC2012_val_00023214_attmap_vitt_gem5.0_supervised.PNG} \\

\fig[\sz]{vittalphaattmaps/ILSVRC2012_val_00035994_orig.PNG} &
\fig[\sz]{vittalphaattmaps/ILSVRC2012_val_00035994_attmap_vitt_gap_cls_supervised.PNG} &
\fig[\sz]{vittalphaattmaps/ILSVRC2012_val_00035994_attmap_vitt_gem0.5_supervised.PNG} &
\fig[\sz]{vittalphaattmaps/ILSVRC2012_val_00035994_attmap_vitt_gem1.0_supervised.PNG} &
\fig[\sz]{vittalphaattmaps/ILSVRC2012_val_00035994_attmap_vitt_gem1.25_supervised.PNG} &
\fig[\sz]{vittalphaattmaps/ILSVRC2012_val_00035994_attmap_vitt_gem1.5_supervised.PNG} &
\fig[\sz]{vittalphaattmaps/ILSVRC2012_val_00035994_attmap_vitt_gem2.0_supervised.PNG} &
\fig[\sz]{vittalphaattmaps/ILSVRC2012_val_00035994_attmap_vitt_gem2.5_supervised.PNG} &
\fig[\sz]{vittalphaattmaps/ILSVRC2012_val_00035994_attmap_vitt_gem3.0_supervised.PNG} & 
\fig[\sz]{vittalphaattmaps/ILSVRC2012_val_00035994_attmap_vitt_gem5.0_supervised.PNG} \\

%\fig[\sz]{vittalphaattmaps/ILSVRC2012_val_00044676_orig.PNG} &
%\fig[\sz]{vittalphaattmaps/ILSVRC2012_val_00044676_attmap_vitt_gap_cls_supervised.PNG} &
%\fig[\sz]{vittalphaattmaps/ILSVRC2012_val_00044676_attmap_vitt_gem0.5_supervised.PNG} &
%\fig[\sz]{vittalphaattmaps/ILSVRC2012_val_00044676_attmap_vitt_gem1.0_supervised.PNG} &
%\fig[\sz]{vittalphaattmaps/ILSVRC2012_val_00044676_attmap_vitt_gem1.25_supervised.PNG} &
%\fig[\sz]{vittalphaattmaps/ILSVRC2012_val_00044676_attmap_vitt_gem1.5_supervised.PNG} &
%\fig[\sz]{vittalphaattmaps/ILSVRC2012_val_00044676_attmap_vitt_gem2.0_supervised.PNG} &
%\fig[\sz]{vittalphaattmaps/ILSVRC2012_val_00044676_attmap_vitt_gem2.5_supervised.PNG} &
%\fig[\sz]{vittalphaattmaps/ILSVRC2012_val_00044676_attmap_vitt_gem3.0_supervised.PNG} & 
%\fig[\sz]{vittalphaattmaps/ILSVRC2012_val_00044676_attmap_vitt_gem5.0_supervised.PNG} \\

\fig[\sz]{vittalphaattmaps/ILSVRC2012_val_00049166_orig.PNG} &
\fig[\sz]{vittalphaattmaps/ILSVRC2012_val_00049166_attmap_vitt_gap_cls_supervised.PNG} &
\fig[\sz]{vittalphaattmaps/ILSVRC2012_val_00049166_attmap_vitt_gem0.5_supervised.PNG} &
\fig[\sz]{vittalphaattmaps/ILSVRC2012_val_00049166_attmap_vitt_gem1.0_supervised.PNG} &
\fig[\sz]{vittalphaattmaps/ILSVRC2012_val_00049166_attmap_vitt_gem1.25_supervised.PNG} &
\fig[\sz]{vittalphaattmaps/ILSVRC2012_val_00049166_attmap_vitt_gem1.5_supervised.PNG} &
\fig[\sz]{vittalphaattmaps/ILSVRC2012_val_00049166_attmap_vitt_gem2.0_supervised.PNG} &
\fig[\sz]{vittalphaattmaps/ILSVRC2012_val_00049166_attmap_vitt_gem2.5_supervised.PNG} &
\fig[\sz]{vittalphaattmaps/ILSVRC2012_val_00049166_attmap_vitt_gem3.0_supervised.PNG} & 
\fig[\sz]{vittalphaattmaps/ILSVRC2012_val_00049166_attmap_vitt_gem5.0_supervised.PNG} \\

\fig[\sz]{vittalphaattmaps/ILSVRC2012_val_00049635_orig.PNG} &
\fig[\sz]{vittalphaattmaps/ILSVRC2012_val_00049635_attmap_vitt_gap_cls_supervised.PNG} &
\fig[\sz]{vittalphaattmaps/ILSVRC2012_val_00049635_attmap_vitt_gem0.5_supervised.PNG} &
\fig[\sz]{vittalphaattmaps/ILSVRC2012_val_00049635_attmap_vitt_gem1.0_supervised.PNG} &
\fig[\sz]{vittalphaattmaps/ILSVRC2012_val_00049635_attmap_vitt_gem1.25_supervised.PNG} &
\fig[\sz]{vittalphaattmaps/ILSVRC2012_val_00049635_attmap_vitt_gem1.5_supervised.PNG} &
\fig[\sz]{vittalphaattmaps/ILSVRC2012_val_00049635_attmap_vitt_gem2.0_supervised.PNG} &
\fig[\sz]{vittalphaattmaps/ILSVRC2012_val_00049635_attmap_vitt_gem2.5_supervised.PNG} &
\fig[\sz]{vittalphaattmaps/ILSVRC2012_val_00049635_attmap_vitt_gem3.0_supervised.PNG} & 
\fig[\sz]{vittalphaattmaps/ILSVRC2012_val_00049635_attmap_vitt_gem5.0_supervised.PNG} \\

\fig[\sz]{vittalphaattmaps/ILSVRC2012_val_00005676_orig.PNG} &
\fig[\sz]{vittalphaattmaps/ILSVRC2012_val_00005676_attmap_vitt_gap_cls_supervised.PNG} &
\fig[\sz]{vittalphaattmaps/ILSVRC2012_val_00005676_attmap_vitt_gem0.5_supervised.PNG} &
\fig[\sz]{vittalphaattmaps/ILSVRC2012_val_00005676_attmap_vitt_gem1.0_supervised.PNG} &
\fig[\sz]{vittalphaattmaps/ILSVRC2012_val_00005676_attmap_vitt_gem1.25_supervised.PNG} &
\fig[\sz]{vittalphaattmaps/ILSVRC2012_val_00005676_attmap_vitt_gem1.5_supervised.PNG} &
\fig[\sz]{vittalphaattmaps/ILSVRC2012_val_00005676_attmap_vitt_gem2.0_supervised.PNG} &
\fig[\sz]{vittalphaattmaps/ILSVRC2012_val_00005676_attmap_vitt_gem2.5_supervised.PNG} &
\fig[\sz]{vittalphaattmaps/ILSVRC2012_val_00005676_attmap_vitt_gem3.0_supervised.PNG} & 
\fig[\sz]{vittalphaattmaps/ILSVRC2012_val_00005676_attmap_vitt_gem5.0_supervised.PNG} \\

\fig[\sz]{vittalphaattmaps/ILSVRC2012_val_00008347_orig.PNG} &
\fig[\sz]{vittalphaattmaps/ILSVRC2012_val_00008347_attmap_vitt_gap_cls_supervised.PNG} &
\fig[\sz]{vittalphaattmaps/ILSVRC2012_val_00008347_attmap_vitt_gem0.5_supervised.PNG} &
\fig[\sz]{vittalphaattmaps/ILSVRC2012_val_00008347_attmap_vitt_gem1.0_supervised.PNG} &
\fig[\sz]{vittalphaattmaps/ILSVRC2012_val_00008347_attmap_vitt_gem1.25_supervised.PNG} &
\fig[\sz]{vittalphaattmaps/ILSVRC2012_val_00008347_attmap_vitt_gem1.5_supervised.PNG} &
\fig[\sz]{vittalphaattmaps/ILSVRC2012_val_00008347_attmap_vitt_gem2.0_supervised.PNG} &
\fig[\sz]{vittalphaattmaps/ILSVRC2012_val_00008347_attmap_vitt_gem2.5_supervised.PNG} &
\fig[\sz]{vittalphaattmaps/ILSVRC2012_val_00008347_attmap_vitt_gem3.0_supervised.PNG} & 
\fig[\sz]{vittalphaattmaps/ILSVRC2012_val_00008347_attmap_vitt_gem5.0_supervised.PNG} \\

\fig[\sz]{vittalphaattmaps/ILSVRC2012_val_00008025_orig.PNG} &
\fig[\sz]{vittalphaattmaps/ILSVRC2012_val_00008025_attmap_vitt_gap_cls_supervised.PNG} &
\fig[\sz]{vittalphaattmaps/ILSVRC2012_val_00008025_attmap_vitt_gem0.5_supervised.PNG} &
\fig[\sz]{vittalphaattmaps/ILSVRC2012_val_00008025_attmap_vitt_gem1.0_supervised.PNG} &
\fig[\sz]{vittalphaattmaps/ILSVRC2012_val_00008025_attmap_vitt_gem1.25_supervised.PNG} &
\fig[\sz]{vittalphaattmaps/ILSVRC2012_val_00008025_attmap_vitt_gem1.5_supervised.PNG} &
\fig[\sz]{vittalphaattmaps/ILSVRC2012_val_00008025_attmap_vitt_gem2.0_supervised.PNG} &
\fig[\sz]{vittalphaattmaps/ILSVRC2012_val_00008025_attmap_vitt_gem2.5_supervised.PNG} &
\fig[\sz]{vittalphaattmaps/ILSVRC2012_val_00008025_attmap_vitt_gem3.0_supervised.PNG} & 
\fig[\sz]{vittalphaattmaps/ILSVRC2012_val_00008025_attmap_vitt_gem5.0_supervised.PNG} \\

\fig[\sz]{vittalphaattmaps/ILSVRC2012_val_00008130_orig.PNG} &
\fig[\sz]{vittalphaattmaps/ILSVRC2012_val_00008130_attmap_vitt_gap_cls_supervised.PNG} &
\fig[\sz]{vittalphaattmaps/ILSVRC2012_val_00008130_attmap_vitt_gem0.5_supervised.PNG} &
\fig[\sz]{vittalphaattmaps/ILSVRC2012_val_00008130_attmap_vitt_gem1.0_supervised.PNG} &
\fig[\sz]{vittalphaattmaps/ILSVRC2012_val_00008130_attmap_vitt_gem1.25_supervised.PNG} &
\fig[\sz]{vittalphaattmaps/ILSVRC2012_val_00008130_attmap_vitt_gem1.5_supervised.PNG} &
\fig[\sz]{vittalphaattmaps/ILSVRC2012_val_00008130_attmap_vitt_gem2.0_supervised.PNG} &
\fig[\sz]{vittalphaattmaps/ILSVRC2012_val_00008130_attmap_vitt_gem2.5_supervised.PNG} &
\fig[\sz]{vittalphaattmaps/ILSVRC2012_val_00008130_attmap_vitt_gem3.0_supervised.PNG} & 
\fig[\sz]{vittalphaattmaps/ILSVRC2012_val_00008130_attmap_vitt_gem5.0_supervised.PNG} \\

\fig[\sz]{vittalphaattmaps/ILSVRC2012_val_00009779_orig.PNG} &
\fig[\sz]{vittalphaattmaps/ILSVRC2012_val_00009779_attmap_vitt_gap_cls_supervised.PNG} &
\fig[\sz]{vittalphaattmaps/ILSVRC2012_val_00009779_attmap_vitt_gem0.5_supervised.PNG} &
\fig[\sz]{vittalphaattmaps/ILSVRC2012_val_00009779_attmap_vitt_gem1.0_supervised.PNG} &
\fig[\sz]{vittalphaattmaps/ILSVRC2012_val_00009779_attmap_vitt_gem1.25_supervised.PNG} &
\fig[\sz]{vittalphaattmaps/ILSVRC2012_val_00009779_attmap_vitt_gem1.5_supervised.PNG} &
\fig[\sz]{vittalphaattmaps/ILSVRC2012_val_00009779_attmap_vitt_gem2.0_supervised.PNG} &
\fig[\sz]{vittalphaattmaps/ILSVRC2012_val_00009779_attmap_vitt_gem2.5_supervised.PNG} &
\fig[\sz]{vittalphaattmaps/ILSVRC2012_val_00009779_attmap_vitt_gem3.0_supervised.PNG} & 
\fig[\sz]{vittalphaattmaps/ILSVRC2012_val_00009779_attmap_vitt_gem5.0_supervised.PNG} \\

%\fig[\sz]{vittalphaattmaps/ILSVRC2012_val_00012354_orig.PNG} &
%\fig[\sz]{vittalphaattmaps/ILSVRC2012_val_00012354_attmap_vitt_gap_cls_supervised.PNG} &
%\fig[\sz]{vittalphaattmaps/ILSVRC2012_val_00012354_attmap_vitt_gem0.5_supervised.PNG} &
%\fig[\sz]{vittalphaattmaps/ILSVRC2012_val_00012354_attmap_vitt_gem1.0_supervised.PNG} &
%\fig[\sz]{vittalphaattmaps/ILSVRC2012_val_00012354_attmap_vitt_gem1.25_supervised.PNG} &
%\fig[\sz]{vittalphaattmaps/ILSVRC2012_val_00012354_attmap_vitt_gem1.5_supervised.PNG} &
%\fig[\sz]{vittalphaattmaps/ILSVRC2012_val_00012354_attmap_vitt_gem2.0_supervised.PNG} &
%\fig[\sz]{vittalphaattmaps/ILSVRC2012_val_00012354_attmap_vitt_gem2.5_supervised.PNG} &
%\fig[\sz]{vittalphaattmaps/ILSVRC2012_val_00012354_attmap_vitt_gem3.0_supervised.PNG} & 
%\fig[\sz]{vittalphaattmaps/ILSVRC2012_val_00012354_attmap_vitt_gem5.0_supervised.PNG} \\

\fig[\sz]{vittalphaattmaps/ILSVRC2012_val_00003189_orig.PNG} &
\fig[\sz]{vittalphaattmaps/ILSVRC2012_val_00003189_attmap_vitt_gap_cls_supervised.PNG} &
\fig[\sz]{vittalphaattmaps/ILSVRC2012_val_00003189_attmap_vitt_gem0.5_supervised.PNG} &
\fig[\sz]{vittalphaattmaps/ILSVRC2012_val_00003189_attmap_vitt_gem1.0_supervised.PNG} &
\fig[\sz]{vittalphaattmaps/ILSVRC2012_val_00003189_attmap_vitt_gem1.25_supervised.PNG} &
\fig[\sz]{vittalphaattmaps/ILSVRC2012_val_00003189_attmap_vitt_gem1.5_supervised.PNG} &
\fig[\sz]{vittalphaattmaps/ILSVRC2012_val_00003189_attmap_vitt_gem2.0_supervised.PNG} &
\fig[\sz]{vittalphaattmaps/ILSVRC2012_val_00003189_attmap_vitt_gem2.5_supervised.PNG} &
\fig[\sz]{vittalphaattmaps/ILSVRC2012_val_00003189_attmap_vitt_gem3.0_supervised.PNG} & 
\fig[\sz]{vittalphaattmaps/ILSVRC2012_val_00003189_attmap_vitt_gem5.0_supervised.PNG} \\

input &
\mr{2}{no $\gamma$} &
\mr{2}{$\gamma = 0.5$} &
\mr{2}{$\gamma = 1.0$} &
\mr{2}{$\gamma = 1.25$} &
\mr{2}{$\gamma = 1.5$} &
\mr{2}{$\gamma = 2.0$} &
\mr{2}{$\gamma = 2.5$} &
\mr{2}{$\gamma = 3.0$} &
\mr{2}{$\gamma = 5.0$} \\

image &
&
&
&
&
&
&
&
&
\\

\end{tabular}
\vspace{3pt}
\caption{\emph{The effect of $\gamma$}. Attention maps of ViT-T~\cite{vit} with \Ours using different values of $\gamma$ trained on \imagenet for 100 epochs under supervision. We use the attention map $\va$~\eq{sp-att}. Input image resolution: $896 \times 896$; patches: $16 \times 16$; output attention map: $56 \times 56$; no $\gamma$: using the average pooling operation $f_{-1}$ instead of $f_\alpha$~\eq{alpha}. We set $\gamma=1.25$ by default for transformers.}
\label{fig:attention-maps-vit-alpha}
\end{figure*}

\autoref{fig:attention-maps-resnet-alpha} and \autoref{fig:attention-maps-vit-alpha} visualize the effect of exponent $\gamma = (1 - \alpha)/2$ of pooling operation $f_\alpha$~\eq{alpha} on the quality of the attention maps of ResNet-18 and ViT-T, respectively. The use of the average pooling operation $f_{-1}$ as opposed to $f_\alpha$~\eq{alpha} is referred to as no $\gamma$. For ResNet-18, we observe that for $\gamma < 1.25$ or $\gamma > 3.0$, the attention maps are of low quality, failing to delineate the object of interest (\eg, rows 4, 5, 11), missing the object of interest partially (\eg, rows 1, 2, 3, 6) or even entirely (\eg, row 7). For ViT-T, it is impressive that for $\gamma$ around or equal to 1.25, the attention map quality is high, attending more (\eg, rows 1, 2, 4, 7) or even exclusively (\eg, rows 3, 6, 11) the object instead of background.

\paragraph{Attention maps: \cls vs. \Ours}

\begin{figure*}
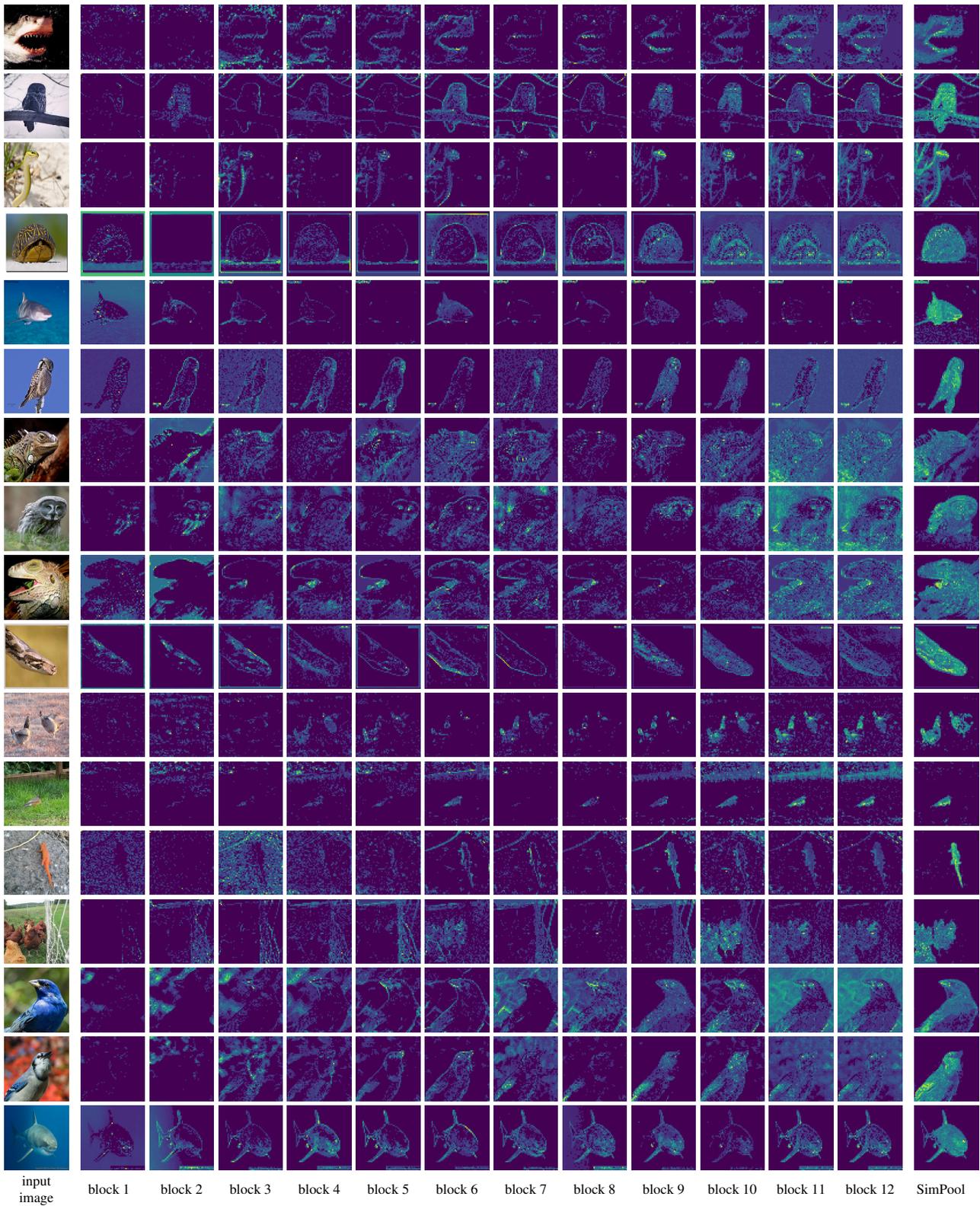

\scriptsize
\centering
\setlength{\tabcolsep}{1.2pt}
\newcommand{\sz}{.064}
%\begin{tabular}{c!{\color{red}\vrule}cccccccccccc!{\color{red}\vrule}c}
\begin{tabular}{cccccccccccccc}

\fig[\sz]{vittblocksattmaps/ILSVRC2012_val_00016168_orig.PNG} & \hspace{2pt}
\fig[\sz]{vittblocksattmaps/ILSVRC2012_val_00016168_attmap_vitt_official_supervised_selfattention_block1.PNG} &
\fig[\sz]{vittblocksattmaps/ILSVRC2012_val_00016168_attmap_vitt_official_supervised_selfattention_block2.PNG} &
\fig[\sz]{vittblocksattmaps/ILSVRC2012_val_00016168_attmap_vitt_official_supervised_selfattention_block3.PNG} &
\fig[\sz]{vittblocksattmaps/ILSVRC2012_val_00016168_attmap_vitt_official_supervised_selfattention_block4.PNG} &
\fig[\sz]{vittblocksattmaps/ILSVRC2012_val_00016168_attmap_vitt_official_supervised_selfattention_block5.PNG} &
\fig[\sz]{vittblocksattmaps/ILSVRC2012_val_00016168_attmap_vitt_official_supervised_selfattention_block6.PNG} &
\fig[\sz]{vittblocksattmaps/ILSVRC2012_val_00016168_attmap_vitt_official_supervised_selfattention_block7.PNG} &
\fig[\sz]{vittblocksattmaps/ILSVRC2012_val_00016168_attmap_vitt_official_supervised_selfattention_block8.PNG} &
\fig[\sz]{vittblocksattmaps/ILSVRC2012_val_00016168_attmap_vitt_official_supervised_selfattention_block9.PNG} &
\fig[\sz]{vittblocksattmaps/ILSVRC2012_val_00016168_attmap_vitt_official_supervised_selfattention_block10.PNG} &
\fig[\sz]{vittblocksattmaps/ILSVRC2012_val_00016168_attmap_vitt_official_supervised_selfattention_block11.PNG} &
\fig[\sz]{vittblocksattmaps/ILSVRC2012_val_00016168_attmap_vitt_official_supervised_selfattention_block12.PNG} & \hspace{2pt}
\fig[\sz]{vittblocksattmaps/ILSVRC2012_val_00016168_attmap_ep100_vitt_gem1.0_supervised_simpool.PNG} \\

\fig[\sz]{vittblocksattmaps/ILSVRC2012_val_00003169_orig.PNG} & \hspace{2pt}
\fig[\sz]{vittblocksattmaps/ILSVRC2012_val_00003169_attmap_ep100_vitt_official_supervised_selfattention_block1.PNG} &
\fig[\sz]{vittblocksattmaps/ILSVRC2012_val_00003169_attmap_ep100_vitt_official_supervised_selfattention_block2.PNG} &
\fig[\sz]{vittblocksattmaps/ILSVRC2012_val_00003169_attmap_ep100_vitt_official_supervised_selfattention_block3.PNG} &
\fig[\sz]{vittblocksattmaps/ILSVRC2012_val_00003169_attmap_ep100_vitt_official_supervised_selfattention_block4.PNG} &
\fig[\sz]{vittblocksattmaps/ILSVRC2012_val_00003169_attmap_ep100_vitt_official_supervised_selfattention_block5.PNG} &
\fig[\sz]{vittblocksattmaps/ILSVRC2012_val_00003169_attmap_ep100_vitt_official_supervised_selfattention_block6.PNG} &
\fig[\sz]{vittblocksattmaps/ILSVRC2012_val_00003169_attmap_ep100_vitt_official_supervised_selfattention_block7.PNG} &
\fig[\sz]{vittblocksattmaps/ILSVRC2012_val_00003169_attmap_ep100_vitt_official_supervised_selfattention_block8.PNG} &
\fig[\sz]{vittblocksattmaps/ILSVRC2012_val_00003169_attmap_ep100_vitt_official_supervised_selfattention_block9.PNG} &
\fig[\sz]{vittblocksattmaps/ILSVRC2012_val_00003169_attmap_ep100_vitt_official_supervised_selfattention_block10.PNG} &
\fig[\sz]{vittblocksattmaps/ILSVRC2012_val_00003169_attmap_ep100_vitt_official_supervised_selfattention_block11.PNG} &
\fig[\sz]{vittblocksattmaps/ILSVRC2012_val_00003169_attmap_ep100_vitt_official_supervised_selfattention_block12.PNG} & \hspace{2pt} 
\fig[\sz]{vittblocksattmaps/ILSVRC2012_val_00003169_attmap_ep100_vitt_gem1.0_supervised_simpool.PNG} \\

\fig[\sz]{vittblocksattmaps/ILSVRC2012_val_00005357_orig.PNG} & \hspace{2pt}
\fig[\sz]{vittblocksattmaps/ILSVRC2012_val_00005357_attmap_ep100_vitt_official_supervised_selfattention_block1.PNG} &
\fig[\sz]{vittblocksattmaps/ILSVRC2012_val_00005357_attmap_ep100_vitt_official_supervised_selfattention_block2.PNG} &
\fig[\sz]{vittblocksattmaps/ILSVRC2012_val_00005357_attmap_ep100_vitt_official_supervised_selfattention_block3.PNG} &
\fig[\sz]{vittblocksattmaps/ILSVRC2012_val_00005357_attmap_ep100_vitt_official_supervised_selfattention_block4.PNG} &
\fig[\sz]{vittblocksattmaps/ILSVRC2012_val_00005357_attmap_ep100_vitt_official_supervised_selfattention_block5.PNG} &
\fig[\sz]{vittblocksattmaps/ILSVRC2012_val_00005357_attmap_ep100_vitt_official_supervised_selfattention_block6.PNG} &
\fig[\sz]{vittblocksattmaps/ILSVRC2012_val_00005357_attmap_ep100_vitt_official_supervised_selfattention_block7.PNG} &
\fig[\sz]{vittblocksattmaps/ILSVRC2012_val_00005357_attmap_ep100_vitt_official_supervised_selfattention_block8.PNG} &
\fig[\sz]{vittblocksattmaps/ILSVRC2012_val_00005357_attmap_ep100_vitt_official_supervised_selfattention_block9.PNG} &
\fig[\sz]{vittblocksattmaps/ILSVRC2012_val_00005357_attmap_ep100_vitt_official_supervised_selfattention_block10.PNG} &
\fig[\sz]{vittblocksattmaps/ILSVRC2012_val_00005357_attmap_ep100_vitt_official_supervised_selfattention_block11.PNG} &
\fig[\sz]{vittblocksattmaps/ILSVRC2012_val_00005357_attmap_ep100_vitt_official_supervised_selfattention_block12.PNG} & \hspace{2pt}
\fig[\sz]{vittblocksattmaps/ILSVRC2012_val_00005357_attmap_ep100_vitt_gem1.0_supervised_simpool.PNG} \\

\fig[\sz]{vittblocksattmaps/ILSVRC2012_val_00007230_orig.PNG} & \hspace{2pt}
\fig[\sz]{vittblocksattmaps/ILSVRC2012_val_00007230_attmap_ep100_vitt_official_supervised_selfattention_block1.PNG} &
\fig[\sz]{vittblocksattmaps/ILSVRC2012_val_00007230_attmap_ep100_vitt_official_supervised_selfattention_block2.PNG} &
\fig[\sz]{vittblocksattmaps/ILSVRC2012_val_00007230_attmap_ep100_vitt_official_supervised_selfattention_block3.PNG} &
\fig[\sz]{vittblocksattmaps/ILSVRC2012_val_00007230_attmap_ep100_vitt_official_supervised_selfattention_block4.PNG} &
\fig[\sz]{vittblocksattmaps/ILSVRC2012_val_00007230_attmap_ep100_vitt_official_supervised_selfattention_block5.PNG} &
\fig[\sz]{vittblocksattmaps/ILSVRC2012_val_00007230_attmap_ep100_vitt_official_supervised_selfattention_block6.PNG} &
\fig[\sz]{vittblocksattmaps/ILSVRC2012_val_00007230_attmap_ep100_vitt_official_supervised_selfattention_block7.PNG} &
\fig[\sz]{vittblocksattmaps/ILSVRC2012_val_00007230_attmap_ep100_vitt_official_supervised_selfattention_block8.PNG} &
\fig[\sz]{vittblocksattmaps/ILSVRC2012_val_00007230_attmap_ep100_vitt_official_supervised_selfattention_block9.PNG} &
\fig[\sz]{vittblocksattmaps/ILSVRC2012_val_00007230_attmap_ep100_vitt_official_supervised_selfattention_block10.PNG} &
\fig[\sz]{vittblocksattmaps/ILSVRC2012_val_00007230_attmap_ep100_vitt_official_supervised_selfattention_block11.PNG} &
\fig[\sz]{vittblocksattmaps/ILSVRC2012_val_00007230_attmap_ep100_vitt_official_supervised_selfattention_block12.PNG} & \hspace{2pt}
\fig[\sz]{vittblocksattmaps/ILSVRC2012_val_00007230_attmap_ep100_vitt_gem1.0_supervised_simpool.PNG} \\

\fig[\sz]{vittblocksattmaps/ILSVRC2012_val_00007743_orig.PNG} & \hspace{2pt}
\fig[\sz]{vittblocksattmaps/ILSVRC2012_val_00007743_attmap_ep100_vitt_official_supervised_selfattention_block1.PNG} &
\fig[\sz]{vittblocksattmaps/ILSVRC2012_val_00007743_attmap_ep100_vitt_official_supervised_selfattention_block2.PNG} &
\fig[\sz]{vittblocksattmaps/ILSVRC2012_val_00007743_attmap_ep100_vitt_official_supervised_selfattention_block3.PNG} &
\fig[\sz]{vittblocksattmaps/ILSVRC2012_val_00007743_attmap_ep100_vitt_official_supervised_selfattention_block4.PNG} &
\fig[\sz]{vittblocksattmaps/ILSVRC2012_val_00007743_attmap_ep100_vitt_official_supervised_selfattention_block5.PNG} &
\fig[\sz]{vittblocksattmaps/ILSVRC2012_val_00007743_attmap_ep100_vitt_official_supervised_selfattention_block6.PNG} &
\fig[\sz]{vittblocksattmaps/ILSVRC2012_val_00007743_attmap_ep100_vitt_official_supervised_selfattention_block7.PNG} &
\fig[\sz]{vittblocksattmaps/ILSVRC2012_val_00007743_attmap_ep100_vitt_official_supervised_selfattention_block8.PNG} &
\fig[\sz]{vittblocksattmaps/ILSVRC2012_val_00007743_attmap_ep100_vitt_official_supervised_selfattention_block9.PNG} &
\fig[\sz]{vittblocksattmaps/ILSVRC2012_val_00007743_attmap_ep100_vitt_official_supervised_selfattention_block10.PNG} &
\fig[\sz]{vittblocksattmaps/ILSVRC2012_val_00007743_attmap_ep100_vitt_official_supervised_selfattention_block11.PNG} &
\fig[\sz]{vittblocksattmaps/ILSVRC2012_val_00007743_attmap_ep100_vitt_official_supervised_selfattention_block12.PNG} & \hspace{2pt}
\fig[\sz]{vittblocksattmaps/ILSVRC2012_val_00007743_attmap_ep100_vitt_gem1.0_supervised_simpool.PNG} \\

\fig[\sz]{vittblocksattmaps/ILSVRC2012_val_00015913_orig.PNG} & \hspace{2pt}
\fig[\sz]{vittblocksattmaps/ILSVRC2012_val_00015913_attmap_ep100_vitt_official_supervised_selfattention_block1.PNG} &
\fig[\sz]{vittblocksattmaps/ILSVRC2012_val_00015913_attmap_ep100_vitt_official_supervised_selfattention_block2.PNG} &
\fig[\sz]{vittblocksattmaps/ILSVRC2012_val_00015913_attmap_ep100_vitt_official_supervised_selfattention_block3.PNG} &
\fig[\sz]{vittblocksattmaps/ILSVRC2012_val_00015913_attmap_ep100_vitt_official_supervised_selfattention_block4.PNG} &
\fig[\sz]{vittblocksattmaps/ILSVRC2012_val_00015913_attmap_ep100_vitt_official_supervised_selfattention_block5.PNG} &
\fig[\sz]{vittblocksattmaps/ILSVRC2012_val_00015913_attmap_ep100_vitt_official_supervised_selfattention_block6.PNG} &
\fig[\sz]{vittblocksattmaps/ILSVRC2012_val_00015913_attmap_ep100_vitt_official_supervised_selfattention_block7.PNG} &
\fig[\sz]{vittblocksattmaps/ILSVRC2012_val_00015913_attmap_ep100_vitt_official_supervised_selfattention_block8.PNG} &
\fig[\sz]{vittblocksattmaps/ILSVRC2012_val_00015913_attmap_ep100_vitt_official_supervised_selfattention_block9.PNG} &
\fig[\sz]{vittblocksattmaps/ILSVRC2012_val_00015913_attmap_ep100_vitt_official_supervised_selfattention_block10.PNG} &
\fig[\sz]{vittblocksattmaps/ILSVRC2012_val_00015913_attmap_ep100_vitt_official_supervised_selfattention_block11.PNG} &
\fig[\sz]{vittblocksattmaps/ILSVRC2012_val_00015913_attmap_ep100_vitt_official_supervised_selfattention_block12.PNG} & \hspace{2pt}
\fig[\sz]{vittblocksattmaps/ILSVRC2012_val_00015913_attmap_ep100_vitt_gem1.0_supervised_simpool.PNG} \\

\fig[\sz]{vittblocksattmaps/ILSVRC2012_val_00000153_orig.PNG} & \hspace{2pt}
\fig[\sz]{vittblocksattmaps/ILSVRC2012_val_00000153_attmap_ep100_vitt_official_supervised_selfattention_block1.PNG} &
\fig[\sz]{vittblocksattmaps/ILSVRC2012_val_00000153_attmap_ep100_vitt_official_supervised_selfattention_block2.PNG} &
\fig[\sz]{vittblocksattmaps/ILSVRC2012_val_00000153_attmap_ep100_vitt_official_supervised_selfattention_block3.PNG} &
\fig[\sz]{vittblocksattmaps/ILSVRC2012_val_00000153_attmap_ep100_vitt_official_supervised_selfattention_block4.PNG} &
\fig[\sz]{vittblocksattmaps/ILSVRC2012_val_00000153_attmap_ep100_vitt_official_supervised_selfattention_block5.PNG} &
\fig[\sz]{vittblocksattmaps/ILSVRC2012_val_00000153_attmap_ep100_vitt_official_supervised_selfattention_block6.PNG} &
\fig[\sz]{vittblocksattmaps/ILSVRC2012_val_00000153_attmap_ep100_vitt_official_supervised_selfattention_block7.PNG} &
\fig[\sz]{vittblocksattmaps/ILSVRC2012_val_00000153_attmap_ep100_vitt_official_supervised_selfattention_block8.PNG} &
\fig[\sz]{vittblocksattmaps/ILSVRC2012_val_00000153_attmap_ep100_vitt_official_supervised_selfattention_block9.PNG} &
\fig[\sz]{vittblocksattmaps/ILSVRC2012_val_00000153_attmap_ep100_vitt_official_supervised_selfattention_block10.PNG} &
\fig[\sz]{vittblocksattmaps/ILSVRC2012_val_00000153_attmap_ep100_vitt_official_supervised_selfattention_block11.PNG} &
\fig[\sz]{vittblocksattmaps/ILSVRC2012_val_00000153_attmap_ep100_vitt_official_supervised_selfattention_block12.PNG} & \hspace{2pt}
\fig[\sz]{vittblocksattmaps/ILSVRC2012_val_00000153_attmap_ep100_vitt_gem1.0_supervised_simpool.PNG} \\

\fig[\sz]{vittblocksattmaps/ILSVRC2012_val_00000318_orig.PNG} & \hspace{2pt}
\fig[\sz]{vittblocksattmaps/ILSVRC2012_val_00000318_attmap_ep100_vitt_official_supervised_selfattention_block1.PNG} &
\fig[\sz]{vittblocksattmaps/ILSVRC2012_val_00000318_attmap_ep100_vitt_official_supervised_selfattention_block2.PNG} &
\fig[\sz]{vittblocksattmaps/ILSVRC2012_val_00000318_attmap_ep100_vitt_official_supervised_selfattention_block3.PNG} &
\fig[\sz]{vittblocksattmaps/ILSVRC2012_val_00000318_attmap_ep100_vitt_official_supervised_selfattention_block4.PNG} &
\fig[\sz]{vittblocksattmaps/ILSVRC2012_val_00000318_attmap_ep100_vitt_official_supervised_selfattention_block5.PNG} &
\fig[\sz]{vittblocksattmaps/ILSVRC2012_val_00000318_attmap_ep100_vitt_official_supervised_selfattention_block6.PNG} &
\fig[\sz]{vittblocksattmaps/ILSVRC2012_val_00000318_attmap_ep100_vitt_official_supervised_selfattention_block7.PNG} &
\fig[\sz]{vittblocksattmaps/ILSVRC2012_val_00000318_attmap_ep100_vitt_official_supervised_selfattention_block8.PNG} &
\fig[\sz]{vittblocksattmaps/ILSVRC2012_val_00000318_attmap_ep100_vitt_official_supervised_selfattention_block9.PNG} &
\fig[\sz]{vittblocksattmaps/ILSVRC2012_val_00000318_attmap_ep100_vitt_official_supervised_selfattention_block10.PNG} &
\fig[\sz]{vittblocksattmaps/ILSVRC2012_val_00000318_attmap_ep100_vitt_official_supervised_selfattention_block11.PNG} &
\fig[\sz]{vittblocksattmaps/ILSVRC2012_val_00000318_attmap_ep100_vitt_official_supervised_selfattention_block12.PNG} & \hspace{2pt}
\fig[\sz]{vittblocksattmaps/ILSVRC2012_val_00000318_attmap_ep100_vitt_gem1.0_supervised_simpool.PNG} \\

\fig[\sz]{vittblocksattmaps/ILSVRC2012_val_00000471_orig.PNG} & \hspace{2pt}
\fig[\sz]{vittblocksattmaps/ILSVRC2012_val_00000471_attmap_ep100_vitt_official_supervised_selfattention_block1.PNG} &
\fig[\sz]{vittblocksattmaps/ILSVRC2012_val_00000471_attmap_ep100_vitt_official_supervised_selfattention_block2.PNG} &
\fig[\sz]{vittblocksattmaps/ILSVRC2012_val_00000471_attmap_ep100_vitt_official_supervised_selfattention_block3.PNG} &
\fig[\sz]{vittblocksattmaps/ILSVRC2012_val_00000471_attmap_ep100_vitt_official_supervised_selfattention_block4.PNG} &
\fig[\sz]{vittblocksattmaps/ILSVRC2012_val_00000471_attmap_ep100_vitt_official_supervised_selfattention_block5.PNG} &
\fig[\sz]{vittblocksattmaps/ILSVRC2012_val_00000471_attmap_ep100_vitt_official_supervised_selfattention_block6.PNG} &
\fig[\sz]{vittblocksattmaps/ILSVRC2012_val_00000471_attmap_ep100_vitt_official_supervised_selfattention_block7.PNG} &
\fig[\sz]{vittblocksattmaps/ILSVRC2012_val_00000471_attmap_ep100_vitt_official_supervised_selfattention_block8.PNG} &
\fig[\sz]{vittblocksattmaps/ILSVRC2012_val_00000471_attmap_ep100_vitt_official_supervised_selfattention_block9.PNG} &
\fig[\sz]{vittblocksattmaps/ILSVRC2012_val_00000471_attmap_ep100_vitt_official_supervised_selfattention_block10.PNG} &
\fig[\sz]{vittblocksattmaps/ILSVRC2012_val_00000471_attmap_ep100_vitt_official_supervised_selfattention_block11.PNG} &
\fig[\sz]{vittblocksattmaps/ILSVRC2012_val_00000471_attmap_ep100_vitt_official_supervised_selfattention_block12.PNG} & \hspace{2pt}
\fig[\sz]{vittblocksattmaps/ILSVRC2012_val_00000471_attmap_ep100_vitt_gem1.0_supervised_simpool.PNG} \\

\fig[\sz]{vittblocksattmaps/ILSVRC2012_val_00000688_orig.PNG} & \hspace{2pt}
\fig[\sz]{vittblocksattmaps/ILSVRC2012_val_00000688_attmap_ep100_vitt_official_supervised_selfattention_block1.PNG} &
\fig[\sz]{vittblocksattmaps/ILSVRC2012_val_00000688_attmap_ep100_vitt_official_supervised_selfattention_block2.PNG} &
\fig[\sz]{vittblocksattmaps/ILSVRC2012_val_00000688_attmap_ep100_vitt_official_supervised_selfattention_block3.PNG} &
\fig[\sz]{vittblocksattmaps/ILSVRC2012_val_00000688_attmap_ep100_vitt_official_supervised_selfattention_block4.PNG} &
\fig[\sz]{vittblocksattmaps/ILSVRC2012_val_00000688_attmap_ep100_vitt_official_supervised_selfattention_block5.PNG} &
\fig[\sz]{vittblocksattmaps/ILSVRC2012_val_00000688_attmap_ep100_vitt_official_supervised_selfattention_block6.PNG} &
\fig[\sz]{vittblocksattmaps/ILSVRC2012_val_00000688_attmap_ep100_vitt_official_supervised_selfattention_block7.PNG} &
\fig[\sz]{vittblocksattmaps/ILSVRC2012_val_00000688_attmap_ep100_vitt_official_supervised_selfattention_block8.PNG} &
\fig[\sz]{vittblocksattmaps/ILSVRC2012_val_00000688_attmap_ep100_vitt_official_supervised_selfattention_block9.PNG} &
\fig[\sz]{vittblocksattmaps/ILSVRC2012_val_00000688_attmap_ep100_vitt_official_supervised_selfattention_block10.PNG} &
\fig[\sz]{vittblocksattmaps/ILSVRC2012_val_00000688_attmap_ep100_vitt_official_supervised_selfattention_block11.PNG} &
\fig[\sz]{vittblocksattmaps/ILSVRC2012_val_00000688_attmap_ep100_vitt_official_supervised_selfattention_block12.PNG} & \hspace{2pt}
\fig[\sz]{vittblocksattmaps/ILSVRC2012_val_00000688_attmap_ep100_vitt_gem1.0_supervised_simpool.PNG} \\

\fig[\sz]{vittblocksattmaps/ILSVRC2012_val_00001461_orig.PNG} & \hspace{2pt}
\fig[\sz]{vittblocksattmaps/ILSVRC2012_val_00001461_attmap_ep100_vitt_official_supervised_selfattention_block1.PNG} &
\fig[\sz]{vittblocksattmaps/ILSVRC2012_val_00001461_attmap_ep100_vitt_official_supervised_selfattention_block2.PNG} &
\fig[\sz]{vittblocksattmaps/ILSVRC2012_val_00001461_attmap_ep100_vitt_official_supervised_selfattention_block3.PNG} &
\fig[\sz]{vittblocksattmaps/ILSVRC2012_val_00001461_attmap_ep100_vitt_official_supervised_selfattention_block4.PNG} &
\fig[\sz]{vittblocksattmaps/ILSVRC2012_val_00001461_attmap_ep100_vitt_official_supervised_selfattention_block5.PNG} &
\fig[\sz]{vittblocksattmaps/ILSVRC2012_val_00001461_attmap_ep100_vitt_official_supervised_selfattention_block6.PNG} &
\fig[\sz]{vittblocksattmaps/ILSVRC2012_val_00001461_attmap_ep100_vitt_official_supervised_selfattention_block7.PNG} &
\fig[\sz]{vittblocksattmaps/ILSVRC2012_val_00001461_attmap_ep100_vitt_official_supervised_selfattention_block8.PNG} &
\fig[\sz]{vittblocksattmaps/ILSVRC2012_val_00001461_attmap_ep100_vitt_official_supervised_selfattention_block9.PNG} &
\fig[\sz]{vittblocksattmaps/ILSVRC2012_val_00001461_attmap_ep100_vitt_official_supervised_selfattention_block10.PNG} &
\fig[\sz]{vittblocksattmaps/ILSVRC2012_val_00001461_attmap_ep100_vitt_official_supervised_selfattention_block11.PNG} &
\fig[\sz]{vittblocksattmaps/ILSVRC2012_val_00001461_attmap_ep100_vitt_official_supervised_selfattention_block12.PNG} & \hspace{2pt}
\fig[\sz]{vittblocksattmaps/ILSVRC2012_val_00001461_attmap_ep100_vitt_gem1.0_supervised_simpool.PNG} \\

\fig[\sz]{vittblocksattmaps/ILSVRC2012_val_00001598_orig.PNG} & \hspace{2pt}
\fig[\sz]{vittblocksattmaps/ILSVRC2012_val_00001598_attmap_ep100_vitt_official_supervised_selfattention_block1.PNG} &
\fig[\sz]{vittblocksattmaps/ILSVRC2012_val_00001598_attmap_ep100_vitt_official_supervised_selfattention_block2.PNG} &
\fig[\sz]{vittblocksattmaps/ILSVRC2012_val_00001598_attmap_ep100_vitt_official_supervised_selfattention_block3.PNG} &
\fig[\sz]{vittblocksattmaps/ILSVRC2012_val_00001598_attmap_ep100_vitt_official_supervised_selfattention_block4.PNG} &
\fig[\sz]{vittblocksattmaps/ILSVRC2012_val_00001598_attmap_ep100_vitt_official_supervised_selfattention_block5.PNG} &
\fig[\sz]{vittblocksattmaps/ILSVRC2012_val_00001598_attmap_ep100_vitt_official_supervised_selfattention_block6.PNG} &
\fig[\sz]{vittblocksattmaps/ILSVRC2012_val_00001598_attmap_ep100_vitt_official_supervised_selfattention_block7.PNG} &
\fig[\sz]{vittblocksattmaps/ILSVRC2012_val_00001598_attmap_ep100_vitt_official_supervised_selfattention_block8.PNG} &
\fig[\sz]{vittblocksattmaps/ILSVRC2012_val_00001598_attmap_ep100_vitt_official_supervised_selfattention_block9.PNG} &
\fig[\sz]{vittblocksattmaps/ILSVRC2012_val_00001598_attmap_ep100_vitt_official_supervised_selfattention_block10.PNG} &
\fig[\sz]{vittblocksattmaps/ILSVRC2012_val_00001598_attmap_ep100_vitt_official_supervised_selfattention_block11.PNG} &
\fig[\sz]{vittblocksattmaps/ILSVRC2012_val_00001598_attmap_ep100_vitt_official_supervised_selfattention_block12.PNG} & \hspace{2pt}
\fig[\sz]{vittblocksattmaps/ILSVRC2012_val_00001598_attmap_ep100_vitt_gem1.0_supervised_simpool.PNG} \\

\fig[\sz]{vittblocksattmaps/ILSVRC2012_val_00001728_orig.PNG} & \hspace{2pt}
\fig[\sz]{vittblocksattmaps/ILSVRC2012_val_00001728_attmap_ep100_vitt_official_supervised_selfattention_block1.PNG} &
\fig[\sz]{vittblocksattmaps/ILSVRC2012_val_00001728_attmap_ep100_vitt_official_supervised_selfattention_block2.PNG} &
\fig[\sz]{vittblocksattmaps/ILSVRC2012_val_00001728_attmap_ep100_vitt_official_supervised_selfattention_block3.PNG} &
\fig[\sz]{vittblocksattmaps/ILSVRC2012_val_00001728_attmap_ep100_vitt_official_supervised_selfattention_block4.PNG} &
\fig[\sz]{vittblocksattmaps/ILSVRC2012_val_00001728_attmap_ep100_vitt_official_supervised_selfattention_block5.PNG} &
\fig[\sz]{vittblocksattmaps/ILSVRC2012_val_00001728_attmap_ep100_vitt_official_supervised_selfattention_block6.PNG} &
\fig[\sz]{vittblocksattmaps/ILSVRC2012_val_00001728_attmap_ep100_vitt_official_supervised_selfattention_block7.PNG} &
\fig[\sz]{vittblocksattmaps/ILSVRC2012_val_00001728_attmap_ep100_vitt_official_supervised_selfattention_block8.PNG} &
\fig[\sz]{vittblocksattmaps/ILSVRC2012_val_00001728_attmap_ep100_vitt_official_supervised_selfattention_block9.PNG} &
\fig[\sz]{vittblocksattmaps/ILSVRC2012_val_00001728_attmap_ep100_vitt_official_supervised_selfattention_block10.PNG} &
\fig[\sz]{vittblocksattmaps/ILSVRC2012_val_00001728_attmap_ep100_vitt_official_supervised_selfattention_block11.PNG} &
\fig[\sz]{vittblocksattmaps/ILSVRC2012_val_00001728_attmap_ep100_vitt_official_supervised_selfattention_block12.PNG} & \hspace{2pt}
\fig[\sz]{vittblocksattmaps/ILSVRC2012_val_00001728_attmap_ep100_vitt_gem1.0_supervised_simpool.PNG} \\

\fig[\sz]{vittblocksattmaps/ILSVRC2012_val_00001906_orig.PNG} & \hspace{2pt}
\fig[\sz]{vittblocksattmaps/ILSVRC2012_val_00001906_attmap_ep100_vitt_official_supervised_selfattention_block1.PNG} &
\fig[\sz]{vittblocksattmaps/ILSVRC2012_val_00001906_attmap_ep100_vitt_official_supervised_selfattention_block2.PNG} &
\fig[\sz]{vittblocksattmaps/ILSVRC2012_val_00001906_attmap_ep100_vitt_official_supervised_selfattention_block3.PNG} &
\fig[\sz]{vittblocksattmaps/ILSVRC2012_val_00001906_attmap_ep100_vitt_official_supervised_selfattention_block4.PNG} &
\fig[\sz]{vittblocksattmaps/ILSVRC2012_val_00001906_attmap_ep100_vitt_official_supervised_selfattention_block5.PNG} &
\fig[\sz]{vittblocksattmaps/ILSVRC2012_val_00001906_attmap_ep100_vitt_official_supervised_selfattention_block6.PNG} &
\fig[\sz]{vittblocksattmaps/ILSVRC2012_val_00001906_attmap_ep100_vitt_official_supervised_selfattention_block7.PNG} &
\fig[\sz]{vittblocksattmaps/ILSVRC2012_val_00001906_attmap_ep100_vitt_official_supervised_selfattention_block8.PNG} &
\fig[\sz]{vittblocksattmaps/ILSVRC2012_val_00001906_attmap_ep100_vitt_official_supervised_selfattention_block9.PNG} &
\fig[\sz]{vittblocksattmaps/ILSVRC2012_val_00001906_attmap_ep100_vitt_official_supervised_selfattention_block10.PNG} &
\fig[\sz]{vittblocksattmaps/ILSVRC2012_val_00001906_attmap_ep100_vitt_official_supervised_selfattention_block11.PNG} &
\fig[\sz]{vittblocksattmaps/ILSVRC2012_val_00001906_attmap_ep100_vitt_official_supervised_selfattention_block12.PNG} & \hspace{2pt}
\fig[\sz]{vittblocksattmaps/ILSVRC2012_val_00001906_attmap_ep100_vitt_gem1.0_supervised_simpool.PNG} \\

\fig[\sz]{vittblocksattmaps/ILSVRC2012_val_00002071_orig.PNG} & \hspace{2pt}
\fig[\sz]{vittblocksattmaps/ILSVRC2012_val_00002071_attmap_ep100_vitt_official_supervised_selfattention_block1.PNG} &
\fig[\sz]{vittblocksattmaps/ILSVRC2012_val_00002071_attmap_ep100_vitt_official_supervised_selfattention_block2.PNG} &
\fig[\sz]{vittblocksattmaps/ILSVRC2012_val_00002071_attmap_ep100_vitt_official_supervised_selfattention_block3.PNG} &
\fig[\sz]{vittblocksattmaps/ILSVRC2012_val_00002071_attmap_ep100_vitt_official_supervised_selfattention_block4.PNG} &
\fig[\sz]{vittblocksattmaps/ILSVRC2012_val_00002071_attmap_ep100_vitt_official_supervised_selfattention_block5.PNG} &
\fig[\sz]{vittblocksattmaps/ILSVRC2012_val_00002071_attmap_ep100_vitt_official_supervised_selfattention_block6.PNG} &
\fig[\sz]{vittblocksattmaps/ILSVRC2012_val_00002071_attmap_ep100_vitt_official_supervised_selfattention_block7.PNG} &
\fig[\sz]{vittblocksattmaps/ILSVRC2012_val_00002071_attmap_ep100_vitt_official_supervised_selfattention_block8.PNG} &
\fig[\sz]{vittblocksattmaps/ILSVRC2012_val_00002071_attmap_ep100_vitt_official_supervised_selfattention_block9.PNG} &
\fig[\sz]{vittblocksattmaps/ILSVRC2012_val_00002071_attmap_ep100_vitt_official_supervised_selfattention_block10.PNG} &
\fig[\sz]{vittblocksattmaps/ILSVRC2012_val_00002071_attmap_ep100_vitt_official_supervised_selfattention_block11.PNG} &
\fig[\sz]{vittblocksattmaps/ILSVRC2012_val_00002071_attmap_ep100_vitt_official_supervised_selfattention_block12.PNG} & \hspace{2pt}
\fig[\sz]{vittblocksattmaps/ILSVRC2012_val_00002071_attmap_ep100_vitt_gem1.0_supervised_simpool.PNG} \\

\fig[\sz]{vittblocksattmaps/ILSVRC2012_val_00002655_orig.PNG} & \hspace{2pt}
\fig[\sz]{vittblocksattmaps/ILSVRC2012_val_00002655_attmap_ep100_vitt_official_supervised_selfattention_block1.PNG} &
\fig[\sz]{vittblocksattmaps/ILSVRC2012_val_00002655_attmap_ep100_vitt_official_supervised_selfattention_block2.PNG} &
\fig[\sz]{vittblocksattmaps/ILSVRC2012_val_00002655_attmap_ep100_vitt_official_supervised_selfattention_block3.PNG} &
\fig[\sz]{vittblocksattmaps/ILSVRC2012_val_00002655_attmap_ep100_vitt_official_supervised_selfattention_block4.PNG} &
\fig[\sz]{vittblocksattmaps/ILSVRC2012_val_00002655_attmap_ep100_vitt_official_supervised_selfattention_block5.PNG} &
\fig[\sz]{vittblocksattmaps/ILSVRC2012_val_00002655_attmap_ep100_vitt_official_supervised_selfattention_block6.PNG} &
\fig[\sz]{vittblocksattmaps/ILSVRC2012_val_00002655_attmap_ep100_vitt_official_supervised_selfattention_block7.PNG} &
\fig[\sz]{vittblocksattmaps/ILSVRC2012_val_00002655_attmap_ep100_vitt_official_supervised_selfattention_block8.PNG} &
\fig[\sz]{vittblocksattmaps/ILSVRC2012_val_00002655_attmap_ep100_vitt_official_supervised_selfattention_block9.PNG} &
\fig[\sz]{vittblocksattmaps/ILSVRC2012_val_00002655_attmap_ep100_vitt_official_supervised_selfattention_block10.PNG} &
\fig[\sz]{vittblocksattmaps/ILSVRC2012_val_00002655_attmap_ep100_vitt_official_supervised_selfattention_block11.PNG} &
\fig[\sz]{vittblocksattmaps/ILSVRC2012_val_00002655_attmap_ep100_vitt_official_supervised_selfattention_block12.PNG} & \hspace{2pt}
\fig[\sz]{vittblocksattmaps/ILSVRC2012_val_00002655_attmap_ep100_vitt_gem1.0_supervised_simpool.PNG} \\

\fig[\sz]{vittblocksattmaps/ILSVRC2012_val_00002752_orig.PNG} & \hspace{2pt}
\fig[\sz]{vittblocksattmaps/ILSVRC2012_val_00002752_attmap_ep100_vitt_official_supervised_selfattention_block1.PNG} &
\fig[\sz]{vittblocksattmaps/ILSVRC2012_val_00002752_attmap_ep100_vitt_official_supervised_selfattention_block2.PNG} &
\fig[\sz]{vittblocksattmaps/ILSVRC2012_val_00002752_attmap_ep100_vitt_official_supervised_selfattention_block3.PNG} &
\fig[\sz]{vittblocksattmaps/ILSVRC2012_val_00002752_attmap_ep100_vitt_official_supervised_selfattention_block4.PNG} &
\fig[\sz]{vittblocksattmaps/ILSVRC2012_val_00002752_attmap_ep100_vitt_official_supervised_selfattention_block5.PNG} &
\fig[\sz]{vittblocksattmaps/ILSVRC2012_val_00002752_attmap_ep100_vitt_official_supervised_selfattention_block6.PNG} &
\fig[\sz]{vittblocksattmaps/ILSVRC2012_val_00002752_attmap_ep100_vitt_official_supervised_selfattention_block7.PNG} &
\fig[\sz]{vittblocksattmaps/ILSVRC2012_val_00002752_attmap_ep100_vitt_official_supervised_selfattention_block8.PNG} &
\fig[\sz]{vittblocksattmaps/ILSVRC2012_val_00002752_attmap_ep100_vitt_official_supervised_selfattention_block9.PNG} &
\fig[\sz]{vittblocksattmaps/ILSVRC2012_val_00002752_attmap_ep100_vitt_official_supervised_selfattention_block10.PNG} &
\fig[\sz]{vittblocksattmaps/ILSVRC2012_val_00002752_attmap_ep100_vitt_official_supervised_selfattention_block11.PNG} &
\fig[\sz]{vittblocksattmaps/ILSVRC2012_val_00002752_attmap_ep100_vitt_official_supervised_selfattention_block12.PNG} & \hspace{2pt}
\fig[\sz]{vittblocksattmaps/ILSVRC2012_val_00002752_attmap_ep100_vitt_gem1.0_supervised_simpool.PNG} \\

input &
\mr{2}{block 1} &
\mr{2}{block 2} &
\mr{2}{block 3} &
\mr{2}{block 4} &
\mr{2}{block 5} &
\mr{2}{block 6} &
\mr{2}{block 7} &
\mr{2}{block 8} &
\mr{2}{block 9} &
\mr{2}{block 10} &
\mr{2}{block 11} &
\mr{2}{block 12} &
\mr{2}{\Ours} \\

image &
&
&
&
&
&
&
&
&
&
&
&
&
\\

\end{tabular}
\vspace{3pt}
\caption{\emph{\cls vs. \Ours}. Attention maps of ViT-T~\cite{vit} trained on \imagenet for 100 epochs under supervision. For \cls, we use the mean attention map of the \cls token of each block. For \Ours, we use the attention map $\va$~\eq{sp-att}. Input image resolution: $896 \times 896$; patches: $16 \times 16$; output attention map: $56 \times 56$.}
\label{fig:attention-maps-vit-cls-simpool}
\end{figure*}

\autoref{fig:attention-maps-vit-cls-simpool} compares the quality of the attention maps of supervised ViT-T trained with \cls to that of \Ours. For \cls, we visualize the mean attention map of the heads of the \cls token for each of the 12 blocks. For \Ours, we visualize the attention map $\va$~\eq{sp-att}. \Ours has attention maps of consistently higher quality, delineating and exclusively focusing on the object of interest (\eg, rows 6, 10, 13). It is impressive that while \cls interacts with patch tokens in 12 different blocks, it is inferior to \Ours, which interacts only once at the end.

\paragraph{Attention maps: ResNet, ConvNeXt}

\autoref{fig:attention-maps-resnet50} and \autoref{fig:attention-maps-convnext} show attention maps of supervised and self-supervised ResNet-50 and ConvNeXt-S, respectively. Both networks are pre-trained on \imagenet for 100 epochs. We use the attention map $\va$~\eq{sp-att}. We observe that \Ours enables the default ResNet-50 and ConvNeXt-S to obtain raw attention maps of high quality, focusing on the object of interest and not on background or other objects. This is not possible with the default global average pooling and is a property commonly thought of vision transformers when self-supervised~\cite{dino}. Between supervised and self-supervised \Ours, the quality differences are small, with self-supervised being slightly superior.

%------------------------------------------------------------------------------

%------------------------------------------------------------------------------
%------------------------------------------------------------------------------

%------------------------------------------------------------------------------
%------------------------------------------------------------------------------

%------------------------------------------------------------------------------
%------------------------------------------------------------------------------

%------------------------------------------------------------------------------
\begin{figure*}
\scriptsize
\centering
\setlength{\tabcolsep}{1.2pt}
\newcommand{\sz}{.1035}
\begin{tabular}{ccccccccc}

\fig[\sz]{resnet50attmaps/ILSVRC2012_val_00000040_orig.PNG} &
\fig[\sz]{resnet50attmaps/ILSVRC2012_val_00000040_attmap_28x28_resnet50_simpool_ep200_lr0.2.PNG} &
\fig[\sz]{resnet50attmaps/ILSVRC2012_val_00000040_attmap_28x28_resnet50_dino_gem_lnkv.PNG} & \hspace{2pt}
\fig[\sz]{resnet50attmaps/ILSVRC2012_val_00000064_orig.PNG} &
\fig[\sz]{resnet50attmaps/ILSVRC2012_val_00000064_attmap_28x28_resnet50_simpool_ep200_lr0.2.PNG} &
\fig[\sz]{resnet50attmaps/ILSVRC2012_val_00000064_attmap_28x28_resnet50_dino_gem_lnkv.PNG} & \hspace{2pt}
\fig[\sz]{resnet50attmaps/ILSVRC2012_val_00000651_orig.PNG} &
\fig[\sz]{resnet50attmaps/ILSVRC2012_val_00000651_attmap_28x28_resnet50_simpool_ep200_lr0.2.PNG} &
\fig[\sz]{resnet50attmaps/ILSVRC2012_val_00000651_attmap_28x28_resnet50_dino_gem_lnkv.PNG} \\

\fig[\sz]{resnet50attmaps/ILSVRC2012_val_00001185_orig.PNG} &
\fig[\sz]{resnet50attmaps/ILSVRC2012_val_00001185_attmap_28x28_resnet50_simpool_ep200_lr0.2.PNG} &
\fig[\sz]{resnet50attmaps/ILSVRC2012_val_00001185_attmap_28x28_resnet50_dino_gem_lnkv.PNG} & \hspace{2pt}
\fig[\sz]{resnet50attmaps/ILSVRC2012_val_00001274_orig.PNG} &
\fig[\sz]{resnet50attmaps/ILSVRC2012_val_00001274_attmap_28x28_resnet50_simpool_ep200_lr0.2.PNG} &
\fig[\sz]{resnet50attmaps/ILSVRC2012_val_00001274_attmap_28x28_resnet50_dino_gem_lnkv.PNG} & \hspace{2pt}
\fig[\sz]{resnet50attmaps/ILSVRC2012_val_00001739_orig.PNG} &
\fig[\sz]{resnet50attmaps/ILSVRC2012_val_00001739_attmap_28x28_resnet50_simpool_ep200_lr0.2.PNG} &
\fig[\sz]{resnet50attmaps/ILSVRC2012_val_00001739_attmap_28x28_resnet50_dino_gem_lnkv.PNG} \\

\fig[\sz]{resnet50attmaps/ILSVRC2012_val_00001887_orig.PNG} &
\fig[\sz]{resnet50attmaps/ILSVRC2012_val_00001887_attmap_28x28_resnet50_simpool_ep200_lr0.2.PNG} &
\fig[\sz]{resnet50attmaps/ILSVRC2012_val_00001887_attmap_28x28_resnet50_dino_gem_lnkv.PNG} & \hspace{2pt}
\fig[\sz]{resnet50attmaps/ILSVRC2012_val_00002338_orig.PNG} &
\fig[\sz]{resnet50attmaps/ILSVRC2012_val_00002338_attmap_28x28_resnet50_simpool_ep200_lr0.2.PNG} &
\fig[\sz]{resnet50attmaps/ILSVRC2012_val_00002338_attmap_28x28_resnet50_dino_gem_lnkv.PNG} & \hspace{2pt}
\fig[\sz]{resnet50attmaps/ILSVRC2012_val_00002928_orig.PNG} &
\fig[\sz]{resnet50attmaps/ILSVRC2012_val_00002928_attmap_28x28_resnet50_simpool_ep200_lr0.2.PNG} &
\fig[\sz]{resnet50attmaps/ILSVRC2012_val_00002928_attmap_28x28_resnet50_dino_gem_lnkv.PNG} \\

\fig[\sz]{resnet50attmaps/ILSVRC2012_val_00003169_orig.PNG} &
\fig[\sz]{resnet50attmaps/ILSVRC2012_val_00003169_attmap_28x28_resnet50_simpool_ep200_lr0.2.PNG} &
\fig[\sz]{resnet50attmaps/ILSVRC2012_val_00003169_attmap_28x28_resnet50_dino_gem_lnkv.PNG} & \hspace{2pt}
\fig[\sz]{resnet50attmaps/ILSVRC2012_val_00003577_orig.PNG} &
\fig[\sz]{resnet50attmaps/ILSVRC2012_val_00003577_attmap_28x28_resnet50_simpool_ep200_lr0.2.PNG} &
\fig[\sz]{resnet50attmaps/ILSVRC2012_val_00003577_attmap_28x28_resnet50_dino_gem_lnkv.PNG} & \hspace{2pt}
\fig[\sz]{resnet50attmaps/ILSVRC2012_val_00003985_orig.PNG} &
\fig[\sz]{resnet50attmaps/ILSVRC2012_val_00003985_attmap_28x28_resnet50_simpool_ep200_lr0.2.PNG} &
\fig[\sz]{resnet50attmaps/ILSVRC2012_val_00003985_attmap_28x28_resnet50_dino_gem_lnkv.PNG} \\

\fig[\sz]{resnet50attmaps/ILSVRC2012_val_00004110_orig.PNG} &
\fig[\sz]{resnet50attmaps/ILSVRC2012_val_00004110_attmap_28x28_resnet50_simpool_ep200_lr0.2.PNG} &
\fig[\sz]{resnet50attmaps/ILSVRC2012_val_00004110_attmap_28x28_resnet50_dino_gem_lnkv.PNG} & \hspace{2pt}
\fig[\sz]{resnet50attmaps/ILSVRC2012_val_00004245_orig.PNG} &
\fig[\sz]{resnet50attmaps/ILSVRC2012_val_00004245_attmap_28x28_resnet50_simpool_ep200_lr0.2.PNG} &
\fig[\sz]{resnet50attmaps/ILSVRC2012_val_00004245_attmap_28x28_resnet50_dino_gem_lnkv.PNG} & \hspace{2pt}
\fig[\sz]{resnet50attmaps/ILSVRC2012_val_00004381_orig.PNG} &
\fig[\sz]{resnet50attmaps/ILSVRC2012_val_00004381_attmap_28x28_resnet50_simpool_ep200_lr0.2.PNG} &
\fig[\sz]{resnet50attmaps/ILSVRC2012_val_00004381_attmap_28x28_resnet50_dino_gem_lnkv.PNG} \\

\fig[\sz]{resnet50attmaps/ILSVRC2012_val_00004747_orig.PNG} &
\fig[\sz]{resnet50attmaps/ILSVRC2012_val_00004747_attmap_28x28_resnet50_simpool_ep200_lr0.2.PNG} &
\fig[\sz]{resnet50attmaps/ILSVRC2012_val_00004747_attmap_28x28_resnet50_dino_gem_lnkv.PNG} & \hspace{2pt}
\fig[\sz]{resnet50attmaps/ILSVRC2012_val_00005290_orig.PNG} &
\fig[\sz]{resnet50attmaps/ILSVRC2012_val_00005290_attmap_28x28_resnet50_simpool_ep200_lr0.2.PNG} &
\fig[\sz]{resnet50attmaps/ILSVRC2012_val_00005290_attmap_28x28_resnet50_dino_gem_lnkv.PNG} & \hspace{2pt}
\fig[\sz]{resnet50attmaps/ILSVRC2012_val_00006524_orig.PNG} &
\fig[\sz]{resnet50attmaps/ILSVRC2012_val_00006524_attmap_28x28_resnet50_simpool_ep200_lr0.2.PNG} &
\fig[\sz]{resnet50attmaps/ILSVRC2012_val_00006524_attmap_28x28_resnet50_dino_gem_lnkv.PNG} \\

\fig[\sz]{resnet50attmaps/ILSVRC2012_val_00008316_orig.PNG} &
\fig[\sz]{resnet50attmaps/ILSVRC2012_val_00008316_attmap_28x28_resnet50_simpool_ep200_lr0.2.PNG} &
\fig[\sz]{resnet50attmaps/ILSVRC2012_val_00008316_attmap_28x28_resnet50_dino_gem_lnkv.PNG} & \hspace{2pt}
\fig[\sz]{resnet50attmaps/ILSVRC2012_val_00009132_orig.PNG} &
\fig[\sz]{resnet50attmaps/ILSVRC2012_val_00009132_attmap_28x28_resnet50_simpool_ep200_lr0.2.PNG} &
\fig[\sz]{resnet50attmaps/ILSVRC2012_val_00009132_attmap_28x28_resnet50_dino_gem_lnkv.PNG} & \hspace{2pt}
\fig[\sz]{resnet50attmaps/ILSVRC2012_val_00009831_orig.PNG} &
\fig[\sz]{resnet50attmaps/ILSVRC2012_val_00009831_attmap_28x28_resnet50_simpool_ep200_lr0.2.PNG} &
\fig[\sz]{resnet50attmaps/ILSVRC2012_val_00009831_attmap_28x28_resnet50_dino_gem_lnkv.PNG} \\

\fig[\sz]{resnet50attmaps/ILSVRC2012_val_00011053_orig.PNG} &
\fig[\sz]{resnet50attmaps/ILSVRC2012_val_00011053_attmap_28x28_resnet50_simpool_ep200_lr0.2.PNG} &
\fig[\sz]{resnet50attmaps/ILSVRC2012_val_00011053_attmap_28x28_resnet50_dino_gem_lnkv.PNG} & \hspace{2pt}
\fig[\sz]{resnet50attmaps/ILSVRC2012_val_00011129_orig.PNG} &
\fig[\sz]{resnet50attmaps/ILSVRC2012_val_00011129_attmap_28x28_resnet50_simpool_ep200_lr0.2.PNG} &
\fig[\sz]{resnet50attmaps/ILSVRC2012_val_00011129_attmap_28x28_resnet50_dino_gem_lnkv.PNG} & \hspace{2pt}
\fig[\sz]{resnet50attmaps/ILSVRC2012_val_00012248_orig.PNG} &
\fig[\sz]{resnet50attmaps/ILSVRC2012_val_00012248_attmap_28x28_resnet50_simpool_ep200_lr0.2.PNG} &
\fig[\sz]{resnet50attmaps/ILSVRC2012_val_00012248_attmap_28x28_resnet50_dino_gem_lnkv.PNG} \\

\fig[\sz]{resnet50attmaps/ILSVRC2012_val_00013265_orig.PNG} &
\fig[\sz]{resnet50attmaps/ILSVRC2012_val_00013265_attmap_28x28_resnet50_simpool_ep200_lr0.2.PNG} &
\fig[\sz]{resnet50attmaps/ILSVRC2012_val_00013265_attmap_28x28_resnet50_dino_gem_lnkv.PNG} & \hspace{2pt}
\fig[\sz]{resnet50attmaps/ILSVRC2012_val_00013945_orig.PNG} &
\fig[\sz]{resnet50attmaps/ILSVRC2012_val_00013945_attmap_28x28_resnet50_simpool_ep200_lr0.2.PNG} &
\fig[\sz]{resnet50attmaps/ILSVRC2012_val_00013945_attmap_28x28_resnet50_dino_gem_lnkv.PNG} & \hspace{2pt}
\fig[\sz]{resnet50attmaps/ILSVRC2012_val_00014082_orig.PNG} &
\fig[\sz]{resnet50attmaps/ILSVRC2012_val_00014082_attmap_28x28_resnet50_simpool_ep200_lr0.2.PNG} &
\fig[\sz]{resnet50attmaps/ILSVRC2012_val_00014082_attmap_28x28_resnet50_dino_gem_lnkv.PNG} \\

\fig[\sz]{resnet50attmaps/ILSVRC2012_val_00000498_orig.PNG} &
\fig[\sz]{resnet50attmaps/ILSVRC2012_val_00000498_attmap_28x28_resnet50_simpool_ep200_lr0.2.PNG} &
\fig[\sz]{resnet50attmaps/ILSVRC2012_val_00000498_attmap_28x28_resnet50_dino_gem_lnkv.PNG} & \hspace{2pt}
\fig[\sz]{resnet50attmaps/ILSVRC2012_val_00003621_orig.PNG} &
\fig[\sz]{resnet50attmaps/ILSVRC2012_val_00003621_attmap_28x28_resnet50_simpool_ep200_lr0.2.PNG} &
\fig[\sz]{resnet50attmaps/ILSVRC2012_val_00003621_attmap_28x28_resnet50_dino_gem_lnkv.PNG} & \hspace{2pt}
\fig[\sz]{resnet50attmaps/ILSVRC2012_val_00002848_orig.PNG} &
\fig[\sz]{resnet50attmaps/ILSVRC2012_val_00002848_attmap_28x28_resnet50_simpool_ep200_lr0.2.PNG} &
\fig[\sz]{resnet50attmaps/ILSVRC2012_val_00002848_attmap_28x28_resnet50_dino_gem_lnkv.PNG} \\

\fig[\sz]{resnet50attmaps/ILSVRC2012_val_00005427_orig.PNG} &
\fig[\sz]{resnet50attmaps/ILSVRC2012_val_00005427_attmap_28x28_resnet50_simpool_ep200_lr0.2.PNG} &
\fig[\sz]{resnet50attmaps/ILSVRC2012_val_00005427_attmap_28x28_resnet50_dino_gem_lnkv.PNG} & \hspace{2pt}
\fig[\sz]{resnet50attmaps/ILSVRC2012_val_00007405_orig.PNG} &
\fig[\sz]{resnet50attmaps/ILSVRC2012_val_00007405_attmap_28x28_resnet50_simpool_ep200_lr0.2.PNG} &
\fig[\sz]{resnet50attmaps/ILSVRC2012_val_00007405_attmap_28x28_resnet50_dino_gem_lnkv.PNG} & \hspace{2pt}
\fig[\sz]{resnet50attmaps/ILSVRC2012_val_00009219_orig.PNG} &
\fig[\sz]{resnet50attmaps/ILSVRC2012_val_00009219_attmap_28x28_resnet50_simpool_ep200_lr0.2.PNG} &
\fig[\sz]{resnet50attmaps/ILSVRC2012_val_00009219_attmap_28x28_resnet50_dino_gem_lnkv.PNG} \\

input &
supervised &
DINO &
input &
supervised &
DINO &
input &
supervised &
DINO \\

image &
\Ours &
\Ours &
image &
\Ours &
\Ours &
image &
\Ours &
\Ours \\

\end{tabular}
\vspace{3pt}
\caption{\emph{Attention maps} of ResNet-50~\cite{resnet} trained on \imagenet for 100 epochs under supervision and self-supervision with DINO~\cite{dino}. We use the attention map $\va$~\eq{sp-att}. Input image resolution: $896 \times 896$; output attention map: $28 \times 28$.}
\label{fig:attention-maps-resnet50}
\end{figure*}
\begin{figure*}
\scriptsize
\centering
\setlength{\tabcolsep}{1.2pt}
\newcommand{\sz}{.1035}
\begin{tabular}{ccccccccc}

\fig[\sz]{convnextmaps/ILSVRC2012_val_00000217_orig.PNG} &
\fig[\sz]{convnextmaps/ILSVRC2012_val_00000217_attmap_28x28_convnext_small_gap_cls.PNG} &
\fig[\sz]{convnextmaps/ILSVRC2012_val_00000217_attmap_28x28_convnext_small_dino_gem.PNG} & \hspace{2pt}
\fig[\sz]{convnextmaps/ILSVRC2012_val_00000807_orig.PNG} &
\fig[\sz]{convnextmaps/ILSVRC2012_val_00000807_attmap_28x28_convnext_small_gap_cls.PNG} &
\fig[\sz]{convnextmaps/ILSVRC2012_val_00000807_attmap_28x28_convnext_small_dino_gem.PNG} & \hspace{2pt}
\fig[\sz]{convnextmaps/ILSVRC2012_val_00000911_orig.PNG} &
\fig[\sz]{convnextmaps/ILSVRC2012_val_00000911_attmap_28x28_convnext_small_gap_cls.PNG} &
\fig[\sz]{convnextmaps/ILSVRC2012_val_00000911_attmap_28x28_convnext_small_dino_gem.PNG} \\

\fig[\sz]{convnextmaps/ILSVRC2012_val_00001213_orig.PNG} &
\fig[\sz]{convnextmaps/ILSVRC2012_val_00001213_attmap_28x28_convnext_small_gap_cls.PNG} &
\fig[\sz]{convnextmaps/ILSVRC2012_val_00001213_attmap_28x28_convnext_small_dino_gem.PNG} & \hspace{2pt}
\fig[\sz]{convnextmaps/ILSVRC2012_val_00001391_orig.PNG} &
\fig[\sz]{convnextmaps/ILSVRC2012_val_00001391_attmap_28x28_convnext_small_gap_cls.PNG} &
\fig[\sz]{convnextmaps/ILSVRC2012_val_00001391_attmap_28x28_convnext_small_dino_gem.PNG} & \hspace{2pt}
\fig[\sz]{convnextmaps/ILSVRC2012_val_00002929_orig.PNG} &
\fig[\sz]{convnextmaps/ILSVRC2012_val_00002929_attmap_28x28_convnext_small_gap_cls.PNG} &
\fig[\sz]{convnextmaps/ILSVRC2012_val_00002929_attmap_28x28_convnext_small_dino_gem.PNG} \\

\fig[\sz]{convnextmaps/ILSVRC2012_val_00003279_orig.PNG} &
\fig[\sz]{convnextmaps/ILSVRC2012_val_00003279_attmap_28x28_convnext_small_gap_cls.PNG} &
\fig[\sz]{convnextmaps/ILSVRC2012_val_00003279_attmap_28x28_convnext_small_dino_gem.PNG} & \hspace{2pt}
\fig[\sz]{convnextmaps/ILSVRC2012_val_00003735_orig.PNG} &
\fig[\sz]{convnextmaps/ILSVRC2012_val_00003735_attmap_28x28_convnext_small_gap_cls.PNG} &
\fig[\sz]{convnextmaps/ILSVRC2012_val_00003735_attmap_28x28_convnext_small_dino_gem.PNG} & \hspace{2pt}
\fig[\sz]{convnextmaps/ILSVRC2012_val_00004306_orig.PNG} &
\fig[\sz]{convnextmaps/ILSVRC2012_val_00004306_attmap_28x28_convnext_small_gap_cls.PNG} &
\fig[\sz]{convnextmaps/ILSVRC2012_val_00004306_attmap_28x28_convnext_small_dino_gem.PNG} \\

\fig[\sz]{convnextmaps/ILSVRC2012_val_00006007_orig.PNG} &
\fig[\sz]{convnextmaps/ILSVRC2012_val_00006007_attmap_28x28_convnext_small_gap_cls.PNG} &
\fig[\sz]{convnextmaps/ILSVRC2012_val_00006007_attmap_28x28_convnext_small_dino_gem.PNG} & \hspace{2pt}
\fig[\sz]{convnextmaps/ILSVRC2012_val_00006694_orig.PNG} &
\fig[\sz]{convnextmaps/ILSVRC2012_val_00006694_attmap_28x28_convnext_small_gap_cls.PNG} &
\fig[\sz]{convnextmaps/ILSVRC2012_val_00006694_attmap_28x28_convnext_small_dino_gem.PNG} & \hspace{2pt}
\fig[\sz]{convnextmaps/ILSVRC2012_val_00007129_orig.PNG} &
\fig[\sz]{convnextmaps/ILSVRC2012_val_00007129_attmap_28x28_convnext_small_gap_cls.PNG} &
\fig[\sz]{convnextmaps/ILSVRC2012_val_00007129_attmap_28x28_convnext_small_dino_gem.PNG} \\

\fig[\sz]{convnextmaps/ILSVRC2012_val_00007288_orig.PNG} &
\fig[\sz]{convnextmaps/ILSVRC2012_val_00007288_attmap_28x28_convnext_small_gap_cls.PNG} &
\fig[\sz]{convnextmaps/ILSVRC2012_val_00007288_attmap_28x28_convnext_small_dino_gem.PNG} & \hspace{2pt}
\fig[\sz]{convnextmaps/ILSVRC2012_val_00008952_orig.PNG} &
\fig[\sz]{convnextmaps/ILSVRC2012_val_00008952_attmap_28x28_convnext_small_gap_cls.PNG} &
\fig[\sz]{convnextmaps/ILSVRC2012_val_00008952_attmap_28x28_convnext_small_dino_gem.PNG} & \hspace{2pt}
\fig[\sz]{convnextmaps/ILSVRC2012_val_00009092_orig.PNG} &
\fig[\sz]{convnextmaps/ILSVRC2012_val_00009092_attmap_28x28_convnext_small_gap_cls.PNG} &
\fig[\sz]{convnextmaps/ILSVRC2012_val_00009092_attmap_28x28_convnext_small_dino_gem.PNG} \\

\fig[\sz]{convnextmaps/ILSVRC2012_val_00009842_orig.PNG} &
\fig[\sz]{convnextmaps/ILSVRC2012_val_00009842_attmap_28x28_convnext_small_gap_cls.PNG} &
\fig[\sz]{convnextmaps/ILSVRC2012_val_00009842_attmap_28x28_convnext_small_dino_gem.PNG} & \hspace{2pt}
\fig[\sz]{convnextmaps/ILSVRC2012_val_00010594_orig.PNG} &
\fig[\sz]{convnextmaps/ILSVRC2012_val_00010594_attmap_28x28_convnext_small_gap_cls.PNG} &
\fig[\sz]{convnextmaps/ILSVRC2012_val_00010594_attmap_28x28_convnext_small_dino_gem.PNG} & \hspace{2pt}
\fig[\sz]{convnextmaps/ILSVRC2012_val_00011650_orig.PNG} &
\fig[\sz]{convnextmaps/ILSVRC2012_val_00011650_attmap_28x28_convnext_small_gap_cls.PNG} &
\fig[\sz]{convnextmaps/ILSVRC2012_val_00011650_attmap_28x28_convnext_small_dino_gem.PNG} \\

\fig[\sz]{convnextmaps/ILSVRC2012_val_00013393_orig.PNG} &
\fig[\sz]{convnextmaps/ILSVRC2012_val_00013393_attmap_28x28_convnext_small_gap_cls.PNG} &
\fig[\sz]{convnextmaps/ILSVRC2012_val_00013393_attmap_28x28_convnext_small_dino_gem.PNG} & \hspace{2pt}
\fig[\sz]{convnextmaps/ILSVRC2012_val_00014741_orig.PNG} &
\fig[\sz]{convnextmaps/ILSVRC2012_val_00014741_attmap_28x28_convnext_small_gap_cls.PNG} &
\fig[\sz]{convnextmaps/ILSVRC2012_val_00014741_attmap_28x28_convnext_small_dino_gem.PNG} & \hspace{2pt}
\fig[\sz]{convnextmaps/ILSVRC2012_val_00015223_orig.PNG} &
\fig[\sz]{convnextmaps/ILSVRC2012_val_00015223_attmap_28x28_convnext_small_gap_cls.PNG} &
\fig[\sz]{convnextmaps/ILSVRC2012_val_00015223_attmap_28x28_convnext_small_dino_gem.PNG} \\

\fig[\sz]{convnextmaps/ILSVRC2012_val_00015516_orig.PNG} &
\fig[\sz]{convnextmaps/ILSVRC2012_val_00015516_attmap_28x28_convnext_small_gap_cls.PNG} &
\fig[\sz]{convnextmaps/ILSVRC2012_val_00015516_attmap_28x28_convnext_small_dino_gem.PNG} & \hspace{2pt}
\fig[\sz]{convnextmaps/ILSVRC2012_val_00020106_orig.PNG} &
\fig[\sz]{convnextmaps/ILSVRC2012_val_00020106_attmap_28x28_convnext_small_gap_cls.PNG} &
\fig[\sz]{convnextmaps/ILSVRC2012_val_00020106_attmap_28x28_convnext_small_dino_gem.PNG} & \hspace{2pt}
\fig[\sz]{convnextmaps/ILSVRC2012_val_00020451_orig.PNG} &
\fig[\sz]{convnextmaps/ILSVRC2012_val_00020451_attmap_28x28_convnext_small_gap_cls.PNG} &
\fig[\sz]{convnextmaps/ILSVRC2012_val_00020451_attmap_28x28_convnext_small_dino_gem.PNG} \\

\fig[\sz]{convnextmaps/ILSVRC2012_val_00022552_orig.PNG} &
\fig[\sz]{convnextmaps/ILSVRC2012_val_00022552_attmap_28x28_convnext_small_gap_cls.PNG} &
\fig[\sz]{convnextmaps/ILSVRC2012_val_00022552_attmap_28x28_convnext_small_dino_gem.PNG} & \hspace{2pt}
\fig[\sz]{convnextmaps/ILSVRC2012_val_00022787_orig.PNG} &
\fig[\sz]{convnextmaps/ILSVRC2012_val_00022787_attmap_28x28_convnext_small_gap_cls.PNG} &
\fig[\sz]{convnextmaps/ILSVRC2012_val_00022787_attmap_28x28_convnext_small_dino_gem.PNG} & \hspace{2pt}
\fig[\sz]{convnextmaps/ILSVRC2012_val_00023185_orig.PNG} &
\fig[\sz]{convnextmaps/ILSVRC2012_val_00023185_attmap_28x28_convnext_small_gap_cls.PNG} &
\fig[\sz]{convnextmaps/ILSVRC2012_val_00023185_attmap_28x28_convnext_small_dino_gem.PNG} \\

\fig[\sz]{convnextmaps/ILSVRC2012_val_00024643_orig.PNG} &
\fig[\sz]{convnextmaps/ILSVRC2012_val_00024643_attmap_28x28_convnext_small_gap_cls.PNG} &
\fig[\sz]{convnextmaps/ILSVRC2012_val_00024643_attmap_28x28_convnext_small_dino_gem.PNG} & \hspace{2pt}
\fig[\sz]{convnextmaps/ILSVRC2012_val_00027301_orig.PNG} &
\fig[\sz]{convnextmaps/ILSVRC2012_val_00027301_attmap_28x28_convnext_small_gap_cls.PNG} &
\fig[\sz]{convnextmaps/ILSVRC2012_val_00027301_attmap_28x28_convnext_small_dino_gem.PNG} & \hspace{2pt}
\fig[\sz]{convnextmaps/ILSVRC2012_val_00027374_orig.PNG} &
\fig[\sz]{convnextmaps/ILSVRC2012_val_00027374_attmap_28x28_convnext_small_gap_cls.PNG} &
\fig[\sz]{convnextmaps/ILSVRC2012_val_00027374_attmap_28x28_convnext_small_dino_gem.PNG} \\

\fig[\sz]{convnextmaps/ILSVRC2012_val_00028099_orig.PNG} &
\fig[\sz]{convnextmaps/ILSVRC2012_val_00028099_attmap_28x28_convnext_small_gap_cls.PNG} &
\fig[\sz]{convnextmaps/ILSVRC2012_val_00028099_attmap_28x28_convnext_small_dino_gem.PNG} & \hspace{2pt}
\fig[\sz]{convnextmaps/ILSVRC2012_val_00029763_orig.PNG} &
\fig[\sz]{convnextmaps/ILSVRC2012_val_00029763_attmap_28x28_convnext_small_gap_cls.PNG} &
\fig[\sz]{convnextmaps/ILSVRC2012_val_00029763_attmap_28x28_convnext_small_dino_gem.PNG} & \hspace{2pt}
\fig[\sz]{convnextmaps/ILSVRC2012_val_00030283_orig.PNG} &
\fig[\sz]{convnextmaps/ILSVRC2012_val_00030283_attmap_28x28_convnext_small_gap_cls.PNG} &
\fig[\sz]{convnextmaps/ILSVRC2012_val_00030283_attmap_28x28_convnext_small_dino_gem.PNG} \\

input &
supervised &
DINO &
input &
supervised &
DINO &
input &
supervised &
DINO \\

image &
\Ours &
\Ours &
image &
\Ours &
\Ours &
image &
\Ours &
\Ours \\

\end{tabular}
\vspace{3pt}
\caption{\emph{Attention maps} of ConvNeXt-S~\cite{convnext} trained on \imagenet for 100 epochs under supervision and self-supervision with DINO~\cite{dino}. We use the attention map $\va$~\eq{sp-att}. Input image resolution: $896 \times 896$; output attention map: $28 \times 28$.}
\label{fig:attention-maps-convnext}
\end{figure*}
%------------------------------------------------------------------------------
%------------------------------------------------------------------------------

\end{document}